\documentclass{article}

\usepackage{longtable}
\usepackage{pdflscape}

\usepackage{tabularx}
\usepackage{array}
\usepackage{ragged2e}
\usepackage{hyperref}

\usepackage{algorithm} 
\usepackage{algpseudocode} 
\usepackage[title]{appendix}

\usepackage[utf8]{inputenc} 
\usepackage[T1]{fontenc}    
\usepackage{hyperref}       
\usepackage{url}            
\usepackage{booktabs}       
\usepackage{amsfonts}       
\usepackage{nicefrac}       
\usepackage{microtype}      
\usepackage{lipsum}
\usepackage{fancyhdr}       
\usepackage{graphicx}       
\graphicspath{{./images/}}     
\usepackage[export]{adjustbox}   
\usepackage{natbib}              
\usepackage[english]{babel}
\usepackage{graphicx}
\usepackage{longtable}
\usepackage{booktabs}
\usepackage{adjustbox}

\usepackage{caption}
\usepackage{lastpage}
\title{SPROCKET: Extending ROCKET to Distance-Based Time-Series Transformations With Prototypes
}

\author{
	Nicholas Harner \\
	Independent Researcher \\
	Dayton, OH\\
	\texttt{memorialsignin@protonmail.com} \\
}

\begin{document}
	
\maketitle 

\begin{abstract}
Classical Time Series Classification algorithms are dominated by feature engineering strategies. One of the most prominent of these transforms is ROCKET, which achieves strong performance through random kernel features. We introduce SPROCKET (\textbf{S}elected \textbf{P}rototype \textbf{R}and\textbf{o}m \textbf{C}onvolutional \textbf{Ke}rnel \textbf{T}ransform), which implements a new feature engineering strategy based on prototypes. On a majority of the UCR and UEA Time Series Classification archives, SPROCKET achieves performance comparable to existing convolutional algorithms and the new MR-HY-SP ( MultiROCKET-HYDRA-SPROCKET) ensemble's average accuracy ranking exceeds HYDRA-MR, the previous best convolutional ensemble's performance. These experimental results demonstrate that prototype-based feature transformation can enhance both accuracy and robustness in time series classification.
\end{abstract}

\section{Introduction}
Time series classification algorithms have undergone rapid development recently, including the introduction of a new family of algorithms based on convolutional kernels for feature extraction. ROCKET\cite{Dempster_2020} exemplified these methods and achieved extremely high accuracy with minimal computation time. Rocket was refined into MiniROCKET\cite{Dempster_2021} and MultiROCKET\cite{tan2022multirocketmultiplepoolingoperators}, which constrained the random selection of convolutional kernels and expanded the pooling operations, respectively. Convolutional methods were then expanded to include HYDRA\cite{dempster2022hydracompetingconvolutionalkernels}, which incorporated dictionary-like methods for extracting time-series subsequences into a ROCKET-like classifier through competing convolutional kernels. The 2023 Bake-Off \cite{Middlehurst_2024} established that an ensemble of MultiROCKET and Hydra was a top performing classifier, surpassed only by the HIVE-COTE ensemble which requires substantially greater computational resources. However, these rankings may have changed with the development of ConvTran\cite{Foumani_2023}. Convolutional feature extraction strategies have been remarkably successful in the time series domain. Their relative simplicity and applicability to both classical and deep learning methods make them a promising avenue for continued research.

Motivated by the success of combining dictionary-like approaches into convolutional methods, SPROCKET incorporates distance measures into ROCKET with minimal changes to the ROCKET architecture. Distances, especially elastic distances, have an extensive history in time series problems and are one of the algorithm families in the Bake-Off\cite{Middlehurst_2024}.  However, they are not trivially compatible with convolutional methods, since they require pairwise comparisons. SPROCKET addresses this issue through a randomized prototyping strategy to achieve baseline accuracy competitive with both ROCKET and HYDRA. Furthermore, the addition of SPROCKET to the HYDRA-MR ensemble improves its performance on a majority of the UCR and UEA Time Series Classification datasets \cite{bagnall2018ueamultivariatetimeseries} \cite{dau2019ucrtimeseriesarchive}.

\section{Intuition}
Distances have been central to time series classification since the development of DTW and NN-DTW served as the baseline for the 2016 Bake-Off\citep{bagnall2016greattimeseriesclassification}. Distance-based approaches remain prominent in clustering \citep{paparrizos2024bridginggapdecadereview} and time series specific distance measures continue to develop, with the recent introduction of amercing \citep{herrmann2021amercingintuitiveeleganteffective} and its incorporation into the Proximity Forest 2.0  \cite{herrmann2023proximityforest20new}. As a robust nonparametric family of classification algorithms, distances have many attractive features. However, they face several fundamental limitations.

\begin{itemize}
\item Elastic distances are computationally expensive.
\subitem Elastic distance measures require a solution to a dynamic programming problem with computational complexity of $O(l^2)$, where $l$ is the length of the series. The search space can be constrained to lower the complexity to $O(lw)$, where $w<<l$.
\item Distances capture only the some of a datasets properties.
\subitem Each distance imposes a single set of relations on a dataset. This single relation inherently limits the diversity of the learnable information from a dataset-distance pair.
\end{itemize}

The first issue can be addressed by selecting prototypes. Prototype based learning relies on representative samples of the dataset, called prototypes, to serve as references for all other points. For each prototype $p$ and instance $a$, the distance $d(p,a)$ is a similarity measure between the pair. With a well selected prototype set, this signal can be used to derive useful information for all points in the dataset. Prototype selection reduces the computational burden of metric learning by limiting the number of points of comparison and their resultant distance calculations.

The issue of a limited derivable information given a dataset and a distance can be partially addressed by utilizing many different similarity measures and transformations of a time series. Taking derivatives is a classic example of this, as seen in DDTW, which applies derivatives to a series before taking the DTW distance. This transformation has been successful in the classification domain and the first derivative is incorporated into the Proximity Forest 2.0 as a baseline strategy for increasing ensemble diversity \cite{herrmann2023proximityforest20new}. However the exceptional performance of convolutional kernels for time series classification suggests an additional approach. Apply the kernels from ROCKET as a feature transformation, then use a distance based approach in this new space. This observation motivates SPROCKET.

\section{ROCKET Architecture and Modification}
Starting with ROCKET is a reasonable first step in combining ROCKET and distances. The transform has one hyperparameter in its base form and only requires a specified number of kernels.

\begin{algorithm}
	\caption{ROCKET} 
	\begin{algorithmic}[1]
		\State Set the number of kernels $K$ for a training set $X$.
		\For {$kernel$ $k_i$ $=1,2,\ldots K$}
		\State Parameterize kernel $k_i$, selecting a random set of weights, number of weights, and dilation.
		\For {training point $p_j$ $ j=1,2,\ldots,|X|$}
		\State Calculate $k_i(p_j)$.
		\State Calculate $M$ pooling features (PPV, Max, and so on) for $k_i(p_j)$.
		\EndFor
		\EndFor
		\State Create matrix $A$ of shape $(|X|, KM)$ and populate it with the pooling features.
		\State Fit a Cross Validated Ridge predictor on $A$.
	\end{algorithmic} 
\end{algorithm}

The key steps of this algorithm are (i) applying kernels, (ii) pooling features, and (iii) using a model. Of these three, the pooling step is the natural place to modify the framework. This is also where prototype selection is required. Note that prototypes are essential here, since the predictor trained on the pooling features requires those features to be consistent. Even if $d(a,b) = d(c,d)$ these values could provide minimal information for a model depending on the relative positions of the points $a, b, c, d$ in the space. In contrast, $d(a,b) = d(a,c)$ is far more likely to be meaningful, since the common point $a$ is included in both features. The shared reference point should greatly increase the likelihood that features are informative for a model.

\begin{algorithm}
\caption{SPROCKET} 
\begin{algorithmic}[1]
\State Set the number of kernels $K$ and number of prototypes $M$ for a training set $X$.
\For {$kernel$ $k_i$ $=1,2,\ldots K$}
\State Parameterize kernel $k_i$ as in ROCKET.
\For {training point $p_j$ $ j=1,2,\ldots,|X|$}
\State Calculate $k_i(p_j)$.
\EndFor
\State Create a prototype set $C_{k_i}$ for each kernel $k_i$, where $C_{k_i} \subset k_i(X), |C_{k_i}| = M$.
\For {training point $p_j$ $ j=1,2,\ldots,|X|$}
\State Calculate the distances $d(c,k_i(p_j))$ between all prototypes in $c\in C_{k_i}$ and $p_j$.
\EndFor
\EndFor
\State Create matrix $A$ of shape  $(|X|, KM)$ and populate it with the calculated distances.
\State Fit a Cross Validated Ridge predictor on $A$.

\end{algorithmic} 
\end{algorithm}

Now there are three remaining key design decisions. SPROCKET requires a distance measure, a prototype selection method, and a number of prototypes. 

\subsection{Number of Prototypes}
SPROCKET needs a number of prototypes, but without the other portions of the algorithm defined, the best heuristic is not obvious. However, we can describe two desirable properties of any method:

\begin{itemize}
\item The number of prototypes should be small.
\subitem ROCKET uses thousands of convolutional kernels in its default configuration. If the number of prototypes is large, SPROCKET will quickly become computationally infeasible as the number of distances calculations increases.
\item The number of prototypes should increase with dataset size.
\subitem NN and other distance based learning methods are consistent, even if their convergence is slow. As datasets increase in size, they could exhibit richer structure that requires more prototypes to capture.
\end{itemize}

These properties compete. As a compromise, the number of prototypes can be set to a monotonic, but slowly increasing function. Any logarithmic function on the size of the dataset would achieve this goal, but for experimental stability it should be consistent. In this paper, the size of the prototype set is always
\begin{equation}
\lceil log_4(|X|) \rceil 
\end{equation} rounding up to ensure a minimum number of prototypes for small datasets. The datasets tested in this paper range from $10-5000$ training points, which results in $2-7$ prototypes per kernel. Alternative strategies are possible and should be explored, but this paper will only demonstrate a minimal baseline algorithm.

\subsection{Prototype Selection}
Prototypes must be chosen, but most principled forms of prototype selection require computing many pairwise distances. Since the number of computed distance measures dominate the runtime differential between SPROCKET and other convolutional algorithms, any increases to computational load should be carefully considered.

Two baseline prototype selection algorithms require no additional distance calculations.
\begin{enumerate} 
	\item Uniform Random Sampling of prototypes from the training set. 
	\item Stratified Random Sampling of prototypes from the training set in proportion to their class prevalence.  
\end{enumerate} 
Both are inexpensive, unbiased, scalable, simple to implement, and standard solutions to this issue.

A natural alternative approach is to perform all or part of a clustering algorithm that provides centroids or medoids. For instance, K-Means++ \cite{arthur2007kmeansplusplus} has natural applications to time series clustering and efficient variants such as KASBA \cite{holder2024rockkasbablazinglyfast} for time series specific measures. The initialization step of Kmeans++ is:

\begin{algorithm}
\caption{Kmeans++ Initialization} 
\begin{algorithmic}[1]
\State Add one center $c$ to the center set $C$, chosen uniformly at random among the data points $X$.
\State For each data point $x\in X\setminus C$, compute $D(x,C)$, or the distance between $x$ and the nearest center in $C$.
\State Add $c_{new}$ to $C$, where $c_{new}$ is chosen with probability proportional to $D(x,C)$ for all $x \in X\setminus C$.
\State Repeat Steps 2 and 3 until $k$ centers have been chosen.
\end{algorithmic} 
\end{algorithm}
Initialization therefore requires, at minimum $|X|-1$ distance calculations. In the worst case, each center added to $C$ will require $|X|-|C|$ distance calculations. Exploiting the triangle inequality can avoid distance calculations if the distance function is a metric, but the worst case initialization of KMeans++ will approximately double the required number of distance calculations relative to random selection. This does not include calculations for subsequent iterations Kmeans itself, which must in turn be applied for all of the hundreds or thousands of kernels in a ROCKET framework. 

The goal of this study is to establish a minimum baseline to combine distance learning with ROCKET. Although prototype selection offers potential to improve accuracy, it comes with substantial computational cost above random selection. Therefore, this paper will confine prototype selection strategies to random sampling, leaving exploration of alternatives to future work.

\subsection{Distance Measures}
Any design requires a set of distance measures used for prototype features. There are many possible measures to consider, but the distance measures present in the Aeon toolkit  \cite{middlehurst2024aeonpythontoolkitlearning} form a readily implementable baseline for this study. For detailed descriptions, we refer the reader to \cite{Holder_2023}.
The selected measures for this study are:

\begin{table}[h!]
	\centering
	\begin{tabularx}{\textwidth}{||c|c|>{\RaggedRight\arraybackslash}X||}
		\hline
		\textbf{Long Name} & \textbf{Abbreviation} & \textbf{Notes} \\[0.5ex]
		\hline\hline
		Euclidean & Euclidean & Baseline non-elastic measure \\
		\hline
		Dynamic Time Warping & DTW & Baseline elastic distance \\
		\hline
		Weighted DTW & WDTW & DTW with a weighted warping penalty \\
		\hline
		Amerced DTW & ADTW & DTW with amercing \\
		\hline
		Edit Distance with Real Penalty & ERP & Levenshtein-style distance with real valued data \\
		\hline
		Time Warp Edit & TWE & Edit distance with explicit time-warping penalty \\
		\hline
		Move-Split-Merge & MSM & Edit-based with local warping penalties \\
		\hline
	\end{tabularx}
	\caption{Elastic and non-elastic distance measures evaluated.}
\end{table}

This set is not exhaustive but does provide a diverse baseline. Exploring additional measures will be left to later work.

\section{Algorithm Analysis}

We will now analyze SPROCKET's computational complexity to determine its scaling factors. Because pairwise distance calculations dominate the runtime of the transform, this analysis will focus on those.

The SPROCKET transform creates $\lceil log_b(|X|) \rceil $ prototypes for each for each kernel, where $b$ is a constant (set to $4$ for our experiments), and $k$ kernels. Now let the dataset have $|X|$ instances, $m$ channels, and length $l$ on each channel. For multivariate series, let the distances be calculated independently on each channel. Then, the number of distance calculations $N_d$ to perform the transform is:
\begin{equation}
N_d = k \cdot n \cdot \log_b(n) \cdot m
\end{equation}

If an elastic distance measure with a Sakoe-Chiba band $w\in {2, 3 ... l}$ is used, then a single distance measure costs $O(lw)$. So SPROCKET transform, excluding the convolutional kernel computational cost inherited from ROCKET costs
\begin{equation}
O(k \cdot n \cdot log_b(n) \cdot m \cdot D)=O(k \cdot n \cdot log_b(n) \cdot m \cdot l \cdot w)
\end{equation}
For non-elastic distances such as the Euclidean distance, the runtime is reduced by a factor of $w$, since the distance calculation has $O(l)$. This complexity equation can then be compared to the Proximity Forest's time complexity, which is:
\begin{equation}
O(n \cdot log_2(n) \cdot m \cdot l \cdot w \cdot r \cdot c \cdot t)
\end{equation}
Where $c$ is the number of classes, $r$ the number of candidate splits per node, and $t$ the number of trees. SPROCKET will have lower complexity in the cases where
\begin{equation}
	r \cdot c \cdot t > k \land b > 2
\end{equation} Since SPROCKET's scaling is independent of the number of classes, it will hold a relative scaling advantage over other distance based methods in multi-class problems. It will also be at a relative advantage in cases where the Proximity Forest requires many splits per node, assuming the construction of the resulting classifier is computationally negligible relative to feature extraction.

\section{Experimental Results}
\subsection{Settings}
For all experiments, parameters were held constant unless otherwise noted. All elastic distances were calculated with a window parameter for the Sakoe-Chiba band set to $\lfloor \sqrt{l} \rfloor$, features were prepared in parallel with 20 threads, and elastic distance parameters were otherwise untuned, using the Aeon-Toolkit's default settings.

The ROCKET variant classification models were trained with scikit-learn's RidgeClassifierCV() classifier with the default settings.

All experiments were conducted on a System76 Adder WS running PopOS, with 16 GB of Memory, and 32 13th Gen Intel® Core™ i9-13900HX cores.

\subsection{Prototype and Distance Selection}

From Section 3, we have two unresolved design concerns:
\begin{itemize}
\item Which distances to use?
\item How to select prototypes?
\end{itemize}

We can start with a dual mode experiment. We will evaluate the SPROCKET algorithm on a representative subset of the UCR archive twice for each distance measure. The first time, we will select the prototypes completely randomly and the second time we will use stratified random sampling. This will yield two tests of the relative rankings for each distance measure and seven tests of the prototyping strategy, one for each distance measure. The selected datasets span a variety of applications and lengths, a full description is included in Appendix A. For this test, we set the number of kernels to be $512$ for all distance measures.

First, we can consider the two relative results for distance measures with true and stratified randomly sampled prototypes.

\begin{figure}[htbp]
\centering
\includegraphics[width=1\textwidth]{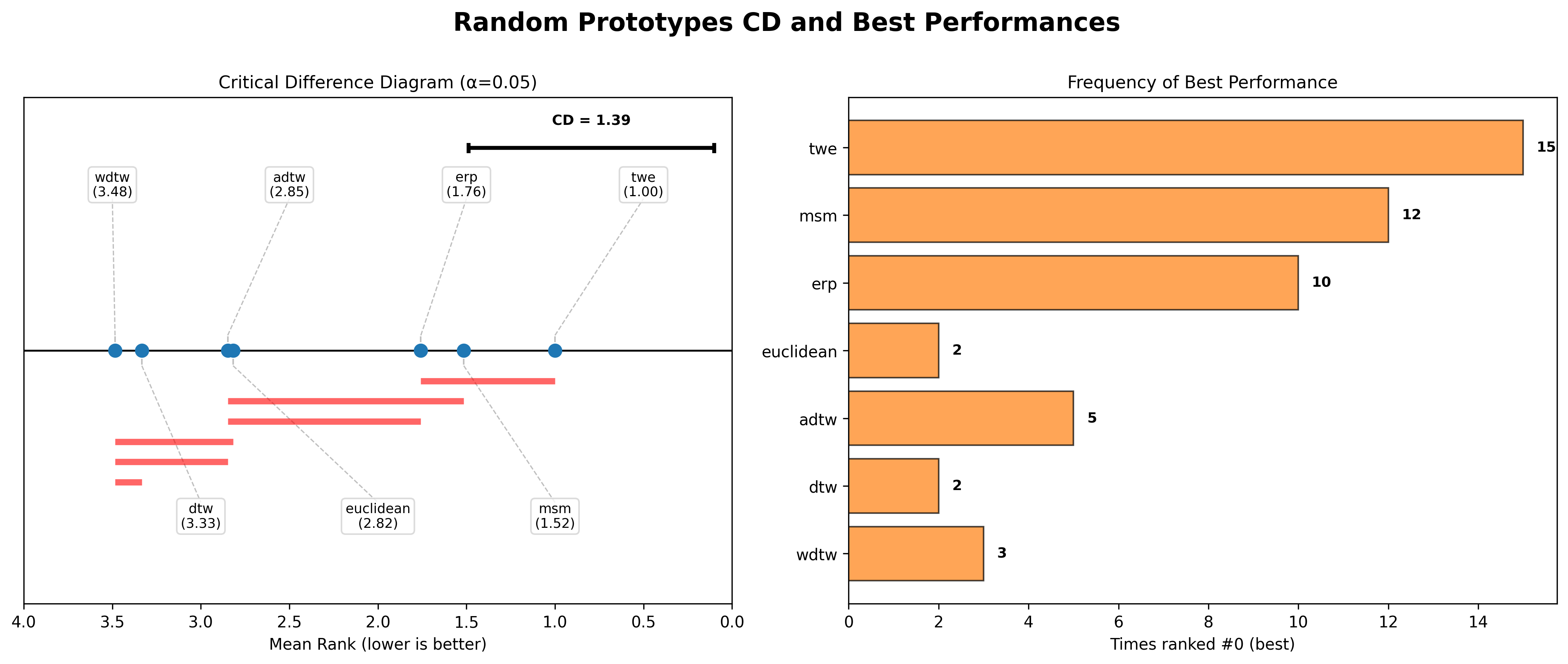}

\caption{Relative comparisons of Distance Measures with the SPROCKET Transform.}
\label{fig:random_sampling_CD_for_distances}
\end{figure}

\begin{figure}[htbp]
\centering
\includegraphics[width=1\textwidth]{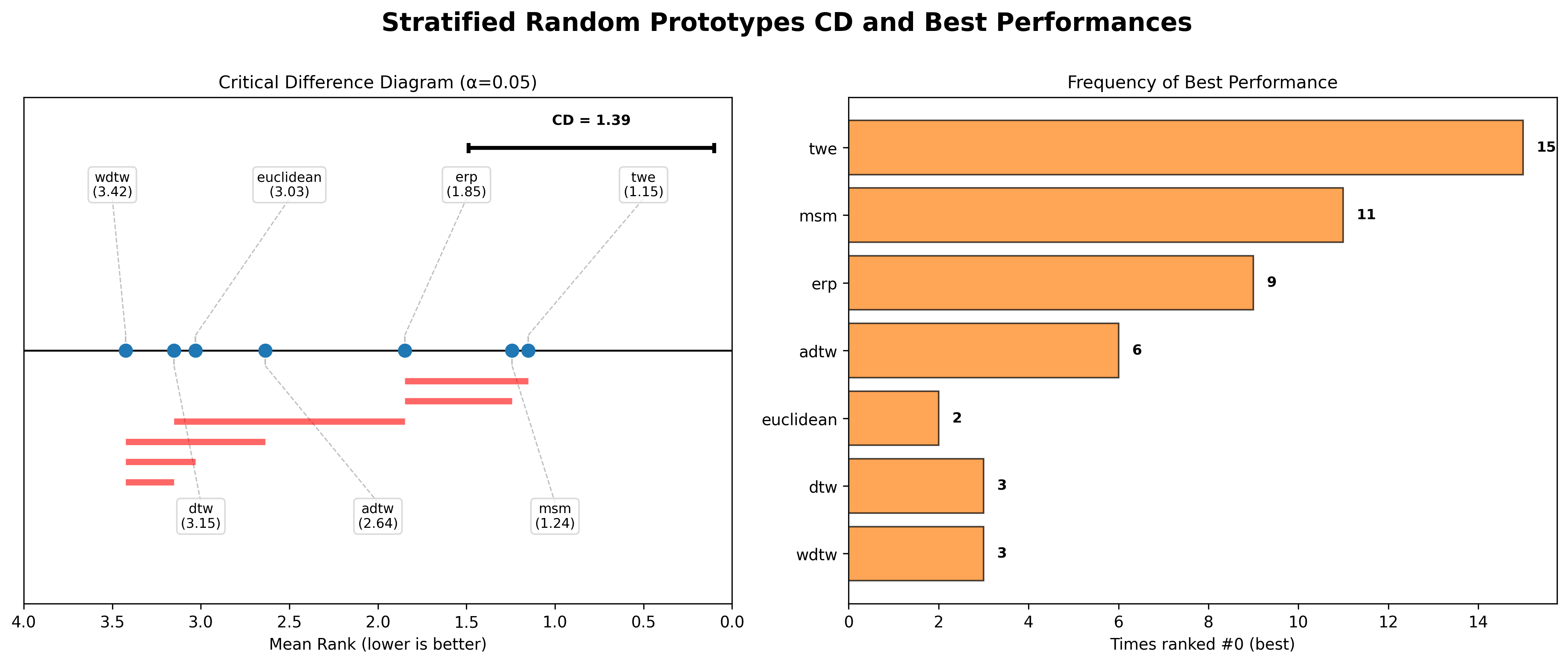}

\caption{Stratified Results maintain a similar rank ordering to true random.}
\label{fig:stratified_sampling_CD_for_distances}
\end{figure}

We can infer some information from these tests about the relative performance of the distance measures. The edit-based distances all perform well, with TWE, MSM, and ERP maintaining the first, second, and third best performing distances in each set of tests. Meanwhile, the warping based distances underperform, with WDTW and DTW failing to outperform the Euclidean baseline on average and ADTW outperforming Euclidean distances only once. However, all of the warping distances are the top performing distances on at least two datasets in both tests and all distances are the top performers on several datasets. We can therefore conclude that the warping distances are not strictly dominated in accuracy, but do perform worse on average than the edit based distances.

We can also examine the relative time each distance takes to prepare the prototype features.

\begin{figure}[htbp]
\centering

\includegraphics[width=1\textwidth]{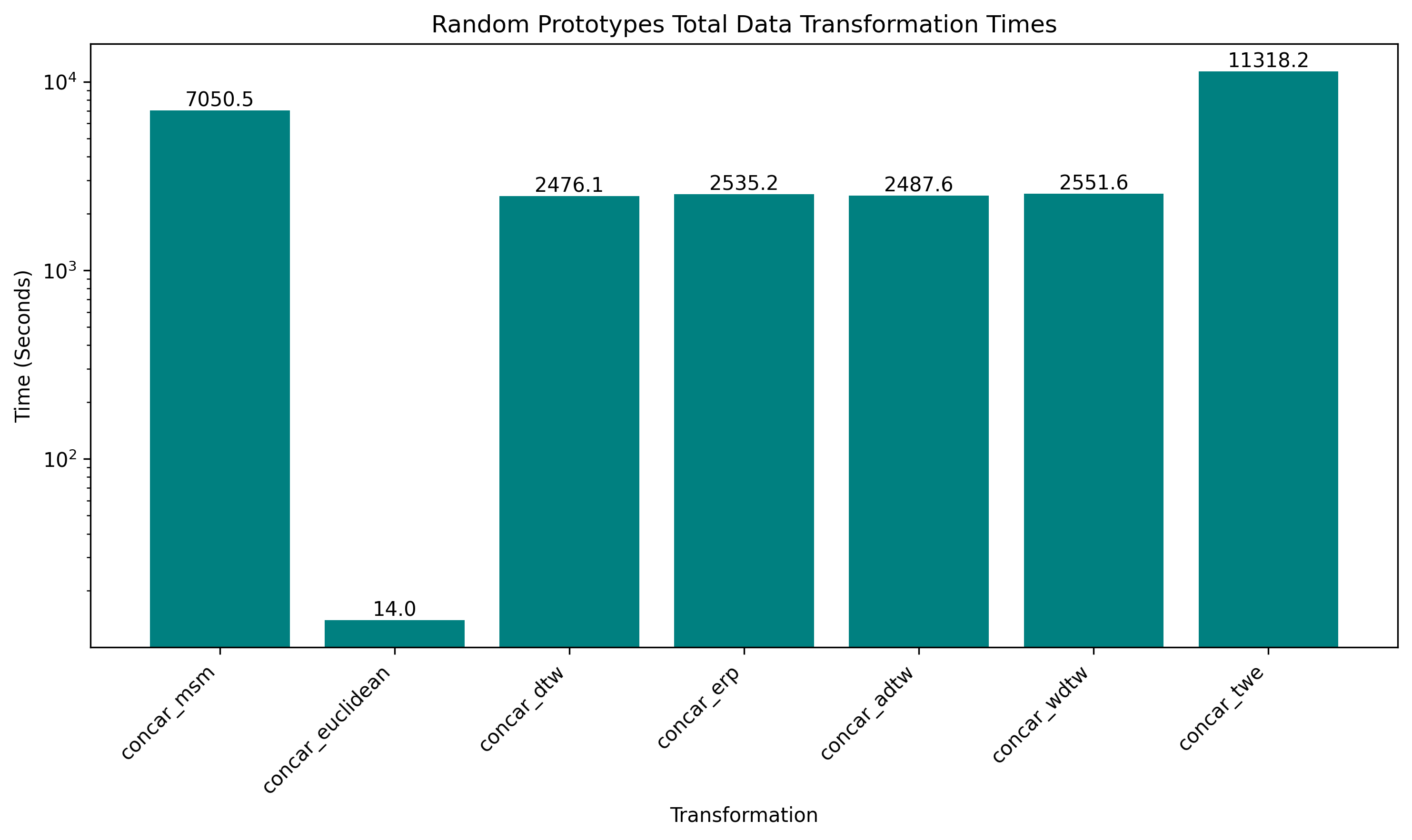}

\caption{Total Feature Transformation Time For Each Distance, Random Prototypes.}
\label{fig:random_sampling_times}
\end{figure}

Here, we have four groups. We have the baseline times for all the warping based distances and TWE. The other two edit based distances take longer, MSM taking $2.8$ times longer than the baseline set of $\{EDR, DTW, ADTW, WDTW\}$ and TWE taking $4.5$ times baseline. Finally, the Euclidean distance is two orders of magnitude faster than all elastic distances, taking under $0.01$ of the base time. Unfortunately, on these datasets the most accurate distances are also the most computationally expensive by substantial factors.

Finally, the distances could be valuable in an ensemble. To examine this possibility, we can consider several ensemble statistics for each distance's predictions.

\begin{figure}[htbp]
\centering
\includegraphics[width=1\textwidth]{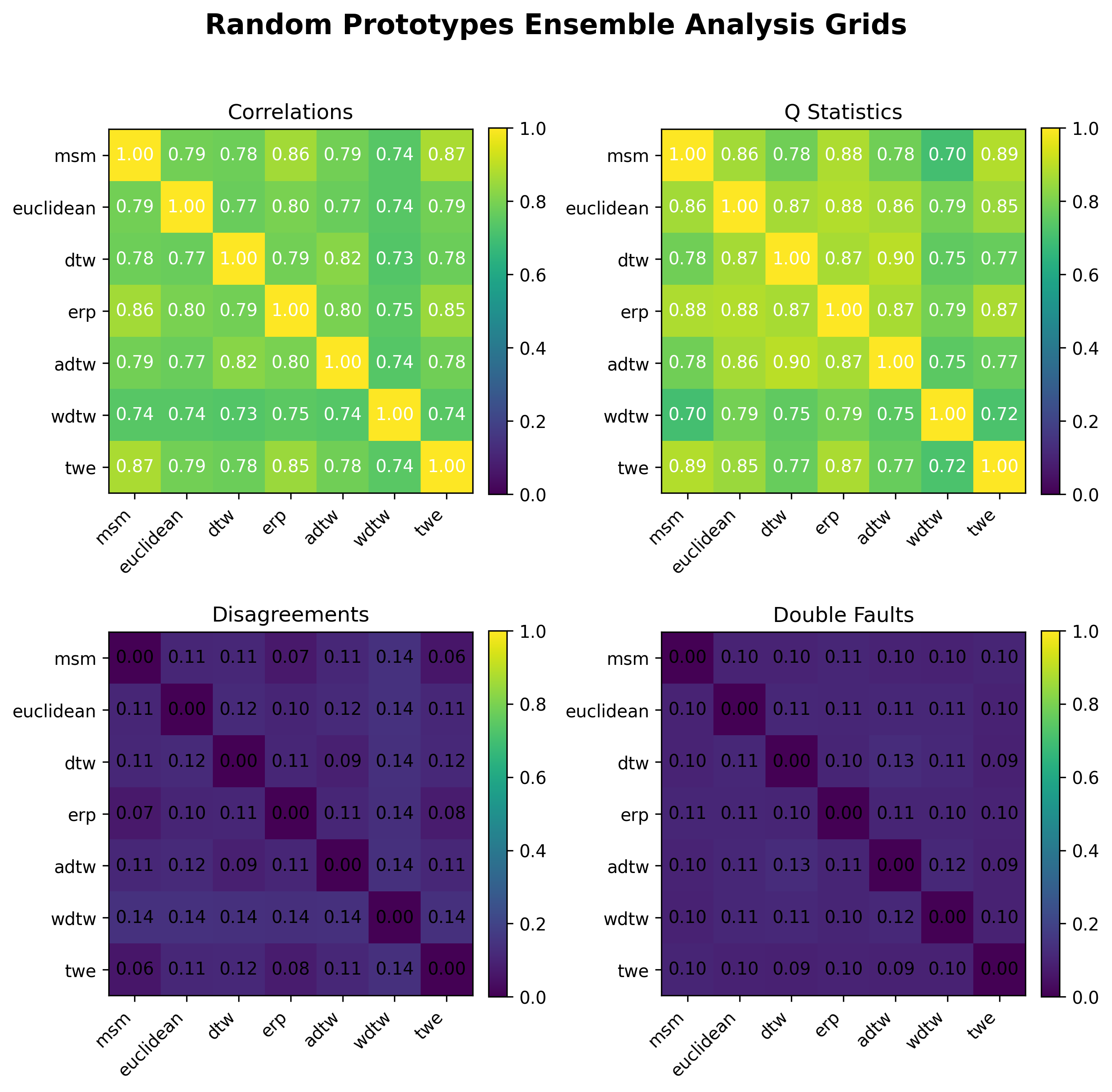}

\caption{Ensemble Grid for Random Prototypes.}
\label{fig:random_prototypes_ensemble_grid}
\end{figure}

\begin{figure}[htbp]
\centering
\includegraphics[width=1\textwidth]{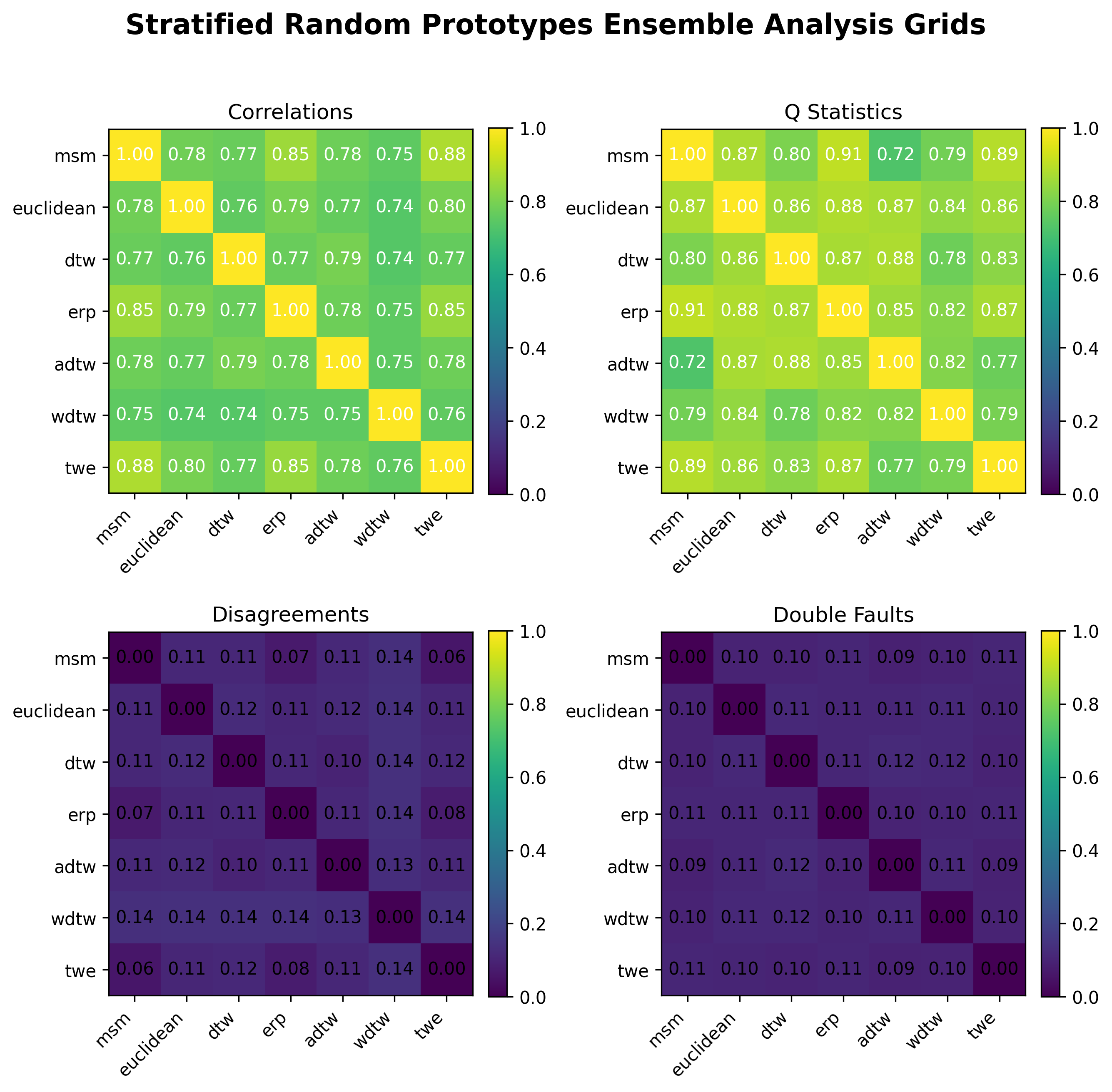}

\caption{Ensemble Grid for Stratified Random Prototypes.}
\label{fig:stratified_random_ensemble_grid}
\end{figure}

All of the distances are highly correlated with each other on these datasets, with a lowest correlation $0.70$ (between DTW and WDTW with Stratified Random Prototypes) and a highest correlation of $0.89$ (between MSM and TWE with Stratified Random Prototypes). The other ensembling statistics are similarly discouraging. We should also note the high correlation between the edit distances. They are far more correlated with each other than with the warping based distances, suggesting that the relative performance improvements from ensembling them will be limited on similar data, even though they are relatively more accurate than the warping based distances.

Finally, we can compare the two prototype selection strategies. The table below counts the number of times Random Selection and Stratified Random Selection outperformed on the $33$ datasets in our small sample for each distance measure.

\begin{table}[h!]
	\centering
	\begin{tabularx}{\textwidth}{||c|>{\RaggedRight\arraybackslash}X|>{\RaggedRight\arraybackslash}X|>{\RaggedRight\arraybackslash}X||}
		\hline
		\textbf{Distance} & \textbf{Random Wins} & \textbf{Stratified Wins} & \textbf{Ties} \\
		\hline\hline
		Euclidean & 17 & 12 & 4 \\
		\hline
		DTW & 14 & 15 & 4 \\
		\hline
		ADTW & 16 & 14 & 3 \\
		\hline
		WDTW & 12 & 14 & 7 \\
		\hline
		ERP & 17 & 12 & 4 \\
		\hline
		MSM & 14 & 14 & 5 \\
		\hline
		TWE & 18 & 11 & 4 \\
		\hline
		Total & 108 & 92 & 31 \\
		\hline
	\end{tabularx}
	\caption{Relative Comparisons of Random and Stratified Random Accuracies on $33$ Datasets with $7$ Distance Measures.}
\end{table}

These results suggest a weak advantage for true random sampling over stratified random sampling on the UCR datasets. Performing a one-sided sign test to determine whether random sampling is preferable to stratified random sampling with the $200$ total nontied trials results in a p-value of $0.14$ in favor of true random. Though this is not statistically significant, combining these results with the relative simplicity advantage of true random sampling over stratified suggests that true random is acceptable for this initial exploratory analysis of SPROCKET. We used true random sampling as the default for the remainder of this paper.

\subsection{Ensemble Experiments}
The previous section suggested that combining multiple distances is unlikely to increase SPROCKET accuracy or reduce computation time for SPROCKET due to the high correlation of distance measures. However, ensembling is well established in the time series distance learning literature and we wished to test whether it would help even in this case. To test this possibility, we conducted a limited test of potential ensembles.

To begin, we divided the distance measures into three groups based on their conceptual framework and the empirical results of our experiments.

\begin{table}[h!]
	\centering
	\begin{tabularx}{\textwidth}{||c|
			>{\RaggedRight\arraybackslash}X|
			>{\RaggedRight\arraybackslash}X|
			>{\RaggedRight\arraybackslash}X|
			>{\RaggedRight\arraybackslash}X||}
		\hline
		\textbf{Label} & \textbf{Distances} & \textbf{Accuracy} & \textbf{Time} & \textbf{Correlation} \\[0.5ex]
		\hline\hline
		Inelastic & Euclidean & Moderate & Very Fast & High \\
		\hline
		Warping & ADTW, DTW, WDTW & Moderate & Slow & High \\
		\hline
		Edit & TWE, MSM, ERP & High & Slow--Very Slow & Very High \\
		\hline
	\end{tabularx}
	\caption{Distance sets and kernel characteristics.}
\end{table}

Then we adopted an ensembling strategy. Ideally, we would validate the various distance measures on holdout data and form an ensembling strategy based on those results, as seen in the Elastic Ensemble\cite{elastic_ensemble}. But that is not necessary for the other convolutional algorithms and would incur substantial additional computational requirements. Instead, for a simple, default version of SPROCKET, we will use a constant fraction of the allotted convolutional kernels and see which ensemble, if any, performs the best.

We tested six ensembles and seven unensembled distance measures on the same subset of the UCR Benchmark in the previous sections. All classifiers will have either $600$ or $1200$ kernels. 

\begin{table}[h!]
	\centering
	\begin{tabularx}{\textwidth}{||c|>{\RaggedRight\arraybackslash}X|>{\RaggedRight\arraybackslash}X||}
		\hline
		\textbf{Label} & \textbf{Distances} & \textbf{Kernel Counts} \\
		\hline\hline
		Top2 & TWE, ADTW & 300, 300 \\
		\hline
		Top2+e & TWE, ADTW, Euclid & 300, 300, 600 \\
		\hline
		Top4 & TWE, ADTW, MSM, DTW & 150, 150, 150, 150 \\
		\hline
		Top4+e & TWE, ADTW, MSM, DTW, Euclid & 150, 150, 150, 150, 600 \\
		\hline
		All Elastic & 
		TWE, ADTW, MSM, DTW, ERP, WDTW & 
		100, 100, 100, 100, 100, 100 \\
		\hline
		All & 
		TWE, ADTW, MSM, DTW, ERP, WDTW, Euclid & 
		100, 100, 100, 100, 100, 100, 600 \\
		\hline
	\end{tabularx}
	\caption{Distance sets and kernel counts.}
\end{table}

These ensemble combinations of distances were chosen to consider contributions from each of the three groups, balance computational considerations, and manage experimental complexity. We will consider equal contributions from each of the two Elastic distance groups, and including the most accurate, two most accurate, and all distances from each of those groups. The three best performing distances in the previous section were all highly correlated and had Q-statistics between $0.86-.089$. This suggests that within group ensembling will be of limited utility and, given the computational limitations at hand, are not worth exploring in this initial assessment. Instead, this test focuses on cross group diversity. There are two equal contributions from each of the two Elastic distance groups, taking the two most accurate from each elastic group in Top2, the four most accurate in Top4, and all distances from each of those groups. Finally, since the Euclidean distance requires orders of magnitude less computation, it can be included with minimal additional computational cost on top of the Elastic distance kernels.

\begin{figure}[htbp]
\centering

\includegraphics[width=1\textwidth]{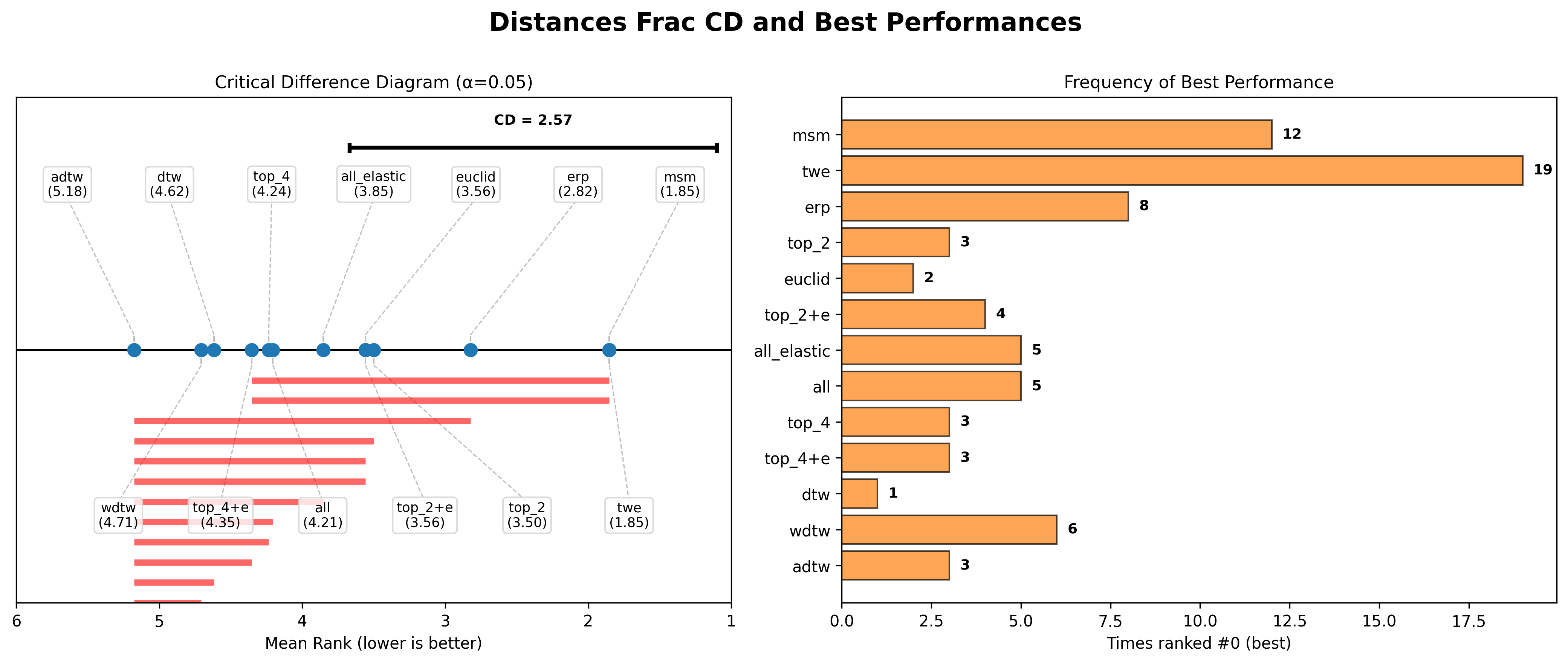}

\caption{Relative Performance of Ensembles and Base Classifiers.}
\label{fig:ensemble CD}
\end{figure}

The top performing ensembles on this test (Top2 and Top2+e) were both outperformed by all edit based versions of SPROCKET (TWE, MSM, and ERP). Top2+e held the same average rank as the Euclidean SPROCKET, and Top2 held a slightly worse average rank, though the difference was small at $0.06$. Both also achieved the best possible rank more times than the Euclidean SPROCKET, with $3$ and $4$ first ranks as compared to the Euclidean $2$, but less than all edit distances, with $8$, $12$, and $19$ best ranks for ERP, MSM, and TWE respectively.

We can also examine the ensembling statistics for all tested classifiers.

\begin{figure}[htbp]
	\centering
	
	\includegraphics[width=1\textwidth]{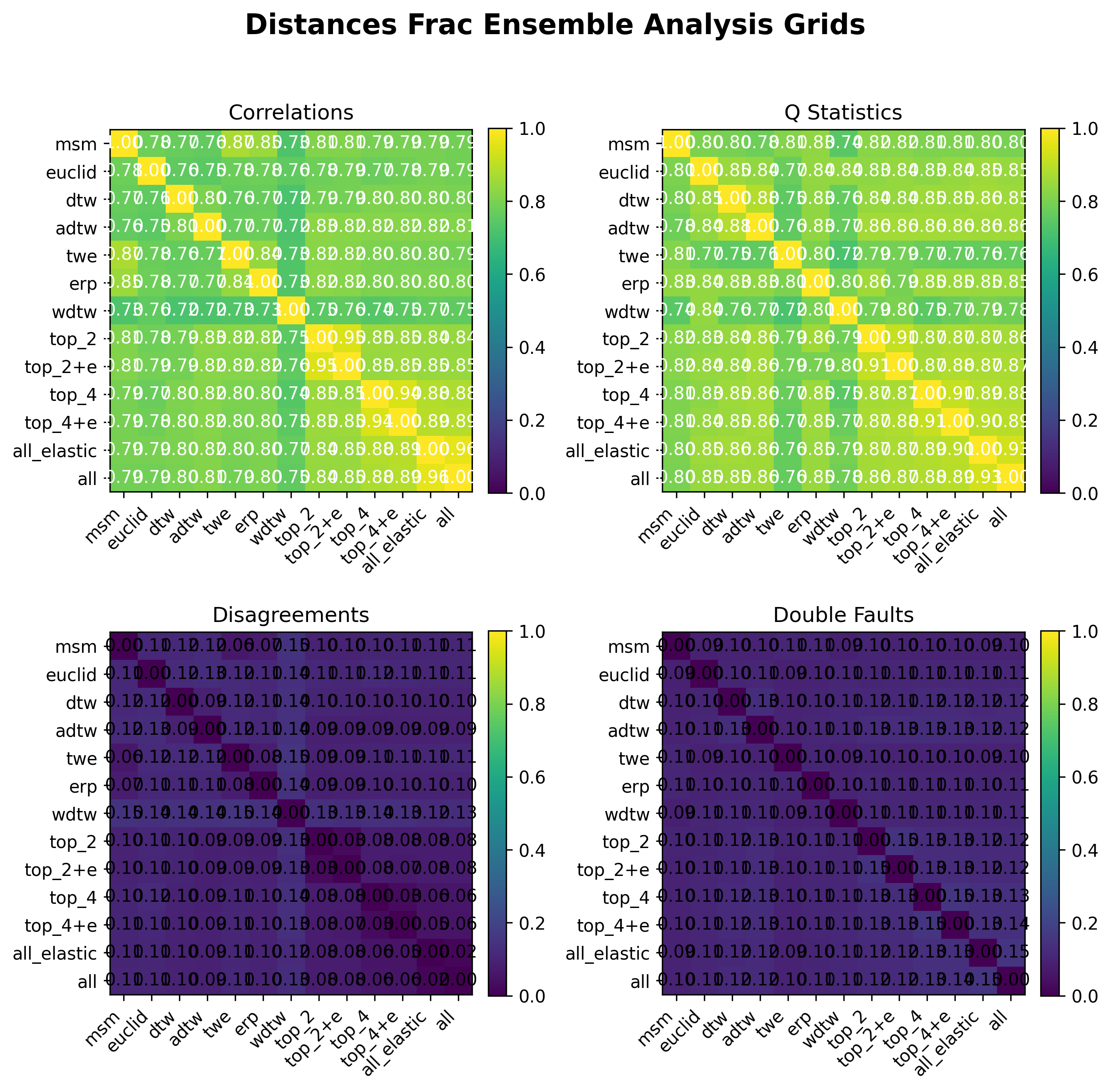}
	
	\caption{Ensemble Statistics for Tested Ensembles.}
	\label{fig:ensemble statistics}
\end{figure}

All Ensembles have higher correlations with each other than the single distance classifiers. The most correlated any ensemble is with any single classifier is $0.80$, which is also the least correlated any ensemble is to any other ensemble. Simultaneously, Ensemble correlations can range as high as $0.90$, which suggests that the ensembles are more similar to each other than any of their components. 

The relative under performance of ensembling multiple distance measures with a naive strategy as opposed to simply using the best distances suggests that hyperparameter tuning may be required to achieve high ensemble performance with measures. The simple concatenated ensembles tested here did not gain significant advantage from diverse distance measures on this dataset. Since more complex ensembling such as stacking and weighted voting require additional computational work for validation, we will not consider it further in this paper.

Instead, we will adopt MSM as the default elastic distance metric for this paper. Although the TWE metric has outranked MSM or tied MSM in average rank on all tests, none of those have been statistically significant. However, the computational requirements for TWE are much larger than MSM, with the MSM transformation taking $63\%$ on average, of TWE's transformation time. 

Furthermore the large computational difference between the Euclidean distance and all other distances suggests that the Euclidean distance should still be considered separately for computationally constrained applications. It also performs relatively well, outranking WDTW in all tests and DTW in some. In light of this, we can consider two distances, one elastic and one inelastic, as default SPROCKET recommendations.

\subsection{Kernel Scaling}
SPROCKET is a convolutional transform and other convolutional transforms are sensitive to the $K$ hyperparameter that determines the number of kernels used. We therefore examine SPROCKET's sensitivity to $K$ in a limited test on the 33 dataset subset before performing a large-scale evaluation on the UCR/UEA benchmark.

To test this, we created MSM classifiers where $K$ is set to every power of two between $8$ and $4096$ and tested them on the small subset of UCR that we previously used.

\begin{figure}[htbp]
	\centering
	
	\includegraphics[width=1\textwidth]{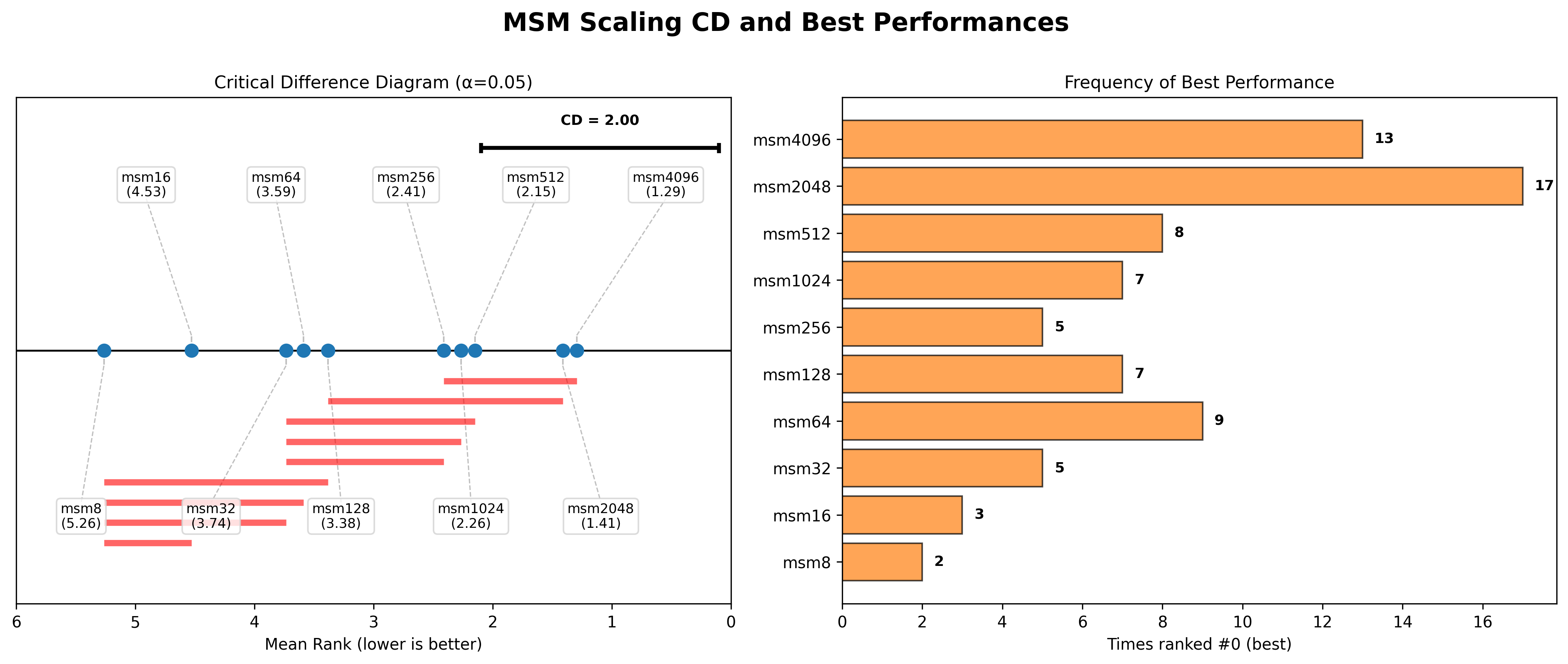}
	
	\caption{Relative Performance of Power of Two Kernel Counts for SPROCKET. Note that msm8 has 8 kernels, msm16 has 16, etc.}
	\label{fig:msm_scaling_cd}
\end{figure}

The Critical Difference diagram reveals that the number of kernels does improve average performance, but not fully uniformly. MSM1024 has a higher average rank than MSM512 by $.11$, a small but notable result. The higher kernel counts are also more frequently the best performing classifiers, but again this is not uniform.

The time statistics are far more uniform and show a clear linear scaling pattern for various values of $K$.

\begin{figure}[htbp]
	\centering
	
	\includegraphics[width=1\textwidth]{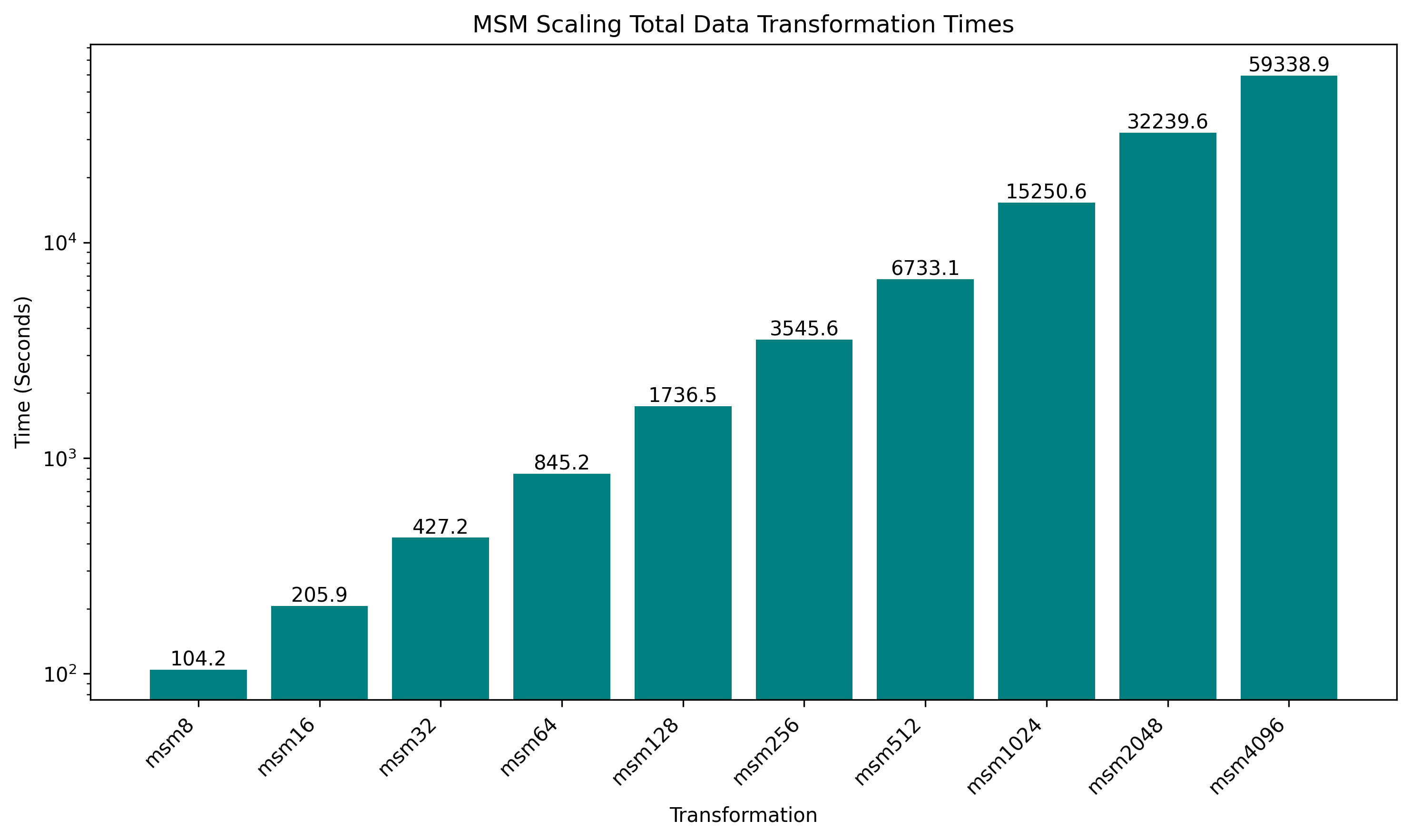}
	
	\caption{Total time to transform data for various MSM kernel counts.}
	\label{fig:msm_scaling_times}
\end{figure}

Finally, correlations and Q-Statistics between different kernel counts of MSM-SPROCKET increase for larger values of K, while disagreements and double faults decrease. These patterns in ensemble statistics suggest diminishing marginal diversity among the independently parameterized MSM-SPROCKET variants. Such diminishing diversity reflects relatively muted gains and expected ensembling accuracy improvements from increased kernel counts as $K$ increases.

\begin{figure}[htbp]
	\centering
	
	\includegraphics[width=1\textwidth]{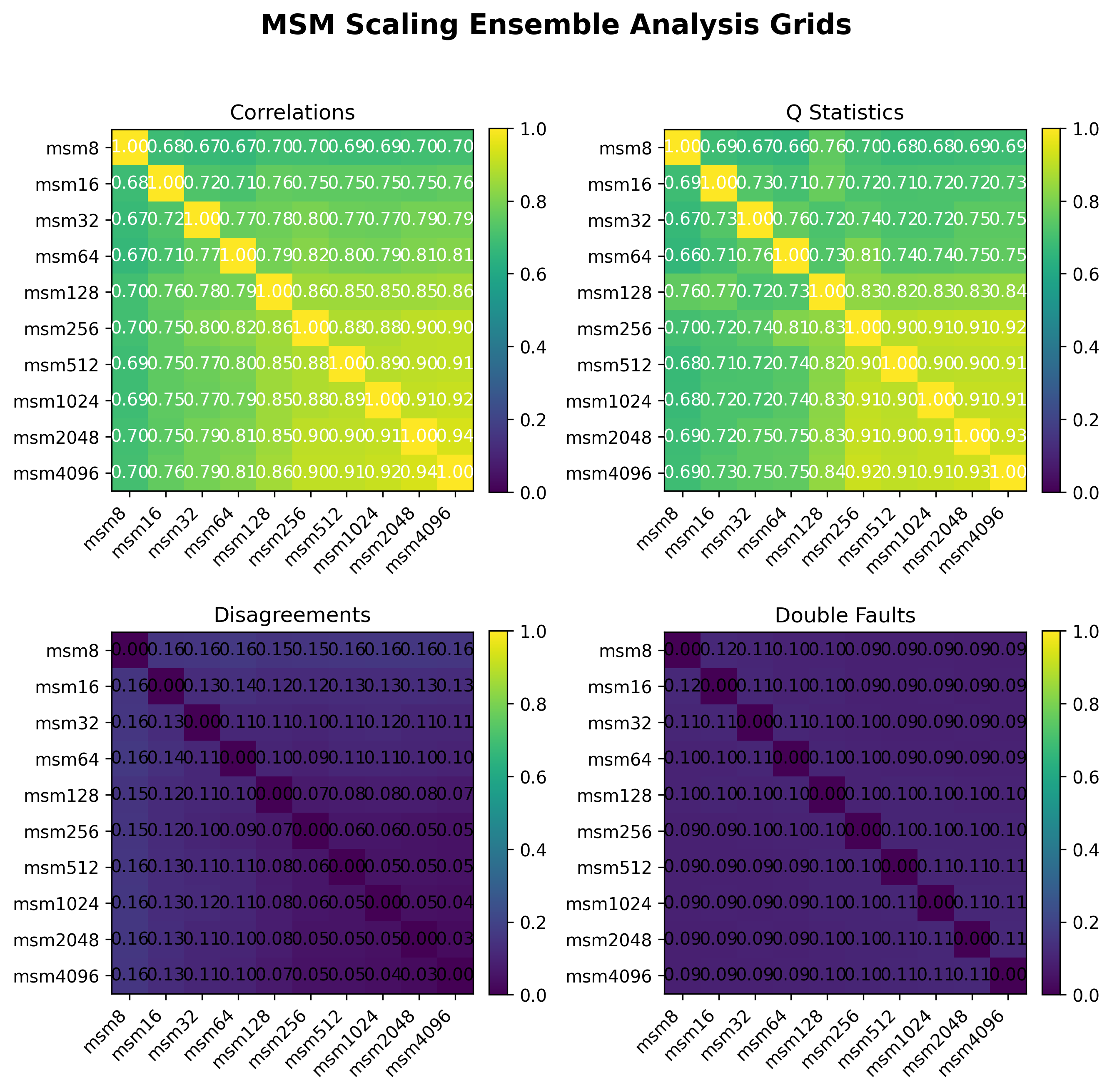}
	
	\caption{Ensemble Statistics for Powers of Two Scaled MSM Sprockets.}
	\label{fig:msm_scaling_ensembles}
\end{figure}

MSM512 is the smallest version of SPROCKET where Q-Statistics plateau at $0.90$ for larger versions of SPROCKET. This suggests that MSM512 is at the point of diminishing returns for larger K. It also has an identical kernel count to HYDRA, another prominent transform that hybridizes convolutional and non-convolutional approaches. This makes $K=512$ a pragmatically attractive default setting for $K$. This small scale testing on 33 datasets cannot definitively establish the optimal K threshold and that threshold could be dataset specific, but 512 kernels balance accuracy and computational requirements within the tested range.

These results indicate that while increasing K generally improves performance, the gains are not linear with increasing values of K, while corresponding increases in computation time are linear. For large-scale evaluation, we therefore adopt K=512 as a balanced default.

\subsection{Large Scale MSM Benchmarking}
We now proceed to a large-scale test of SPROCKET on the UCR Time Series Classification Library. We tested MultiROCKET, HYDRA, SPROCKET, MR-HY (MultiROCKET-HYDRA), MR-SP (MultiROCKET-SPROCKET), HY-SP (HYDRA-SPROCKET), MR-HY-SP (MultiROCKET-HYDRA-SPROCKET), and QUANT\cite{dempster2023quantminimalistintervalmethod}, as a non-convolutional baseline. The MultiROCKET, HYDRA, and SPROCKET ensembles were formed by concatenating their transformed features and using a single RidgeCV classifier, as is standard for the existing MR-HYDRA ensemble. All non-SPROCKET transforms utilized their standard configurations in the Aeon-Toolkit and the tested SPROCKET configuration utilizes MSM distances, 512 kernels, $\lceil \log_4(|X|) \rceil$ random prototypes, and a window parameter of $\lfloor \sqrt{l} \rfloor$, where $l$ is the length of the series.

For this test, we sought to capture a wide selection of datasets from the UCR Classification Benchmark Library. We excluded:
\begin{itemize}
	\item Datasets with unequal length series, which are not compatible with any of the tested classifiers.
	\item Datasets with more than $5000$ training instances. Due to the large number of features produced by MultiROCKET, those datasets would require stochastic gradient descent (SGD) on available hardware instead of closed-form Ridge regression used elsewhere, reducing comparability.
	\item Datasets with length greater than $500$. Since SPROCKET scales superlinearly with length, a maximum length was needed.
	\item Datasets with length under $9$. Since MultiROCKET requires datasets to have a length of at least $9$, this is to maintain comparability  between all classifiers.
\end{itemize}
This resulted in 98 datasets, which are detailed in Appendix A. These datasets were tested 5 times with different random seeds and the standard train-test split provided by the UCR datasets. The results were averaged and presented below.

\begin{figure}[htbp]
	\centering
	
	\includegraphics[width=1\textwidth]{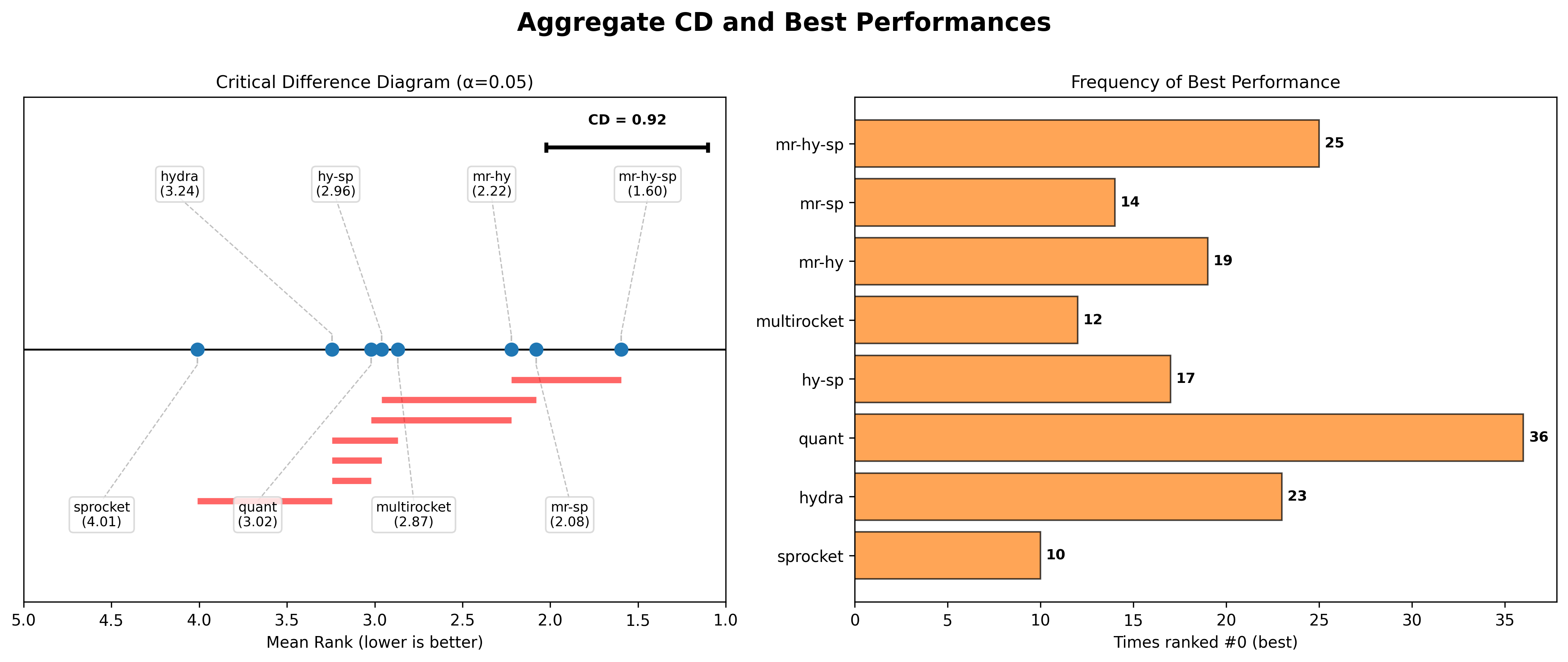}
	
	\caption{Aggregate CD and Best Performances}
	\label{fig:aggregated_acc_CD}
\end{figure}

These results establish that, although SPROCKET is the worst performing in individual algorithm, it ensembles at least as well as HYDRA and MultiROCKET for convolutional algorithms. It also achieves the best rank on $10$ of the $98$ datasets, showing that it is not strictly dominated as an individual algorithm. More importantly, the SPROCKET ensembles perform very well. The MR-HY-SP ensemble outranks all other algorithms in this test and the MR-SP ensemble outranks MR-HY, despite the fact that HYDRA outranks SPROCKET individually. The large number of individual best ranks for QUANT are likely attributable to it being the only nonlinear algorithm in this sample, but are still notable.

Detailed head-to-head comparison scatterplots of each algorithm are not included here because of their high number. To examine them in detail, please see Appendix B.

We can also examine the ensembling statistics for the tested algorithms.

\begin{figure}[htbp]
	\centering
	
	\includegraphics[width=1\textwidth]{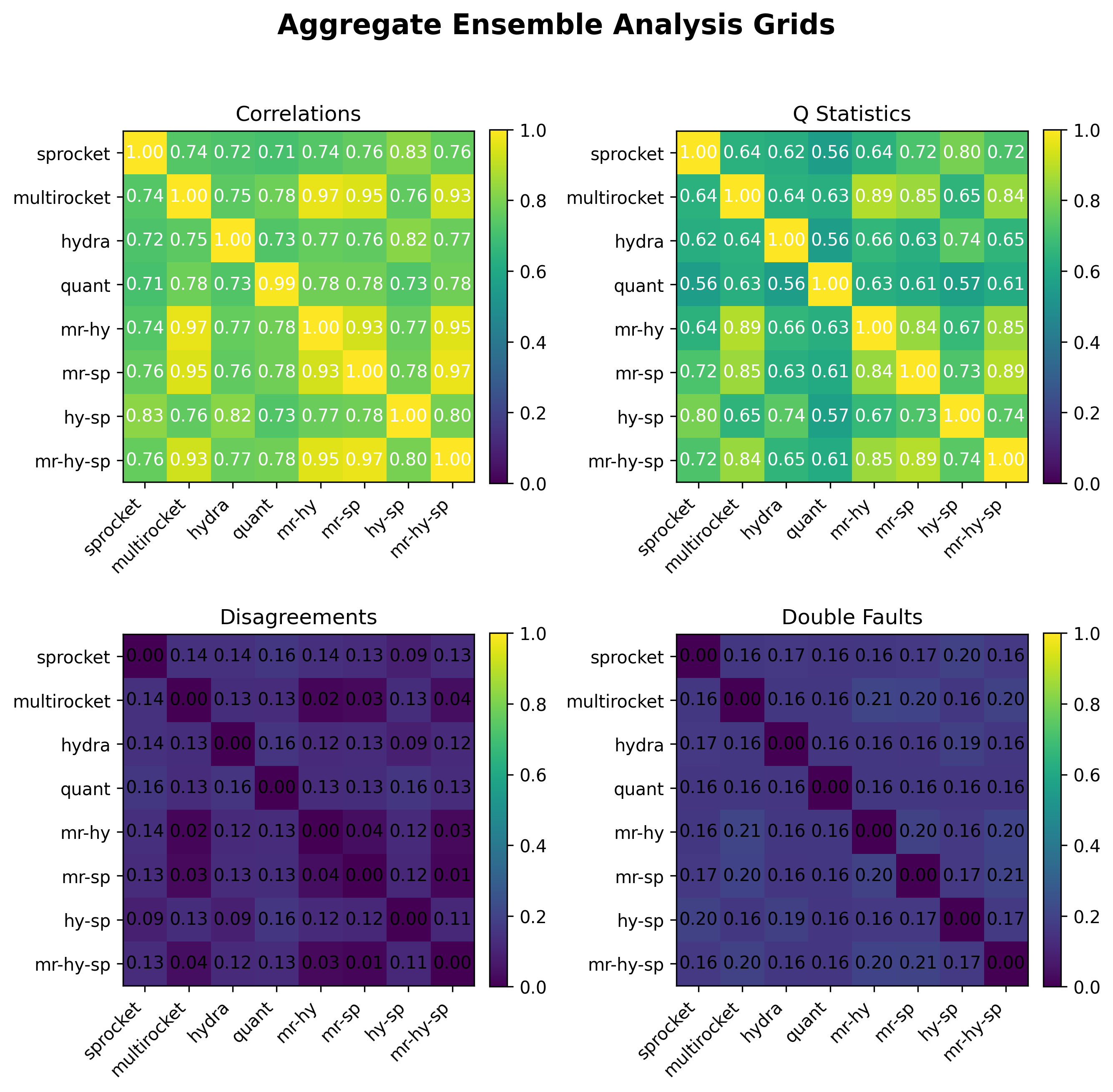}
	
	\caption{Ensembling Statistics for Large Scale MSM Test}
	\label{fig:aggregated_ensemble}
\end{figure}

The Pairwise Q-Statistics for MultiROCKET-SPROCKET, MultiROCKET-HYDRA, and HYDRA-SPROCKET are $0.64$, $0.64$, and $0.62$ respectively. This relationship may explain why the pairwise ensembles all perform similarly. All the ensembles containing MultiROCKET are more correlated with MultiROCKET than any of their other subcomponents, suggesting that MultiROCKET is dominant subcomponent in the ensembles. Finally, QUANT retains similar ensembling statistics to each convolutional method, which suggests that the predictions produced by the covolutional methods are still meaningfully different than at least one non-convolutional method.

We can now turn to the running times. By taking the average transformation time of the SPROCKET transforms against the predicted value from our theoretical analysis in Section 4, we can confirm that the theoretical results are somewhat representative of our observed results, which is to be expected due variance caused by randomization in the SPROCKET algorithm and hardware implementation details.

\begin{figure}[htbp]
	\centering
	
	\includegraphics[width=1\textwidth]{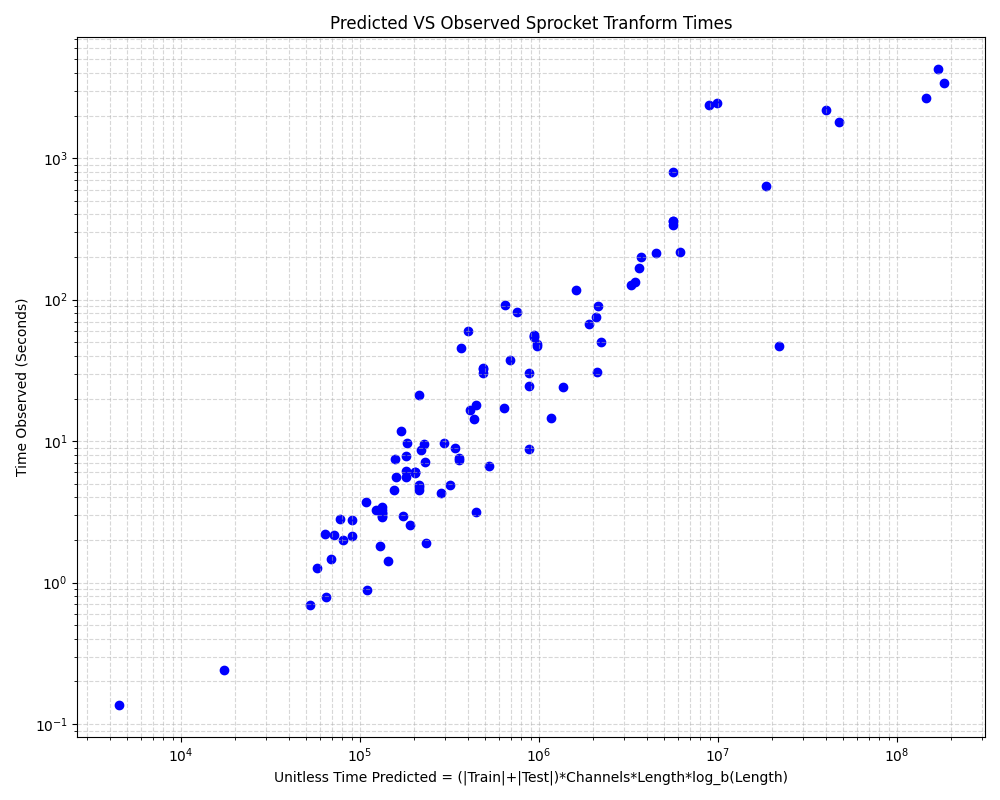}
	
	\caption{SPROCKET Predictions Vs Observed Times}
	\label{fig:predicted_vs_observed_times}
\end{figure}

Finally, we can see that the SPROCKET transform with the MSM distance is substantially more computationally expensive than the other compared transforms.

\begin{table}[h!]
	\centering
	\begin{tabularx}{\textwidth}{||c|>{\RaggedRight\arraybackslash}X|}
		\hline
		\textbf{Algorithm} & \textbf{Total Time (Seconds)} \\
		\hline\hline
		SPROCKET & 24353 \\
		\hline
		MultiROCKET & 364 \\
		\hline
		HYDRA & 193 \\
		\hline
		QUANT & 1373 \\
		\hline
		MR-HY & 590 \\
		\hline
		MR-SP & 24657 \\
		\hline
		HY-SP & 24415 \\
		\hline
		MR-HY-SP & 24796 \\
		\hline
	\end{tabularx}
	\caption{Total Time Taken For all Tests in Large Scale MSM Testing.}
\end{table}

\subsection{Large Scale Euclidean Benchmarking}
Because of the large computational difference between MSM distance SPROCKET and the other convolutional algorithms, we repeated the large scale benchmark with the Euclidean distance measure. The experimental setup was otherwise identical. All non-SPROCKET transforms utilized their standard configurations in the Aeon-Toolkit and the tested SPROCKET configuration utilizes Euclidean distances, 512 kernels, $\lceil \log_4(|X|) \rceil $ random prototypes, and no window parameter, since it is not required for the Euclidean distance.

The experiment was repeated 5 times with different random seeds and the average results are presented below.

\begin{figure}[htbp]
	\centering
	
	\includegraphics[width=1\textwidth]{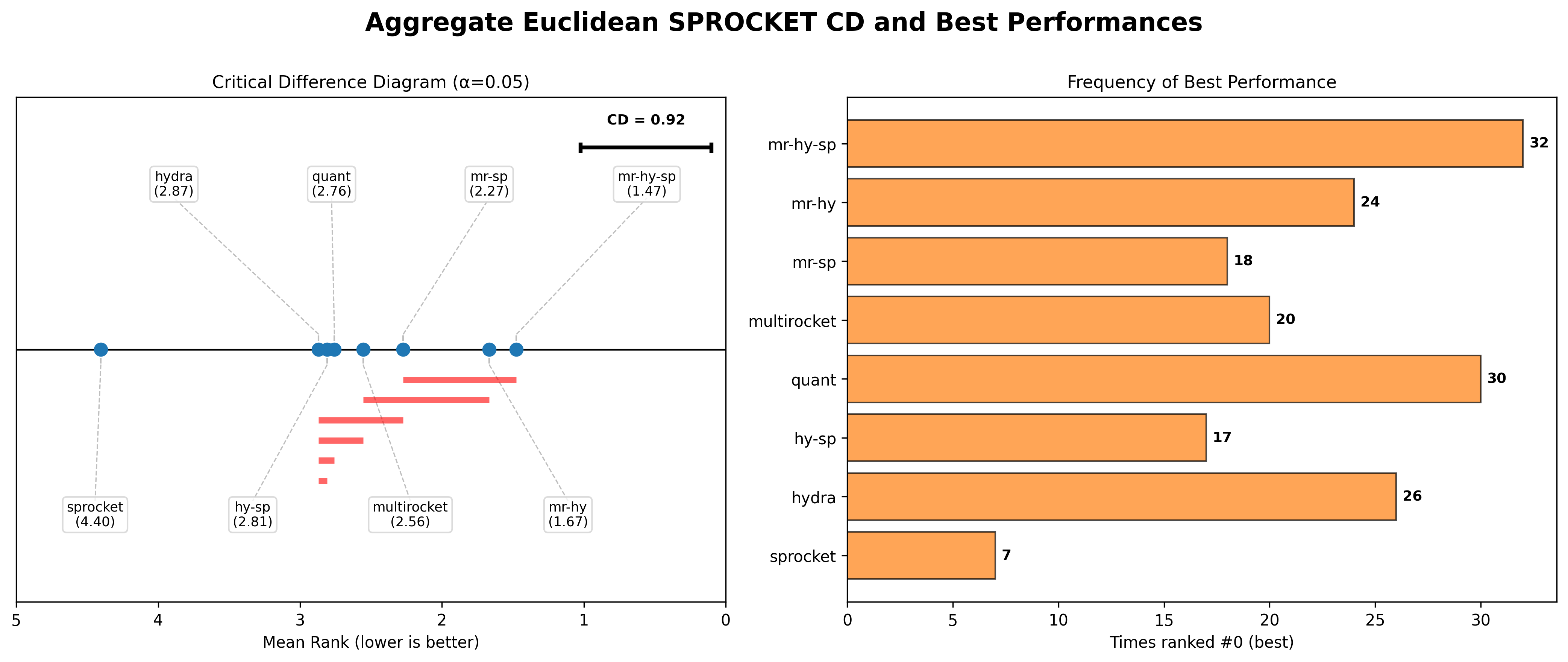}
	
	\caption{Euclidean Aggregate CD and Best Performances}
	\label{fig:euclidean_aggregated_acc_CD}
\end{figure}

Although the ranking is less clear cut than in the MSM case, a similar ranking is present in for the Euclidean SPROCKET. The MR-HY-SP ensemble holds the best possible rank, though the difference between it and the other algorithms has decreased. Interestingly, the MR-HY-SP algorithm has more overall best finishes in this configuration, even though the sum of its ranks has increased. Furthermore, the HY-SP ensemble is now ranked under QUANT, but the gap is not large.

Detailed head-to-head comparison scatterplots of each algorithm are again in Appendix B.

\begin{figure}[htbp]
	\centering
	
	\includegraphics[width=1\textwidth]{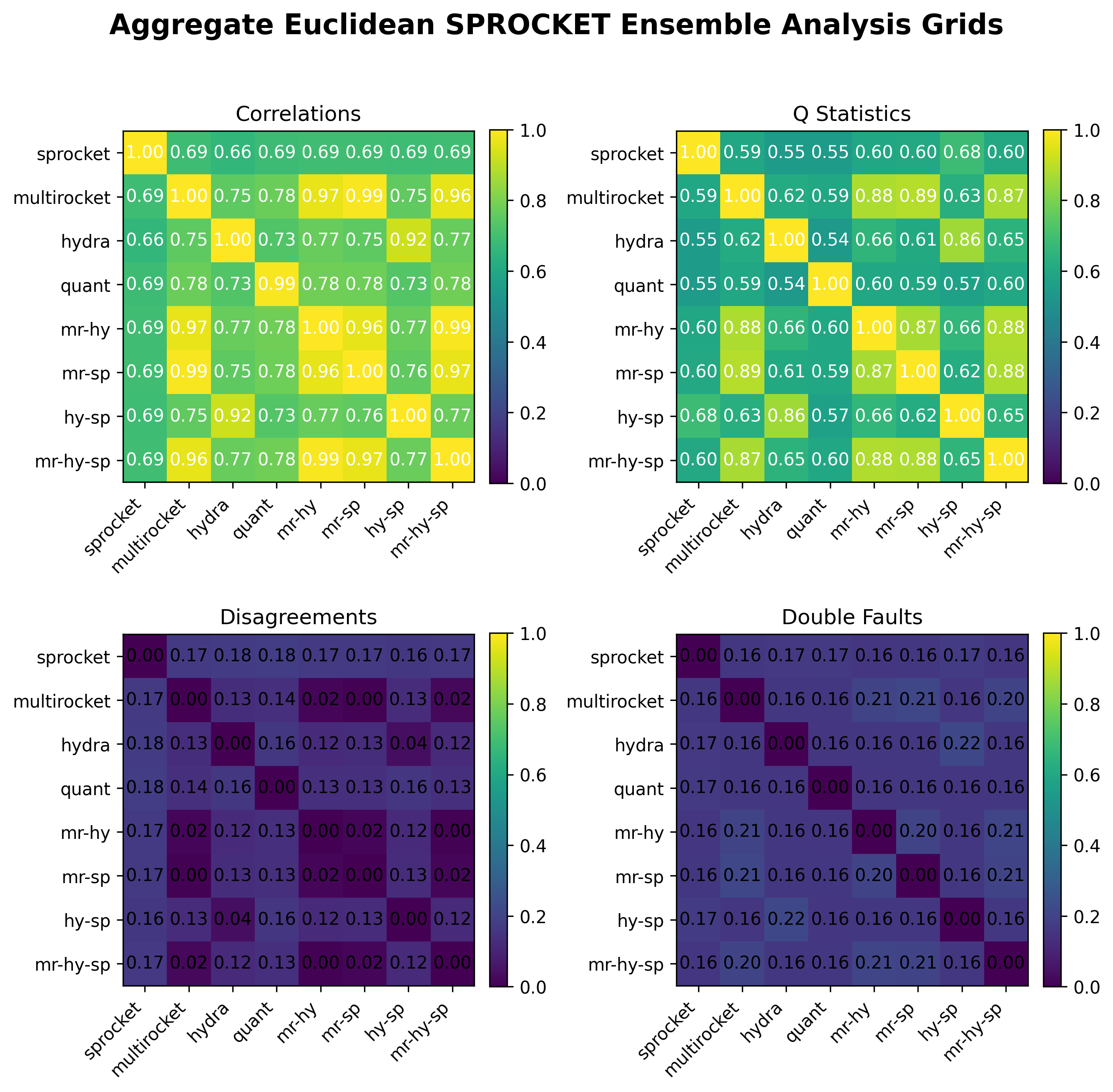}
	
	\caption{Ensembling Statistics for Large Scale Euclidean Test}
	\label{fig:euclidean_aggregated_ensemble}
\end{figure}

The Pairwise Q-Statistics for MultiROCKET-SPROCKET, MultiROCKET-HYDRA, have decreased to  $0.59$ and $0.55$, which may explain why the accuracy decrease is relatively small under ensembling. Euclidean SPROCKET is slightly more independent than MSM SPROCKET from existing convolutional algorithms, though it is less accurate.

We can also repeat our test of predicted vs observed running times. Note that our prediction function has changed, since the Euclidean distance contains no windowing parameter and is instead strictly linear on length.

\begin{figure}[htbp]
	\centering
	
	\includegraphics[width=1\textwidth]{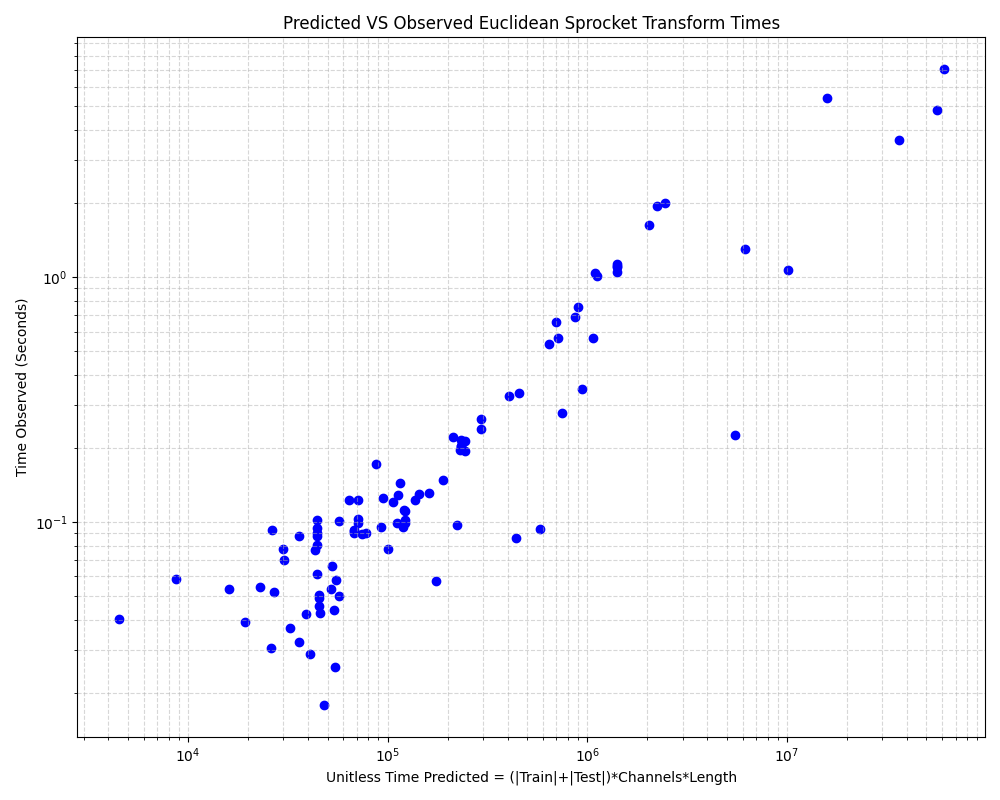}
	
	\caption{Euclidean SPROCKET Predictions Vs Observed Times}
	\label{fig:euclidean_observed_vs_predicted}
\end{figure}

Finally, we can see that the Euclidean SPROCKET transform is much more computationally efficient than the MSM version, with the lowest total computation time of any tested algorithms.

\begin{table}[h!]
	\centering
	\begin{tabularx}{\textwidth}{||c|>{\RaggedRight\arraybackslash}X|}
		\hline
		\textbf{Algorithm} & \textbf{Total Time (Seconds)} \\
		\hline\hline
		SPROCKET & 169 \\
		\hline
		MultiROCKET & 437 \\
		\hline
		HYDRA & 212 \\
		\hline
		QUANT & 1325 \\
		\hline
		MR-HY & 778 \\
		\hline
		MR-SP & 668 \\
		\hline
		HY-SP & 254 \\
		\hline
		MR-HY-SP & 819 \\
		\hline
	\end{tabularx}
	\caption{Total Time Taken For all Tests in Large Scale Euclidean Testing.}
\end{table}

\section{Future Work}
This paper introduces SPROCKET as a proof of concept for prototype driven integration of distance based time series methods into ROCKET. This initial step leaves several promising avenues for continued research and optimization.
\subsection{Improvements on SPROCKET Design}
The design process employed in this research is limited. It did not examine many design decisions, including but not limited to:
\begin{itemize}
	\item Non-random prototype selection.
	\item Weighted or adaptive distance measure ensembles.
	\item The $\lceil \log_4(|X|) \rceil$ prototype scaling heuristic.
	\item Pre and post convolution transformations (normalization, differencing, scaling, etc).
	\item Other distance measures beyond the seven evaluated.
\end{itemize}
All of these could be potentially improved and represent avenues for further research.
\subsection{Improvements from Convolutional Transforms}
This SPROCKET baseline uses the same convolutional architecture as baseline ROCKET, and does not include subsequent refinements since ROCKET's introduction. Convolutional strategy developments may be of interest and include:
\begin{itemize}
	\item First order difference transforms before kernel transformations, as seen in the MultiROCKET.
	\item Constrained kernel randomization to improve computation, feature extraction efficiency, or consistency as seen in limited kernel randomization in Mini and MultiROCKET.
	\item Pruning the kernels after initial training. Several strategies exist to perform this in the ROCKET framework, including Detach-ROCKET \cite{uribarri2024detachrocketsequentialfeatureselection} and S-ROCKET \cite{salehinejad2022srocketselectiverandomconvolution}. The computational improvements from reducing distance calculations will be higher for SPROCKET than for other ROCKET variants, due to the higher cost of distance calculations per kernel.
\end{itemize}
\subsection{Improvements from Distance Classifiers}
SPROCKET is a hybrid Convolutional-Distance transformation and may benefit from the strategies employed by distance based time series classifiers. These include:
\begin{itemize}
	\item Derivative transformations before or after kernel transformations, as seen in the Proximity Forest 2.0. This transformation is similar to the transformation use in MultiROCKET.
	\item Combining HYDRA and Minkowski distances as a non-elastic distance measure, as in the Proximity Forest 2.0
	\item Early abandonment and other computational optimizations of distance calculations to reduce runtime without sacrificing accuracy.
\end{itemize}
Any of these could improve both the accuracy and computational efficiency of SPROCKET.
\subsection{Comprehensive Empirical Characterization}

The empirical evaluation in this paper was constrained by computational resources and focused on datasets with moderate length $(\leq 500)$ and size $(\leq 5000$ instances. Several important questions remain:

\begin{itemize}
	\item Performance on long time series (length >500), potentially using approximation methods or sampling strategies for distance computation.
	\item Scalability to large datasets (>5,000 instances) and testing on the Monash MONSTER benchmark, utilizing SGD or other scaling methods.
	\item Sensitivity analysis for key hyperparameters beyond kernel count K, including prototype count, window parameters, and distance-specific settings.
	\item Evaluation on extrinsic regression and forecasting tasks to extend beyond simple classification to broader time series learning problems.
	\item Performance on domain-specific benchmarks (e.g., audio classification, medical time series, financial datasets).
	\item Stability and variance across different random initializations, particularly for datasets where SPROCKET shows high variability.
	\item Theoretical analysis of SPROCKET's approximation properties and relationship to nearest-neighbor classification and prototype methods.
\end{itemize}

\subsection{Implementation and Accessibility}

The reference implementation of SPROCKET was adapted from existing transformations in Aeon and inherits SciKit-Learn compatibility. It is available at \url{https://github.com/username76543/sprocket_public} This base level of integration could be built on in several ways:
\begin{itemize}
	\item Optimized implementations leveraging additional parallelization and vectorization, beyond the numba optimizations included in the reference code.
	\item GPU acceleration for distance computations, particularly for large-scale applications.
	\item Integration into popular time series classification libraries such as aeon, sktime, and tslearn. The provided reference code was built utilizing aeon standards, but support should be provided for other popular libraries.
	\item Automated hyperparameter selection tools to reduce manual tuning burden.
	\item Comprehensive documentation and usage examples for different application domains.
\end{itemize}

While SPROCKET is not individually the top performing convolutional transform, its contribution to convolutional ensemble learning is demonstrably valuable and represents an improvement on existing methods. Further developments could build on this contribution to offer new fast and accurate convolutional methods. We invite the time series research community to explore any of these proposed research directions or others not listed.

\section{Conclusion}
This paper demonstrates that distance-based prototype features can be effectively integrated into convolutional time series classification frameworks. SPROCKET, while not individually state-of-the-art, substantially improves ensemble performance: the MR-HY-SP ensemble achieves the best average rank across 98 UCR benchmark datasets, surpassing previous convolutional ensembles. This improvement validates that complementary predictions from different algorithmic families enhance classification accuracy beyond what individual methods achieve.

The computational cost of elastic distance calculations presents a tradeoff: MSM-based SPROCKET requires significantly more computation than other convolutional methods and Euclidean SPROCKET reduces this by orders of magnitude. Both significantly improve convolutional ensembles, but selecting the appropriate measure will depend on application constraints. The simple baseline design employed here—random prototypes, logarithmic scaling prototype numbers, standard distances—leaves substantial room for optimization through improved prototype selection, kernel pruning, and computational optimization.

By bridging convolutional and distance-based approaches, SPROCKET opens a productive research direction at the intersection of two major time series classification paradigms.

\section{Citations} 
\bibliography{references}


%

\begin{appendices}
\section{Datasets}
Two sets of datasets were used for the empirical evaluation. All were taken from the UCR Time Series Classification Archive and the UEA Multivariate Time Series Archive. For all tests, they were imported in their standard Train/Test splits via the Aeon Toolkit.

The first, smaller set of datasets were used in the parameter evaluation studies. These were selected to cover a variety of train and test sizes, dimensions, number of classes, and application areas, while allowing for faster iteration through a reduced sampling space.

\begin{table}[H]
	\centering
	\begin{tabularx}{\textwidth}{||>{\RaggedRight\arraybackslash}X|c|c|c|c||}
		\hline
		\textbf{Dataset Name} & \textbf{Train Size} & \textbf{Test Size} & \textbf{Length} & \textbf{Classes} \\ [0.5ex]
		\hline\hline
		Coffee & 28 & 28 & 286 & 2 \\ 
		\hline
		Beef & 30 & 30 & 470 & 5 \\
		\hline
		FaceFour & 24 & 88 & 350 & 4 \\
		\hline
		Lightning7 & 70 & 73 & 319 & 7 \\
		\hline
		Wine & 57 & 54 & 234 & 2 \\
		\hline
		Meat & 60 & 60 & 448 & 3 \\
		\hline
		Ham & 109 & 105 & 431 & 2 \\
		\hline
		ECG200 & 100 & 100 & 96 & 2 \\
		\hline
		ItalyPowerDemand & 67 & 1029 & 24 & 2 \\
		\hline
		MoteStrain & 20 & 1252 & 84 & 2 \\
		\hline
		SonyAIBORobotSurface1 & 20 & 601 & 70 & 2 \\
		\hline
		SonyAIBORobotSurface2 & 27 & 953 & 65 & 2 \\
		\hline
		TwoLeadECG & 23 & 1139 & 82 & 2 \\
		\hline
		GunPoint & 50 & 150 & 150 & 2 \\
		\hline
		Trace & 100 & 100 & 275 & 4 \\
		\hline
		CBF & 30 & 900 & 128 & 3 \\
		\hline
		SyntheticControl & 300 & 300 & 60 & 60 \\
		\hline
		DiatomSizeReduction & 16 & 306 & 345 & 4 \\
		\hline
		SwedishLeaf & 500 & 625 & 128 & 15 \\
		\hline
		Wafer & 1000 & 6164 & 152 & 2 \\
		\hline
		ECG5000 & 500 & 4500 & 140 & 5 \\
		\hline
		FaceAll & 560 & 1690 & 131 & 14 \\
		\hline
		FacesUCR & 200 & 2050 & 131 & 14 \\
		\hline
		WordSynonyms & 267 & 638 & 270 & 25 \\
		\hline
		Adiac & 390 & 391 & 176 & 37 \\
		\hline
		ChlorineConcentration & 467 & 3840 & 166 & 3 \\
		\hline
		Yoga & 300 & 3000 & 426 & 2 \\
		\hline
		MedicalImages & 381 & 760 & 99 & 10 \\
		\hline
		FordA & 3601 & 1320 & 500 & 2 \\
		\hline
		FordB & 3636 & 810 & 500 & 2 \\
		\hline
		ElectricDevices & 8926 & 7711 & 96 & 7 \\
		\hline
		SpokenArabicDigits & 6599 & 2199 & 93 & 10 \\
		\hline
		Handwriting & 150 & 850 & 152 & 26 \\
		\hline
	\end{tabularx}
	\caption{Datasets used for Parameter Selection.}
\end{table}

The larger set of datasets was also selected from the UCR archive, but chosen more broadly to include 98 of the 132 datasets available. It includes all datasets included in the previous test except for Electric Devices and SpokenArabicDigits, which would have required a switch to SGD from standard Ridge Regression to fit in available hardware memory. This switch may have affected comparability and necessitated excluding the datasets. The additional datasets included are listed below.

\begin{center}[H]
\small
\begin{longtable}{||c c c c c||}
	\hline
	Dataset Name & Train Size & Test Size & Length & Classes \\[0.5ex]
	\hline\hline
	\endfirsthead
	
	\hline
	Dataset Name & Train Size & Test Size & Length & Classes \\[0.5ex]
	\hline\hline
	\endhead
	
	\hline
	\multicolumn{5}{r}{Continued on next page}\\
	\endfoot
	
	\hline
	\endlastfoot
	DuckDuckGeese & 60 & 40 & 270 & 5 \\ \hline
	PEMS-SF & 267 & 173 & 144 & 7 \\ \hline
	MindReading & 727 & 653 & 200 & 5 \\ \hline
	PhonemeSpectra & 3315 & 3353 & 217 & 39 \\ \hline
	ShapeletSim & 20 & 180 & 500 & 2 \\ \hline
	EMOPain & 1093 & 50 & 180 & 3 \\ \hline
	SmoothSubspace & 150 & 150 & 15 & 3 \\ \hline
	MelbournePedestrian & 1194 & 2439 & 24 & 10 \\ \hline
	Chinatown & 20 & 345 & 24 & 2 \\ \hline
	JapaneseVowels & 270 & 370 & 29 & 9 \\ \hline
	RacketSports & 151 & 152 & 30 & 4 \\ \hline
	LSST & 2459 & 2466 & 36 & 14 \\ \hline
	Libras & 180 & 180 & 45 & 15 \\ \hline
	FingerMovements & 316 & 100 & 50 & 2 \\ \hline
	NATOPS & 180 & 180 & 51 & 6 \\ \hline
	SharePriceIncrease & 965 & 965 & 60 & 2 \\ \hline
	ERing & 30 & 270 & 65 & 6 \\ \hline
	PhalangesOutlinesCorrect & 1800 & 858 & 80 & 2 \\ \hline
	ProximalPhalanxOutlineCorrect & 600 & 291 & 80 & 2 \\ \hline
	MiddlePhalanxOutlineCorrect & 600 & 291 & 80 & 2 \\ \hline
	DistalPhalanxOutlineCorrect & 600 & 276 & 80 & 2 \\ \hline
	ProximalPhalanxTW & 400 & 139 & 80 & 2 \\ \hline
	ProximalPhalanxOutlineAgeGroup & 400 & 205 & 80 & 3 \\ \hline
	MiddlePhalanxOutlineAgeGroup & 400 & 154 & 80 & 3 \\ \hline
	MiddlePhalanxTW & 399 & 154 & 80 & 6 \\ \hline
	DistalPhalanxTW & 400 & 139 & 80 & 6 \\ \hline
	DistalPhalanxOutlineAgeGroup & 400 & 139 & 80 & 3 \\ \hline
	BasicMotions & 40 & 40 & 100 & 4 \\ \hline
	TwoPatterns & 1000 & 4000 & 128 & 4 \\ \hline
	BME & 30 & 150 & 128 & 3 \\ \hline
	EyesOpenShut & 56 & 42 & 128 & 2 \\ \hline
	ECGFiveDays & 23 & 861 & 136 & 2 \\ \hline
	ArticularyWordRecognition & 275 & 300 & 144 & 25 \\ \hline
	PowerCons & 180 & 180 & 144 & 2 \\ \hline
	Plane & 105 & 105 & 144 & 7 \\ \hline
	GunPointOldVersusYoung & 135 & 316 & 150 & 2 \\ \hline
	GunPointMaleVersusFemale & 135 & 316 & 150 & 2 \\ \hline
	GunPointAgeSpan & 135 & 316 & 150 & 2 \\ \hline
	UMD & 36 & 144 & 150 & 3 \\ \hline
	Epilepsy2 & 80 & 11420 & 178 & 2 \\ \hline
	Colposcopy & 100 & 100 & 180 & 6 \\ \hline
	Fungi & 18 & 186 & 201 & 18 \\ \hline
	WalkingSittingStanding & 7352 & 2947 & 206 & 6 \\ \hline
	Epilepsy & 137 & 138 & 207 & 4 \\ \hline
	Strawberry & 613 & 370 & 235 & 2 \\ \hline
	ElectricDeviceDetection & 623 & 3767 & 256 & 2 \\ \hline
	FiftyWords & 450 & 455 & 270 & 50 \\ \hline
	ToeSegmentation1 & 40 & 228 & 277 & 2 \\ \hline
	DodgerLoopWeekend & 20 & 138 & 288 & 2 \\ \hline
	DodgerLoopGame & 20 & 138 & 288 & 2 \\ \hline
	DodgerLoopDay & 78 & 80 & 288 & 7 \\ \hline
	CricketZ & 390 & 390 & 300 & 12 \\ \hline
	CricketY & 390 & 390 & 300 & 12 \\ \hline
	CricketX & 390 & 390 & 300 & 12 \\ \hline
	FreezerRegularTrain & 150 & 2850 & 301 & 2 \\ \hline
	FreezerSmallTrain & 28 & 2850 & 301 & 2 \\ \hline
	UWaveGestureLibraryZ & 896 & 3582 & 315 & 8 \\ \hline
	UWaveGestureLibraryY & 896 & 3582 & 315 & 8 \\ \hline
	UWaveGestureLibraryX & 896 & 3582 & 315 & 8 \\ \hline
	UWaveGestureLibrary & 2238 & 3582 & 315 & 8 \\ \hline
	ToeSegmentation2 & 36 & 130 & 343 & 2 \\ \hline
	GestureMidAirD3 & 208 & 130 & 360 & 26 \\ \hline
	GestureMidAirD2 & 208 & 130 & 360 & 26 \\ \hline
	GestureMidAirD1 & 208 & 130 & 360 & 26 \\ \hline
	Symbols & 25 & 995 & 398 & 6 \\ \hline
	HandMovementDirection & 160 & 74 & 400 & 4 \\ \hline
	Heartbeat & 204 & 205 & 405 & 2 \\ \hline
	OSULeaf & 200 & 242 & 427 & 6 \\ \hline
	Fish & 175 & 175 & 463 & 7 \\ \hline
	
\end{longtable}
\end{center}

\section{Pairwise Accuracy Comparisons}

\subsection{Randomized Prototype and Distance Metric Parameter Testing}

\begin{figure}[H]
	\centering
	
	\includegraphics[width=1\textwidth]{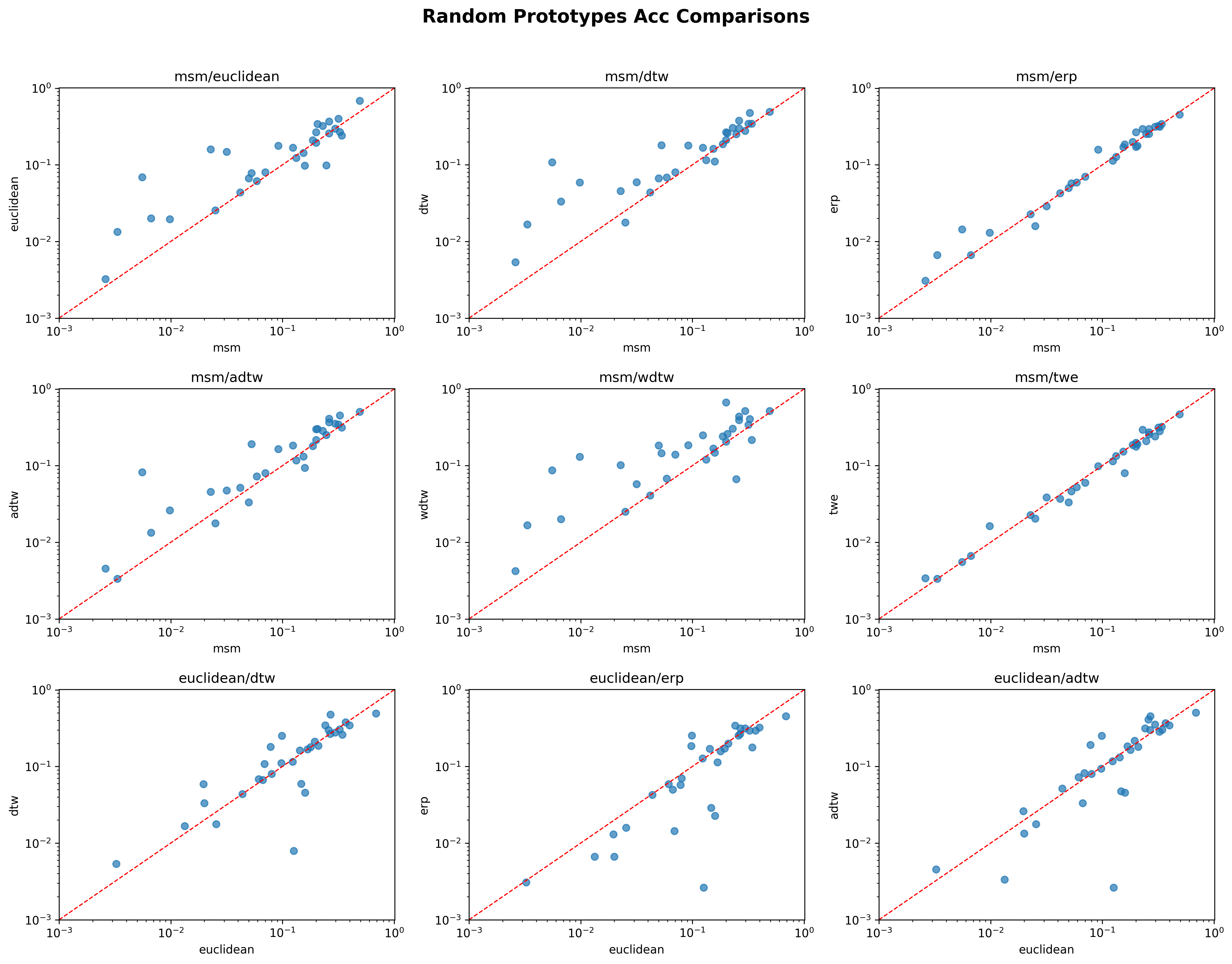}
	
	\caption{Random Prototype Distance Metrics Comparison 1}
	\label{fig:randomized prototype distance comparisons 1}
\end{figure}

\begin{figure}[H]
	\centering
	
	\includegraphics[width=1\textwidth]{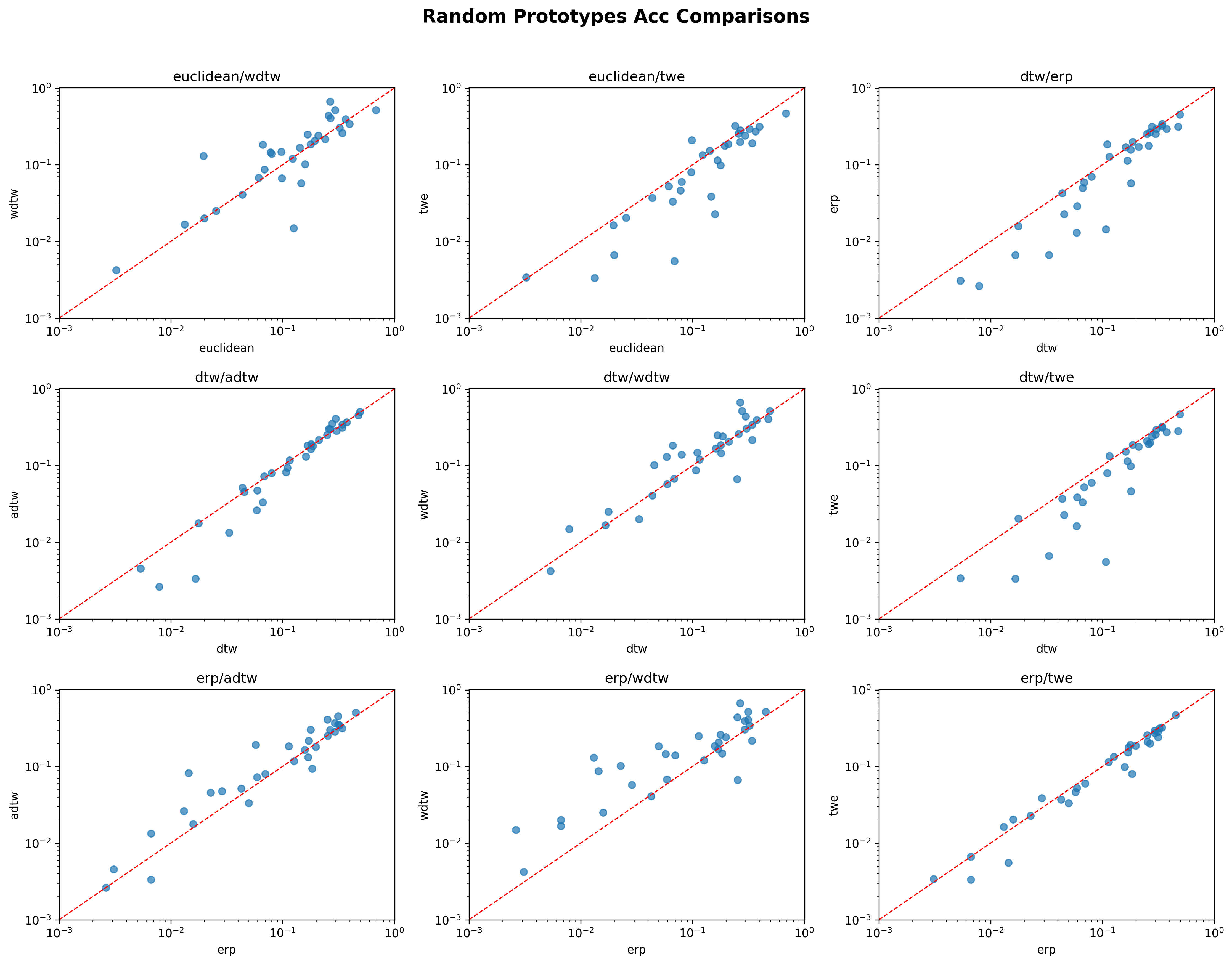}
	
	\caption{Random Prototype Distance Metrics Comparison 2}
	\label{fig:randomized prototype distance comparisons 2}
\end{figure}

\begin{figure}[H]
	\centering
	
	\includegraphics[width=1\textwidth]{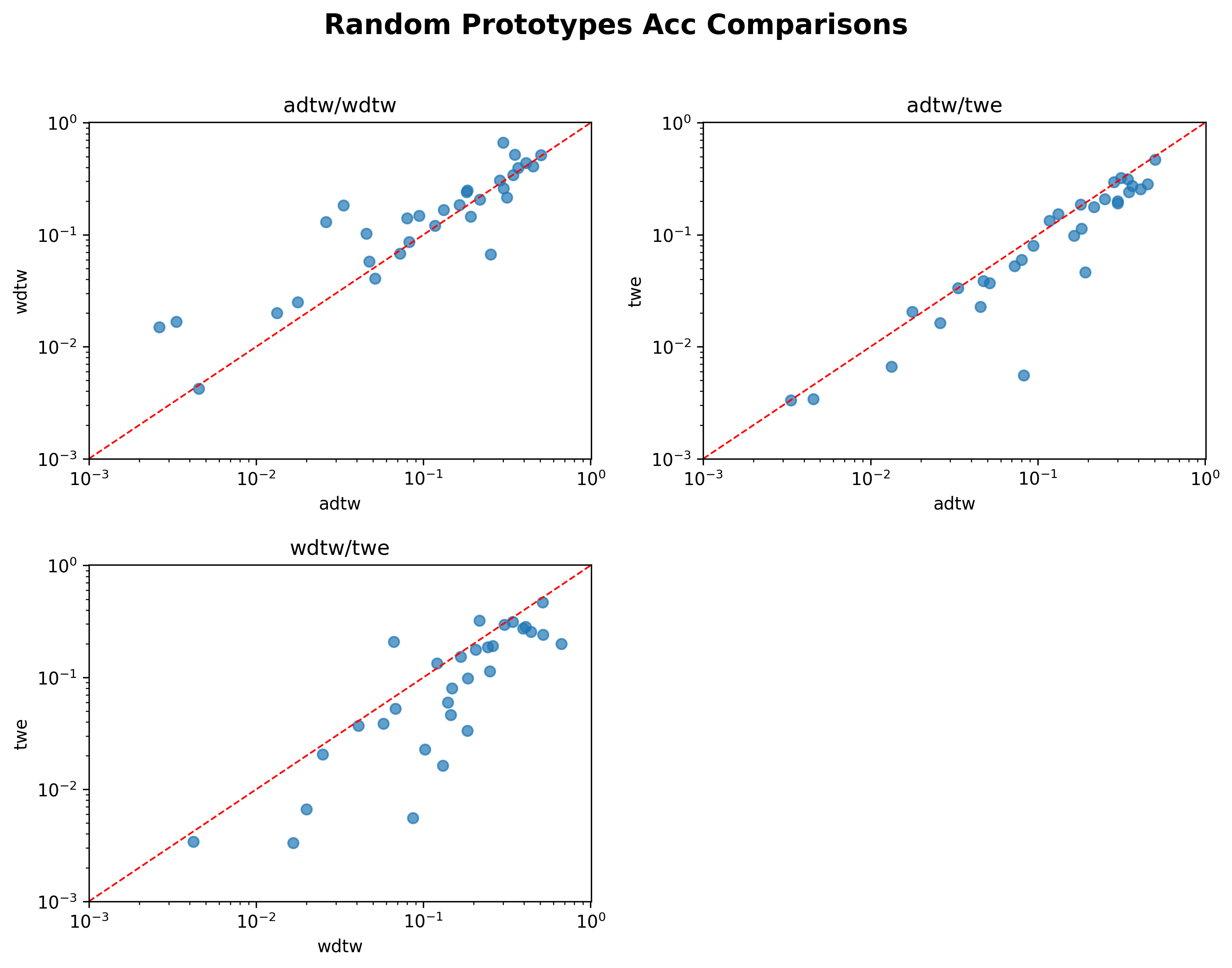}
	
	\caption{Random Prototype Distance Metrics Comparison 3}
	\label{fig:randomized prototype distance comparisons 3}
\end{figure}

\subsection{Stratified Random Prototype and Distance Metric Parameter Testing}

\begin{figure}[H]
	\centering
	
	\includegraphics[width=1\textwidth]{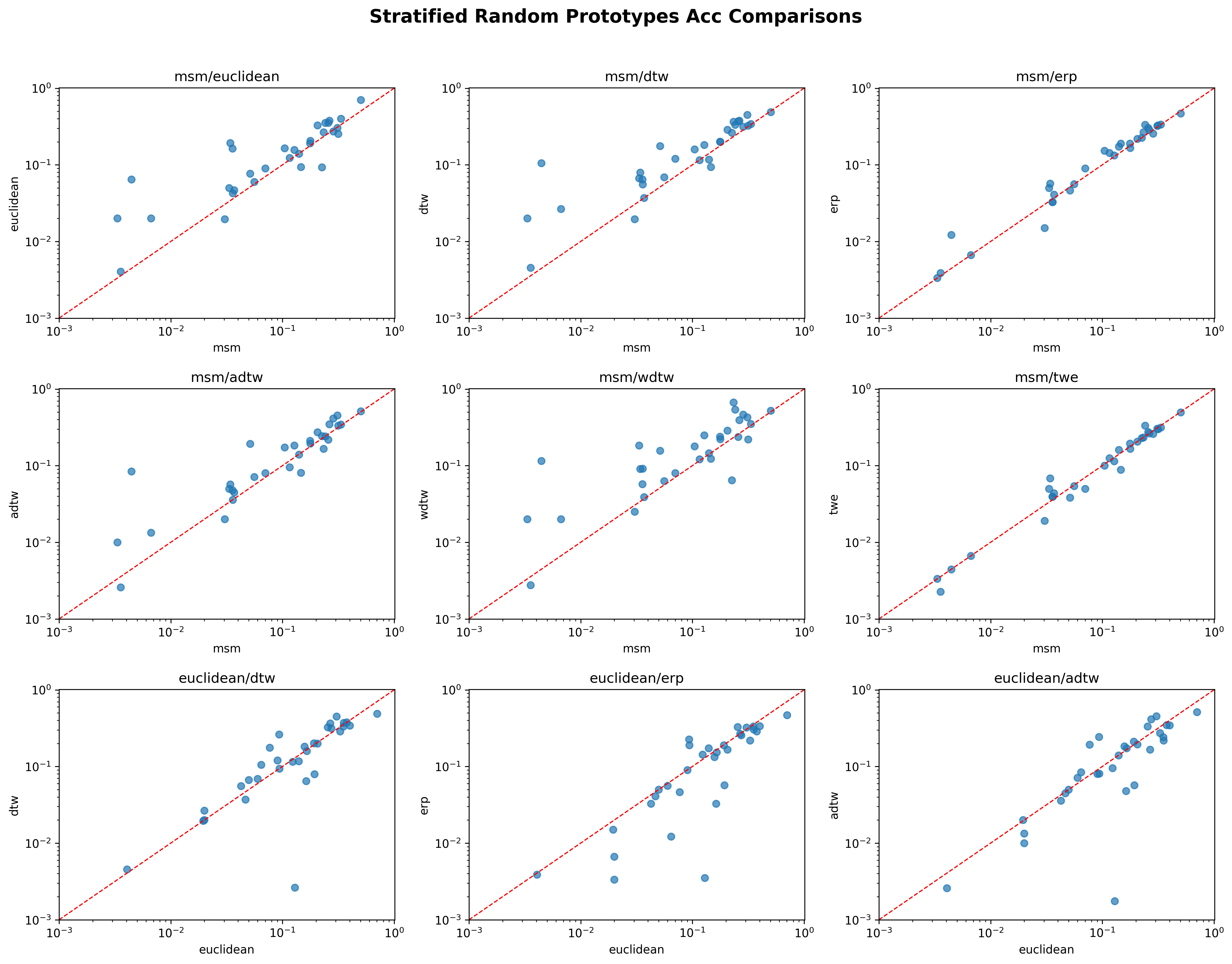}
	
	\caption{Stratified Random Prototype Distance Metrics Comparison 1}
	\label{fig:stratified random prototype distance comparisons 1}
\end{figure}

\begin{figure}[H]
	\centering
	
	\includegraphics[width=1\textwidth]{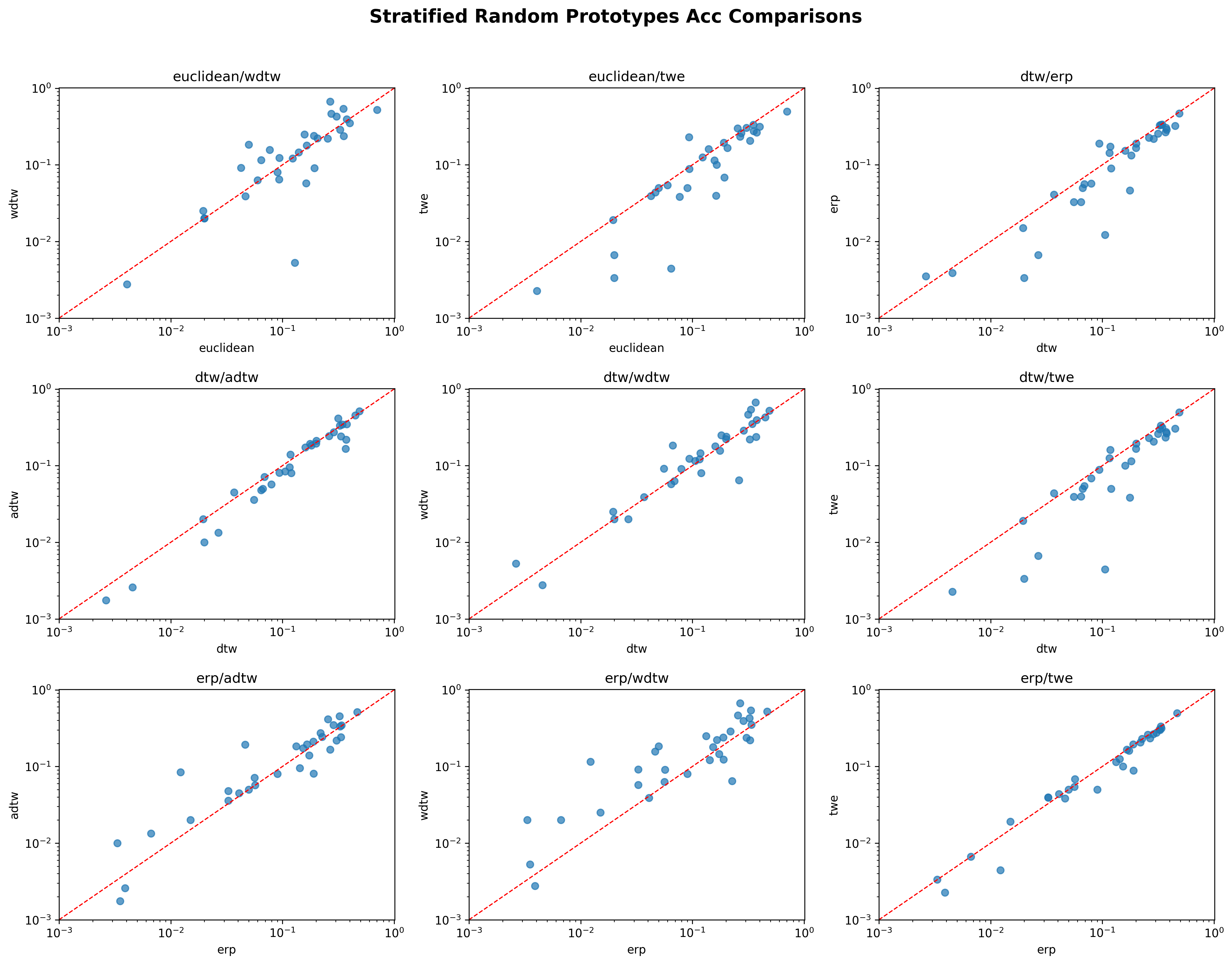}
	
	\caption{Stratified Random Prototype Distance Metrics Comparison 2}
	\label{fig:stratified random prototype distance comparisons 2}
\end{figure}

\begin{figure}[H]
	\centering
	
	\includegraphics[width=1\textwidth]{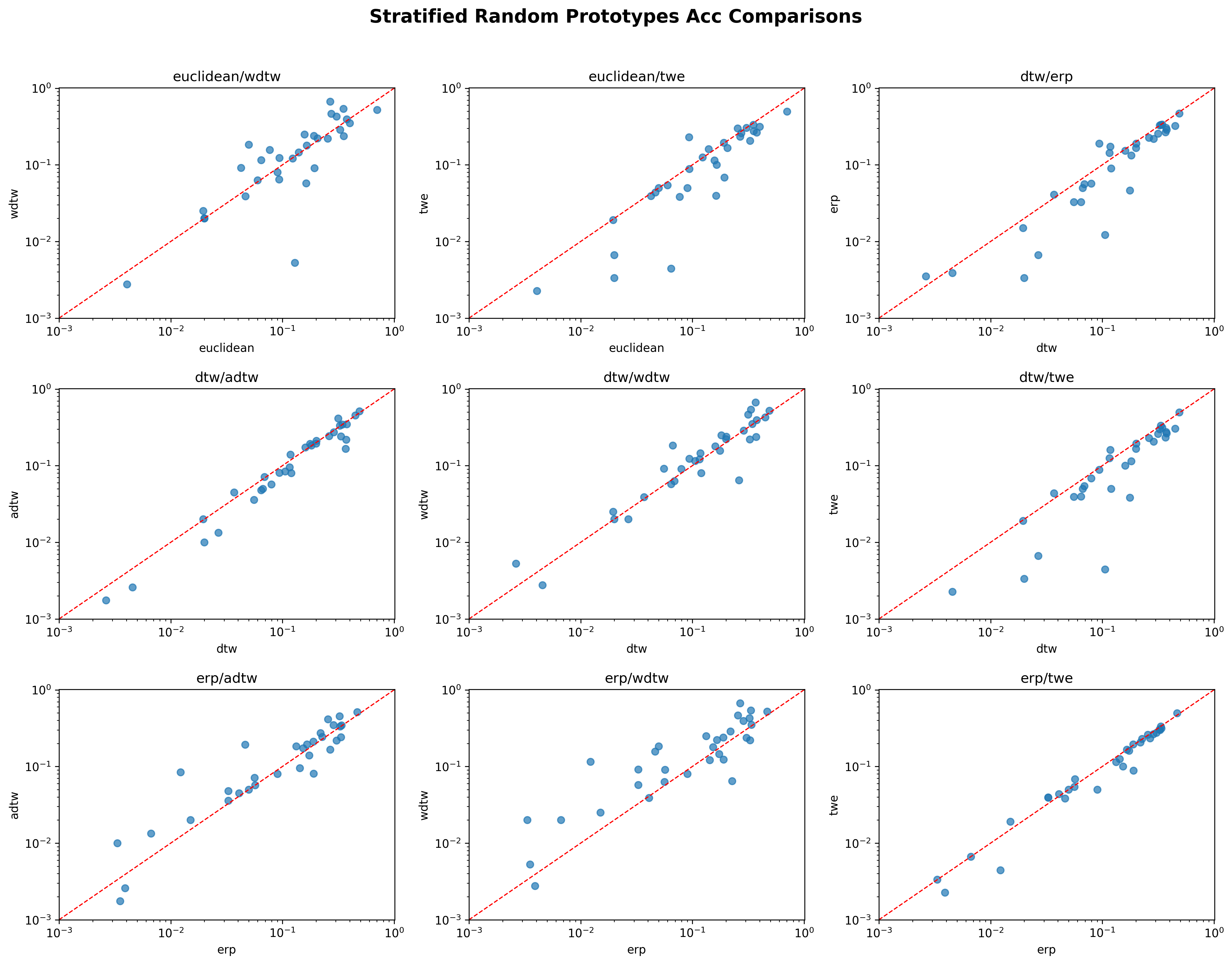}
	
	\caption{Stratified Random Prototype Distance Metrics Comparison 3}
	\label{fig:stratified random prototype distance comparisons 3}
\end{figure}

\subsection{Simple Ensemble Comparisons}

\begin{figure}[H]
	\centering
	
	\includegraphics[width=1\textwidth]{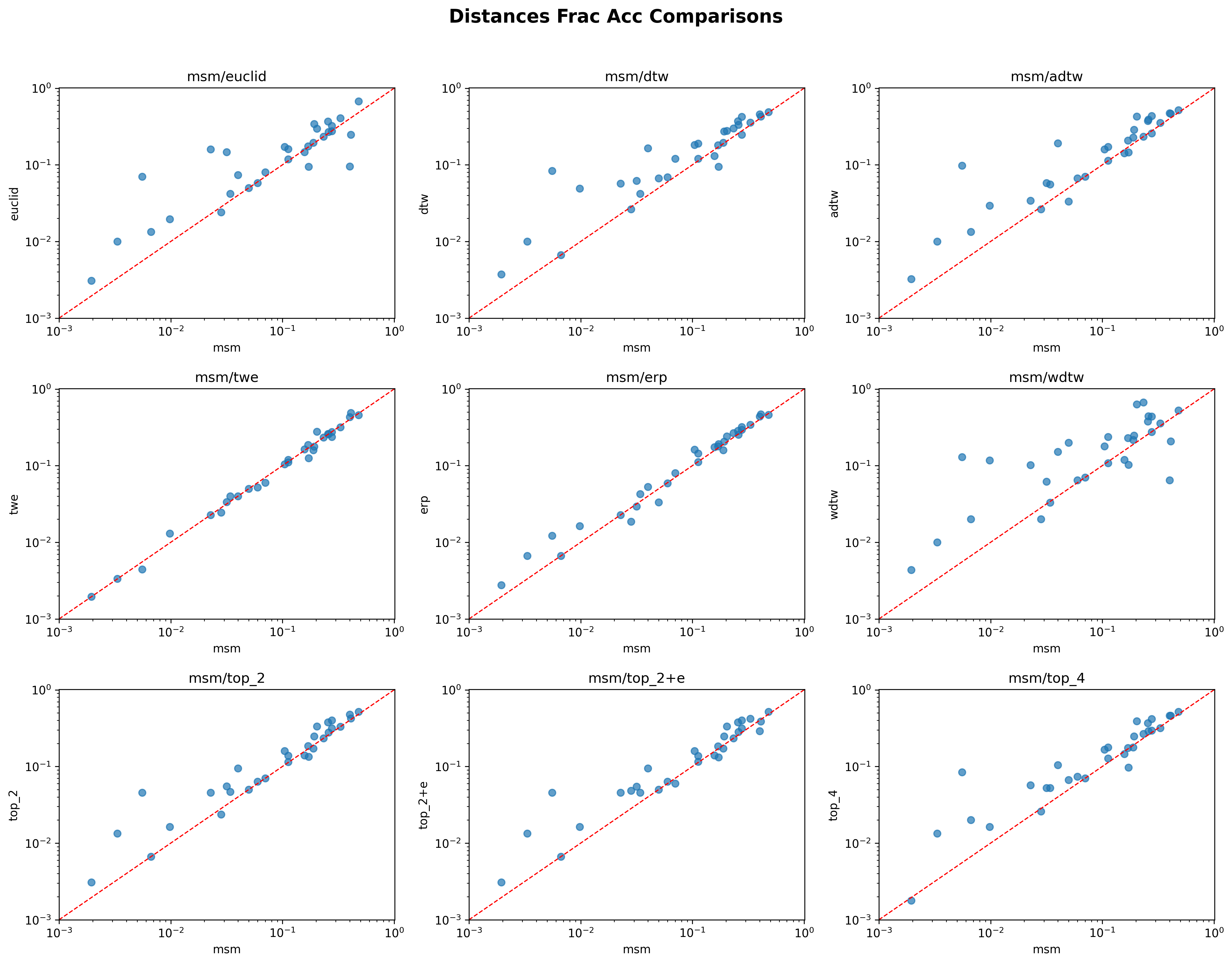}
	
	\caption{Distance Ensemble Comparison 1}
	\label{fig:distance frac comparison 1}
\end{figure}

\begin{figure}[H]
	\centering
	
	\includegraphics[width=1\textwidth]{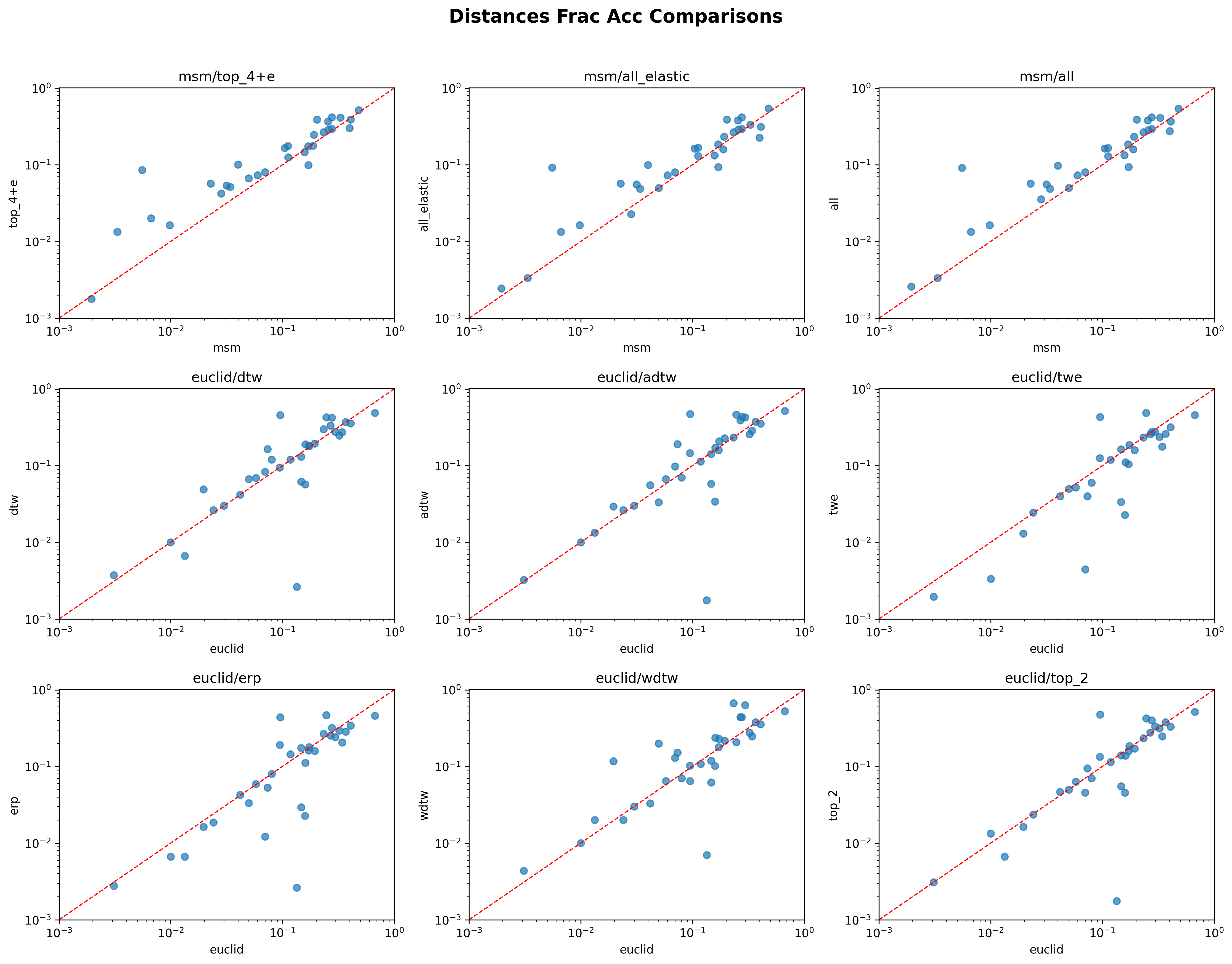}
	
	\caption{Distance Ensemble Comparison 2}
	\label{fig:distance frac comparison 2}
\end{figure}

\begin{figure}[H]
	\centering
	
	\includegraphics[width=1\textwidth]{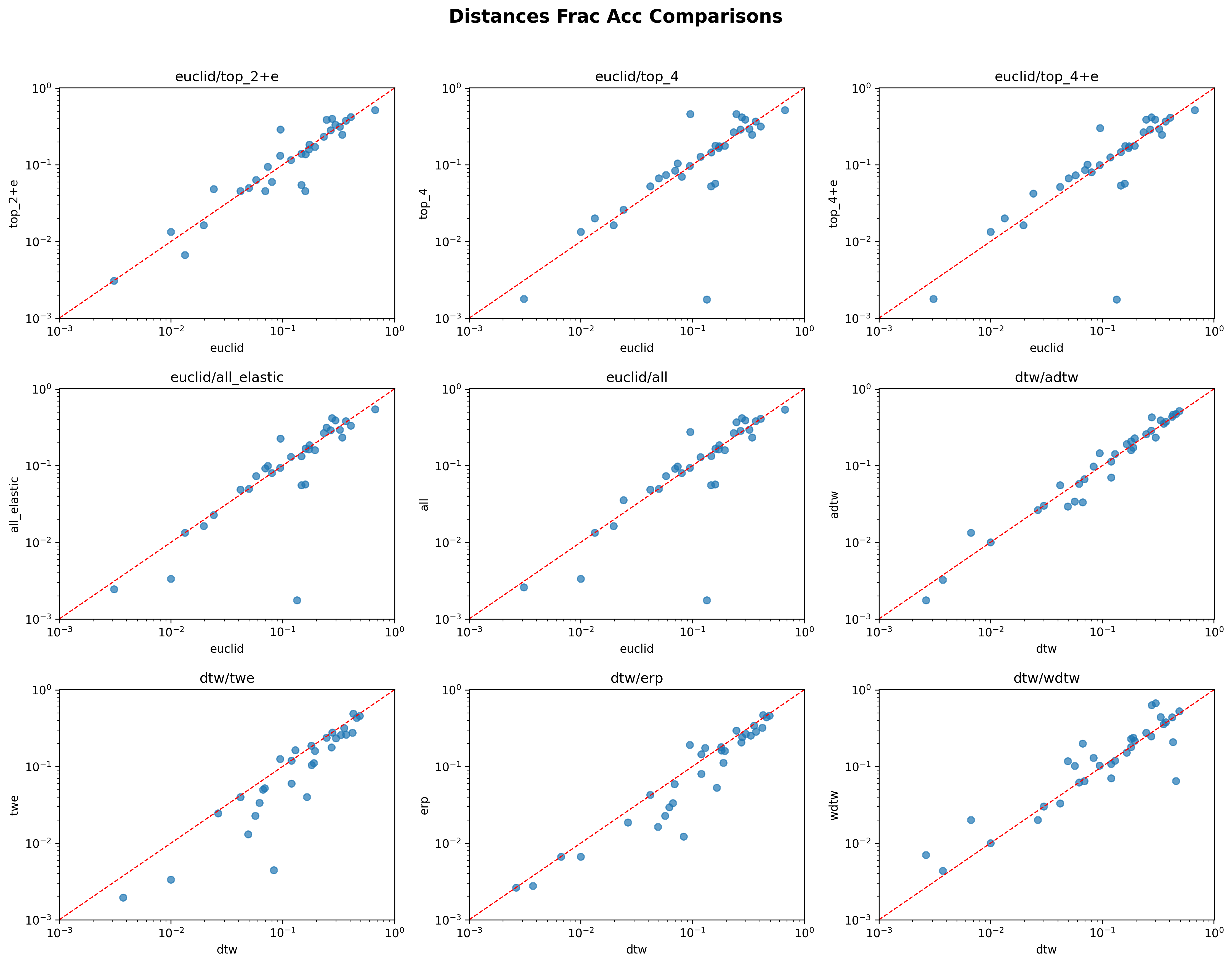}
	
	\caption{Distance Ensemble Comparison 3}
	\label{fig:distance frac comparison 3}
\end{figure}

\begin{figure}[H]
	\centering
	
	\includegraphics[width=1\textwidth]{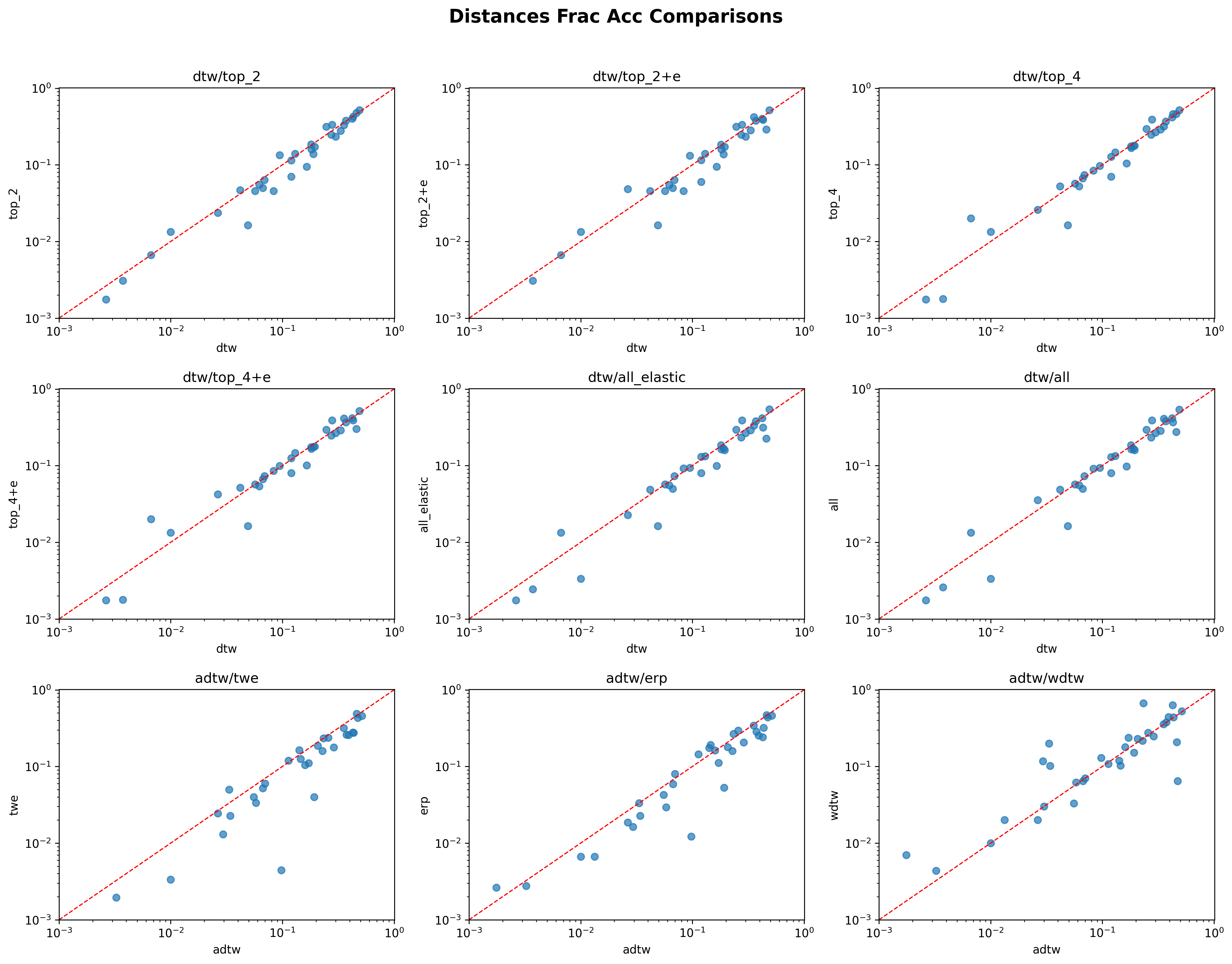}
	
	\caption{Distance Ensemble Comparison 4}
	\label{fig:distance frac comparison 4}
\end{figure}

\begin{figure}[H]
	\centering
	
	\includegraphics[width=1\textwidth]{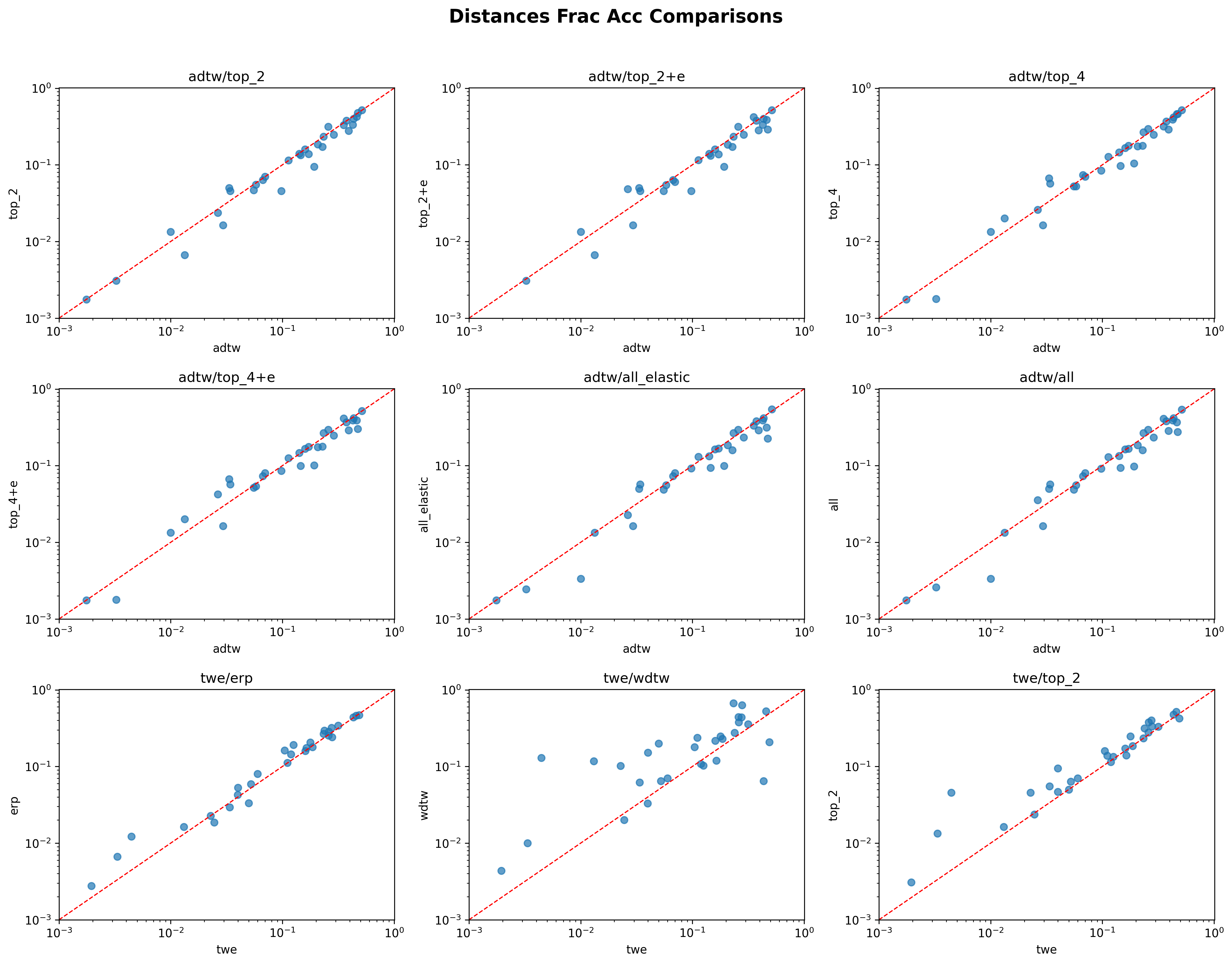}
	
	\caption{Distance Ensemble Comparison 5}
	\label{fig:distance frac comparison 5}
\end{figure}

\begin{figure}[H]
	\centering
	
	\includegraphics[width=1\textwidth]{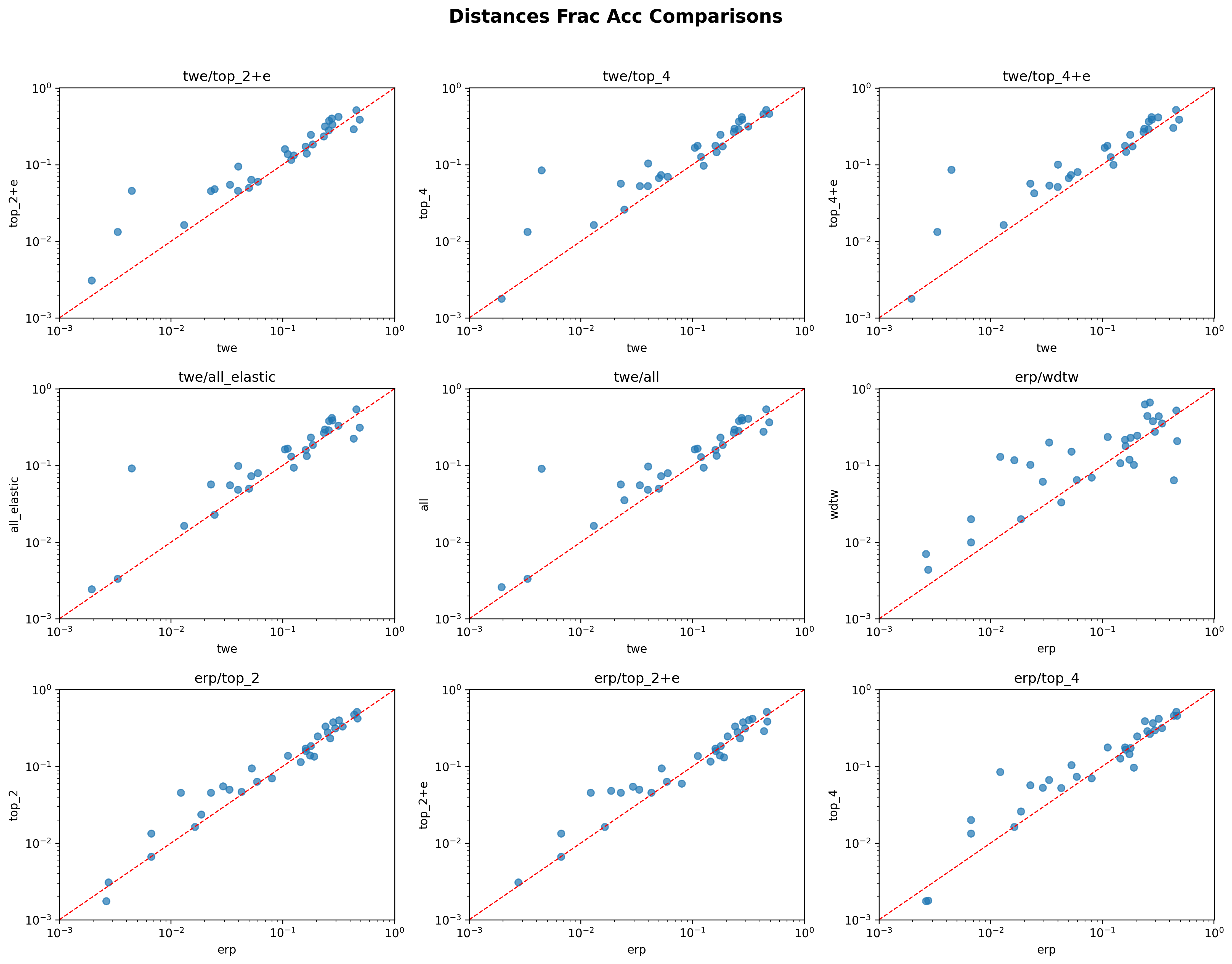}
	
	\caption{Distance Ensemble Comparison 6}
	\label{fig:distance frac comparison 6}
\end{figure}

\begin{figure}[H]
	\centering
	
	\includegraphics[width=1\textwidth]{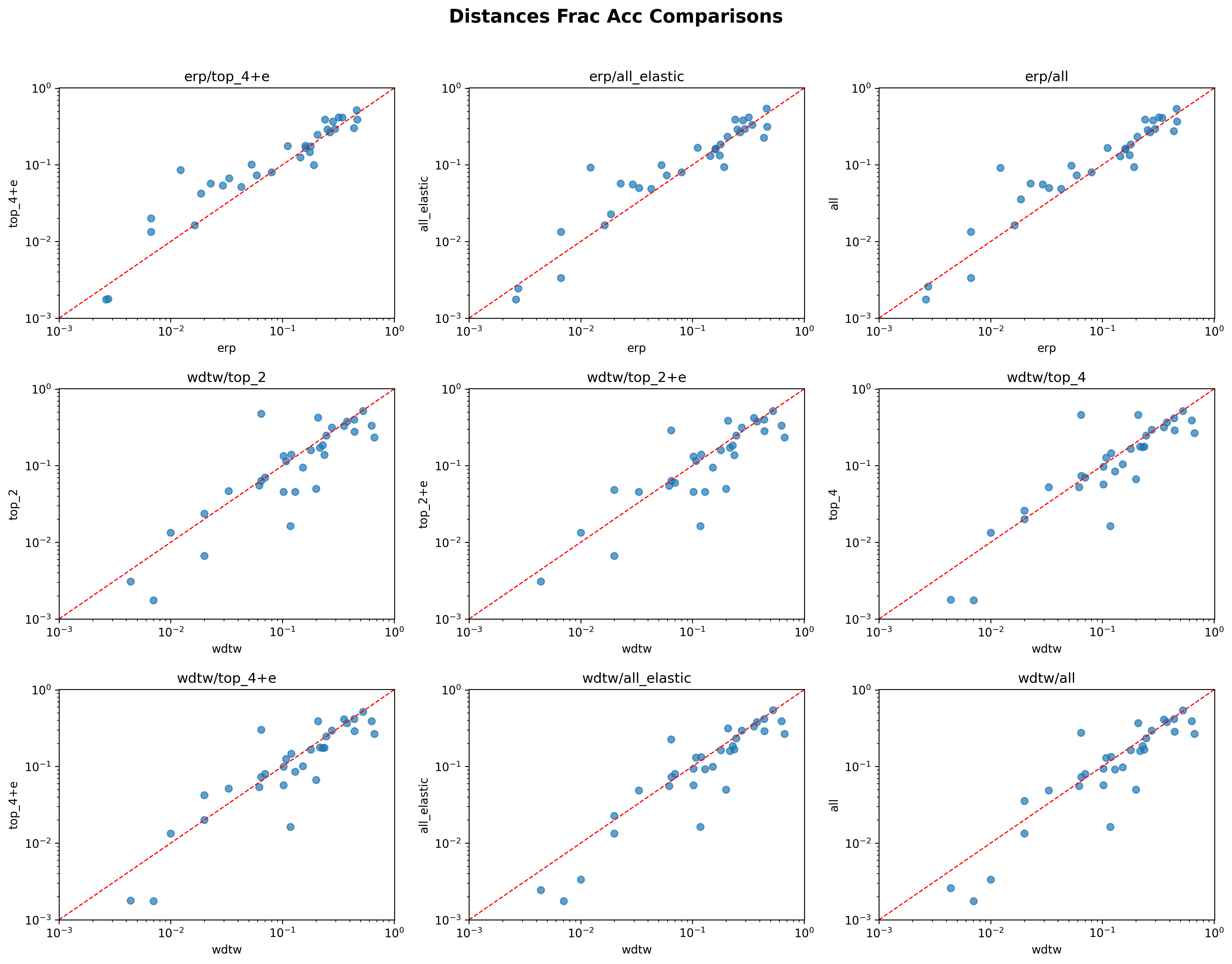}
	
	\caption{Distance Ensemble Comparison 7}
	\label{fig:distance frac comparison 7}
\end{figure}

\begin{figure}[H]
	\centering
	
	\includegraphics[width=1\textwidth]{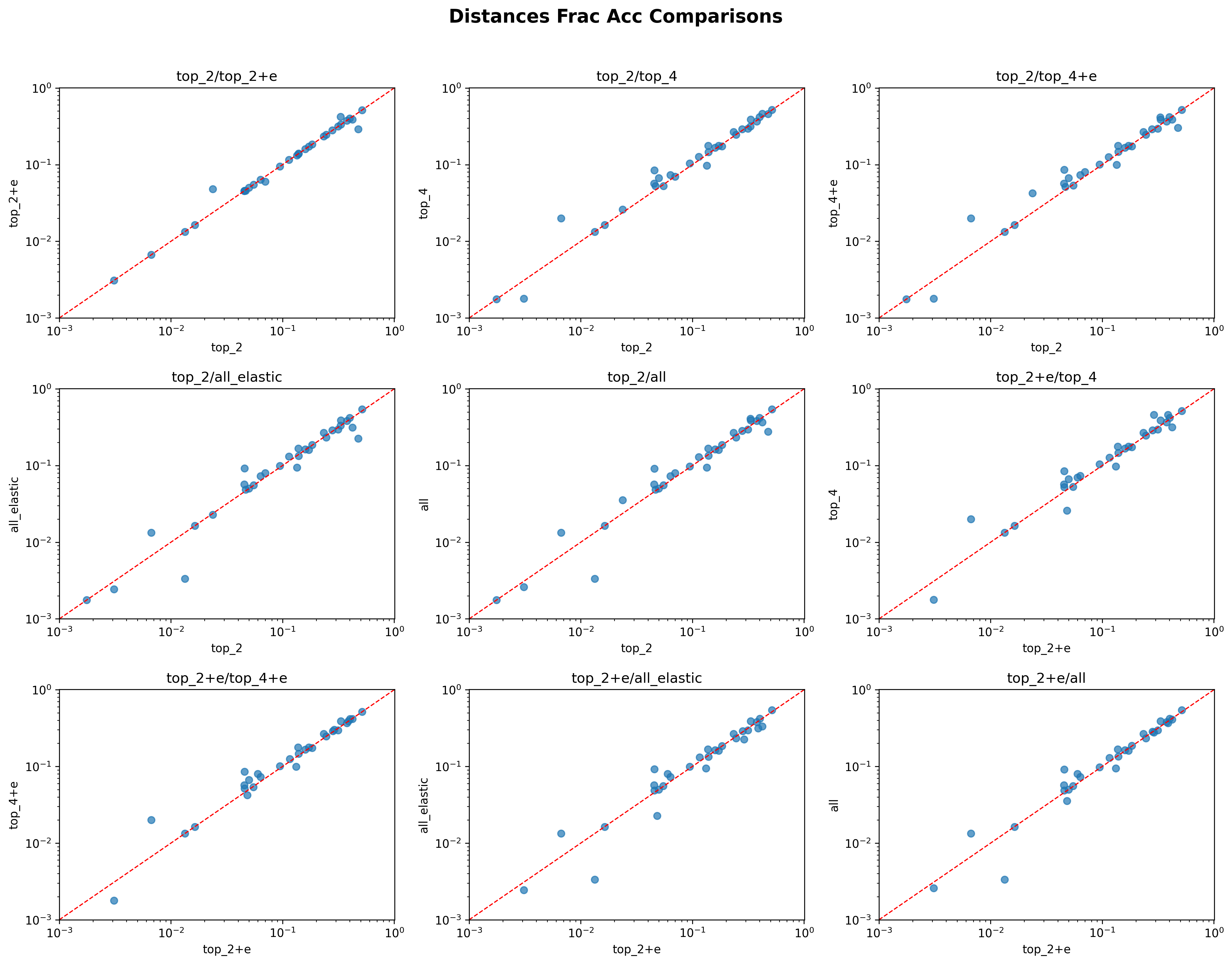}
	
	\caption{Distance Ensemble Comparison 8}
	\label{fig:distance frac comparison 8}
\end{figure}

\begin{figure}[H]
	\centering
	
	\includegraphics[width=1\textwidth]{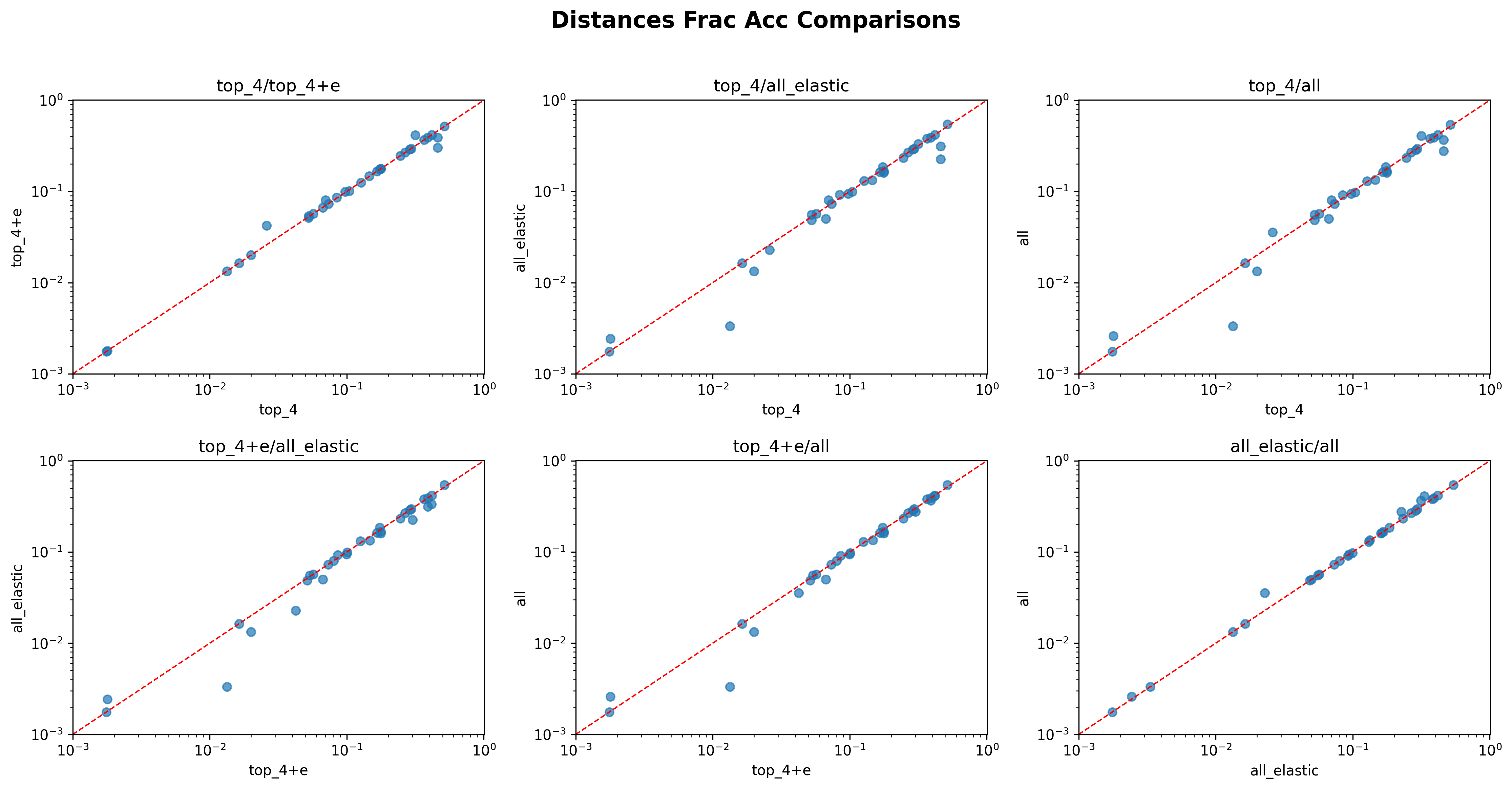}
	
	\caption{Distance Ensemble Comparison 9}
	\label{fig:distance frac comparison 9}
\end{figure}

\subsection{MSM Scaling Comparisons}

\begin{figure}[H]
	\centering
	
	\includegraphics[width=1\textwidth]{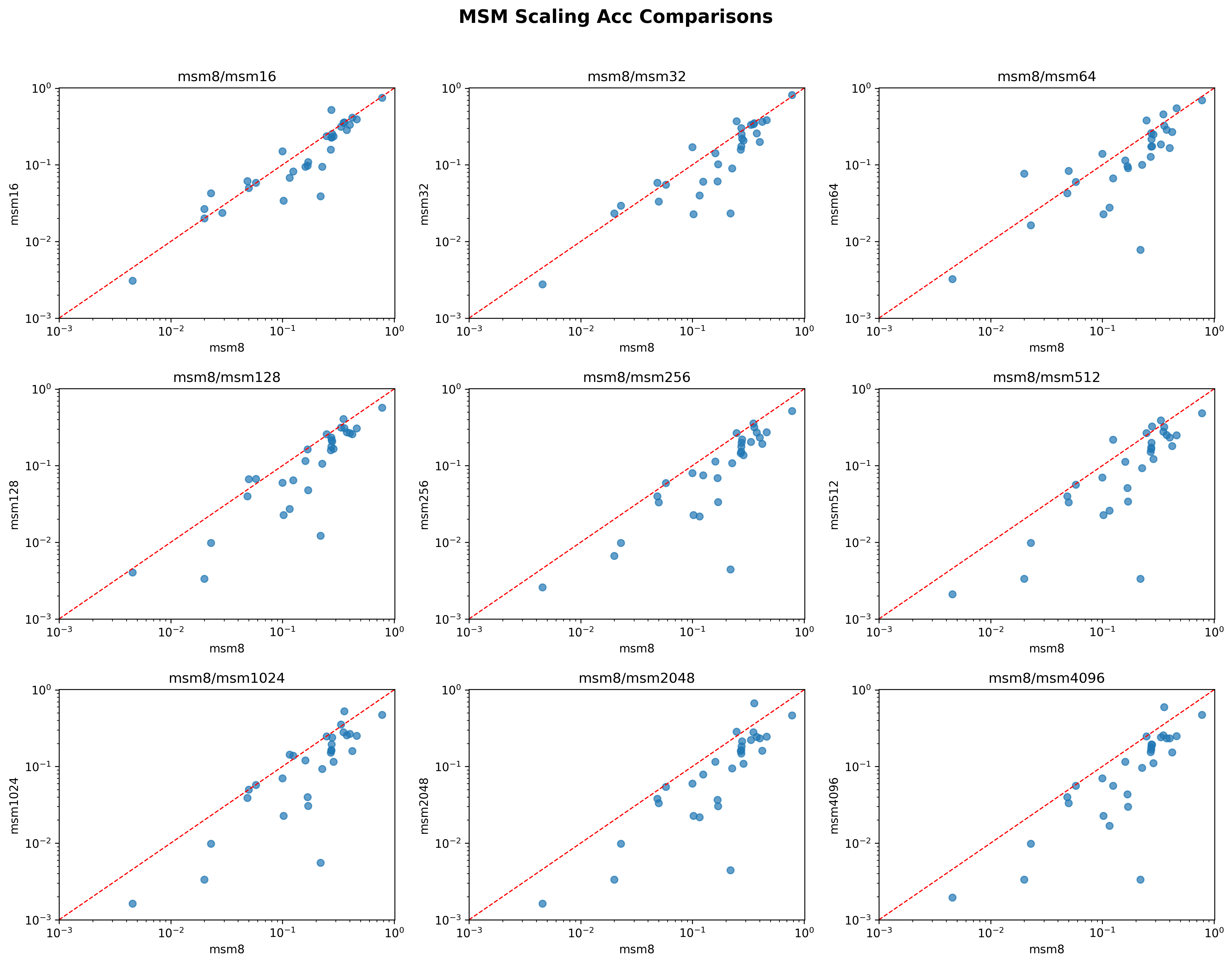}
	
	\caption{MSM Scaling Comparison 1}
	\label{fig:msm scaling distance comparisons 1}
\end{figure}

\begin{figure}[H]
	\centering
	
	\includegraphics[width=1\textwidth]{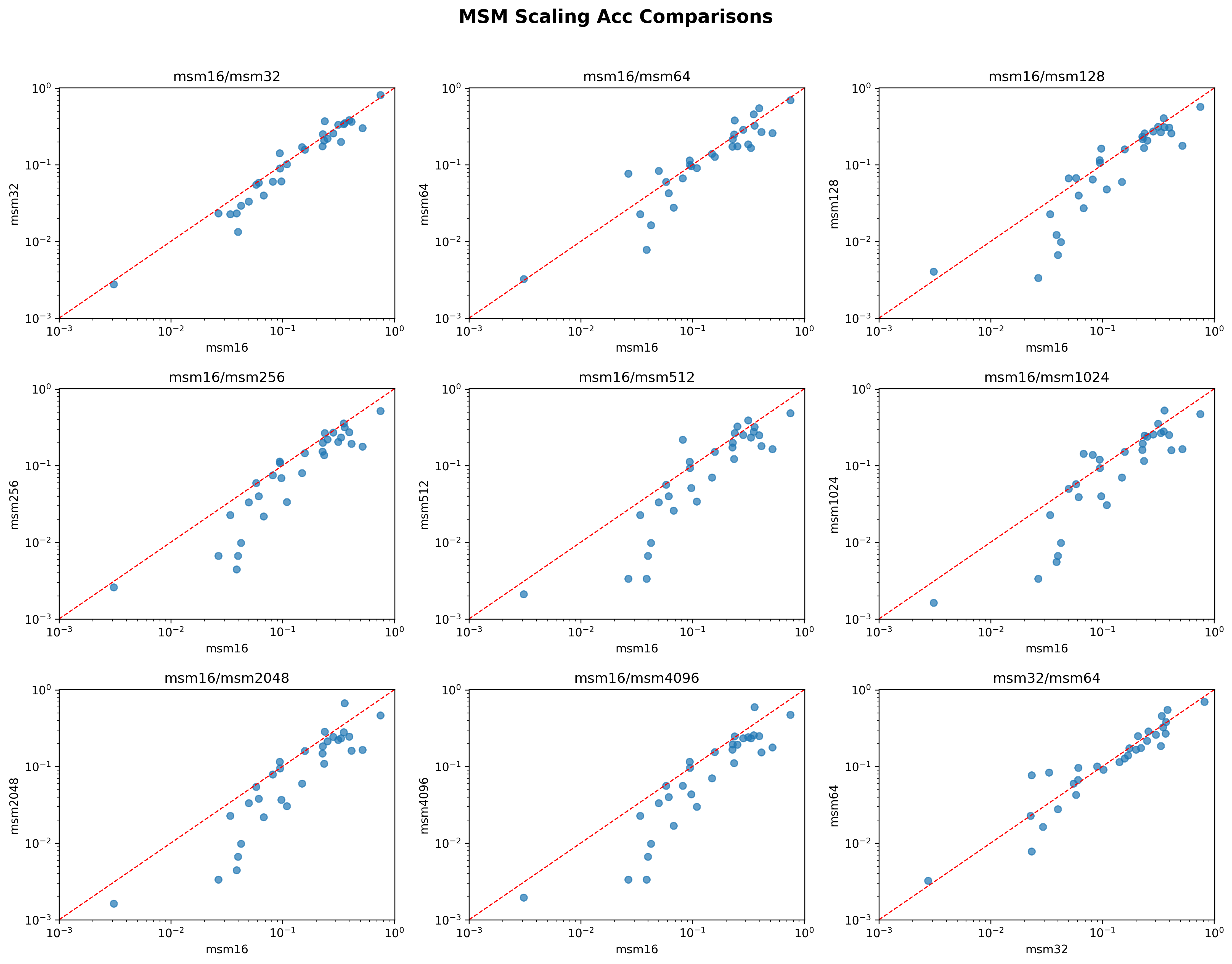}
	
	\caption{MSM Scaling Comparison 2}
	\label{fig:msm scaling distance comparisons 2}
\end{figure}

\begin{figure}[H]
	\centering
	
	\includegraphics[width=1\textwidth]{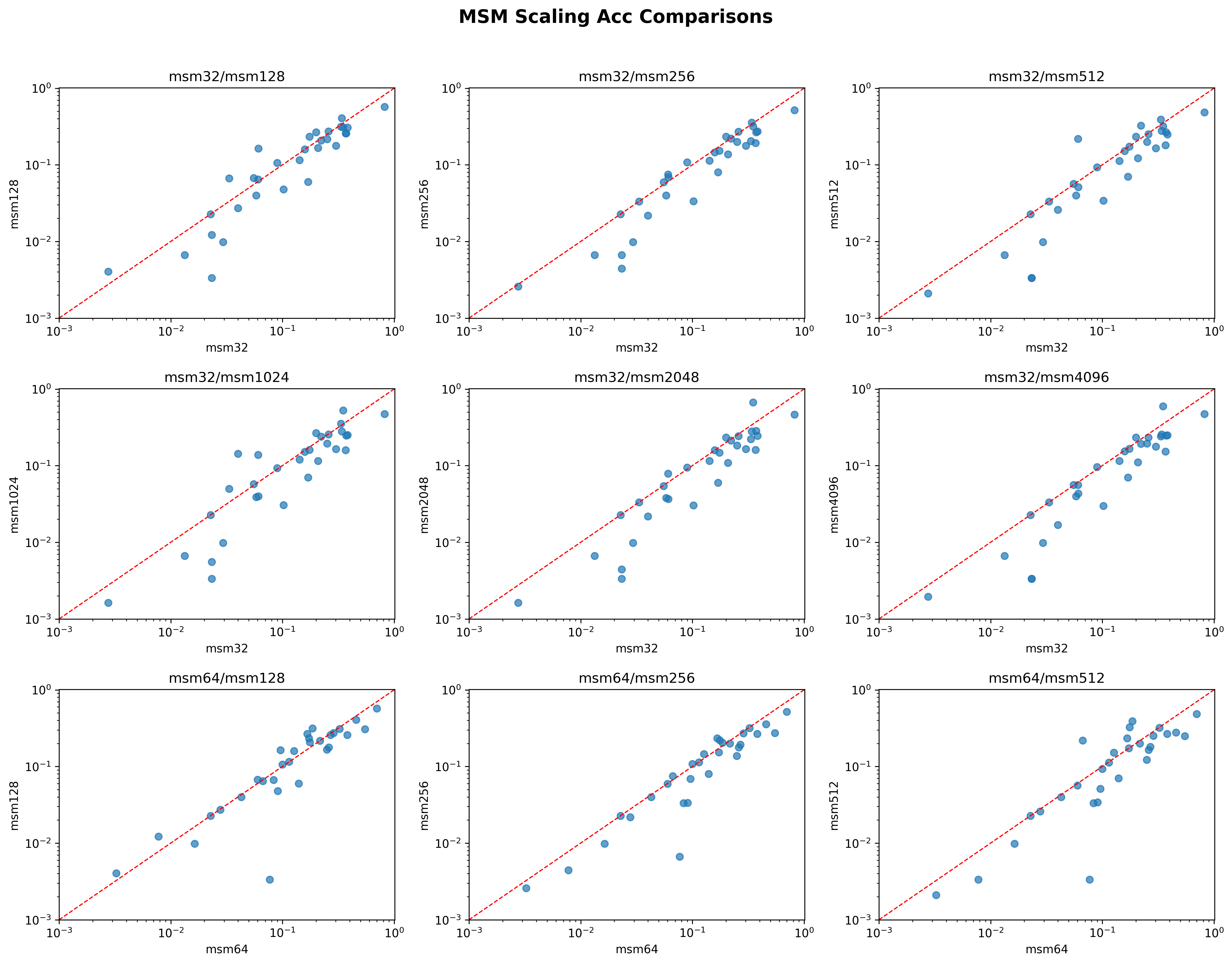}
	
	\caption{MSM Scaling Comparison 3}
	\label{fig:msm scaling distance comparisons 3}
\end{figure}

\begin{figure}[H]
	\centering
	
	\includegraphics[width=1\textwidth]{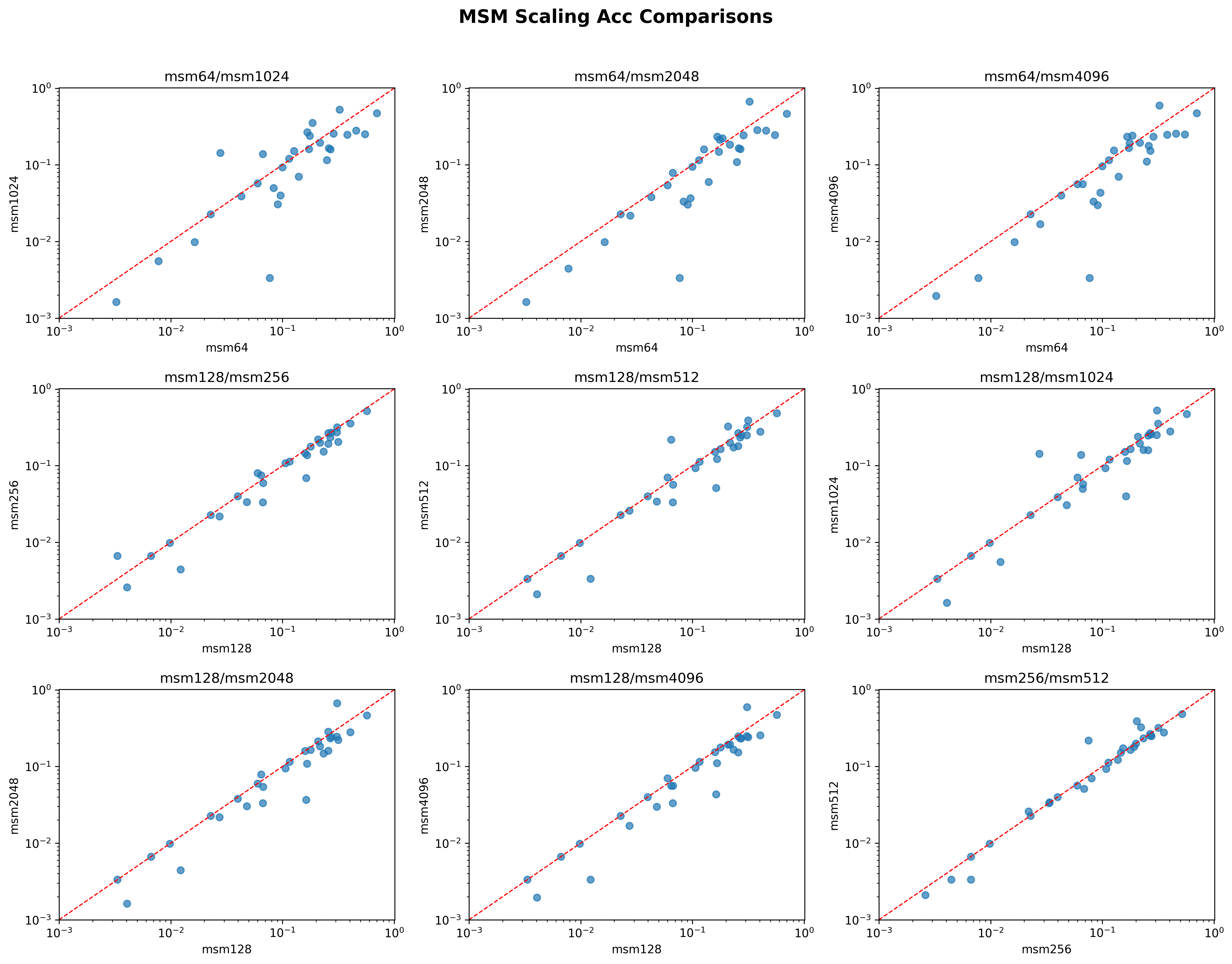}
	
	\caption{MSM Scaling Comparison 4}
	\label{fig:msm scaling distance comparisons 4}
\end{figure}

\begin{figure}[H]
	\centering
	
	\includegraphics[width=1\textwidth]{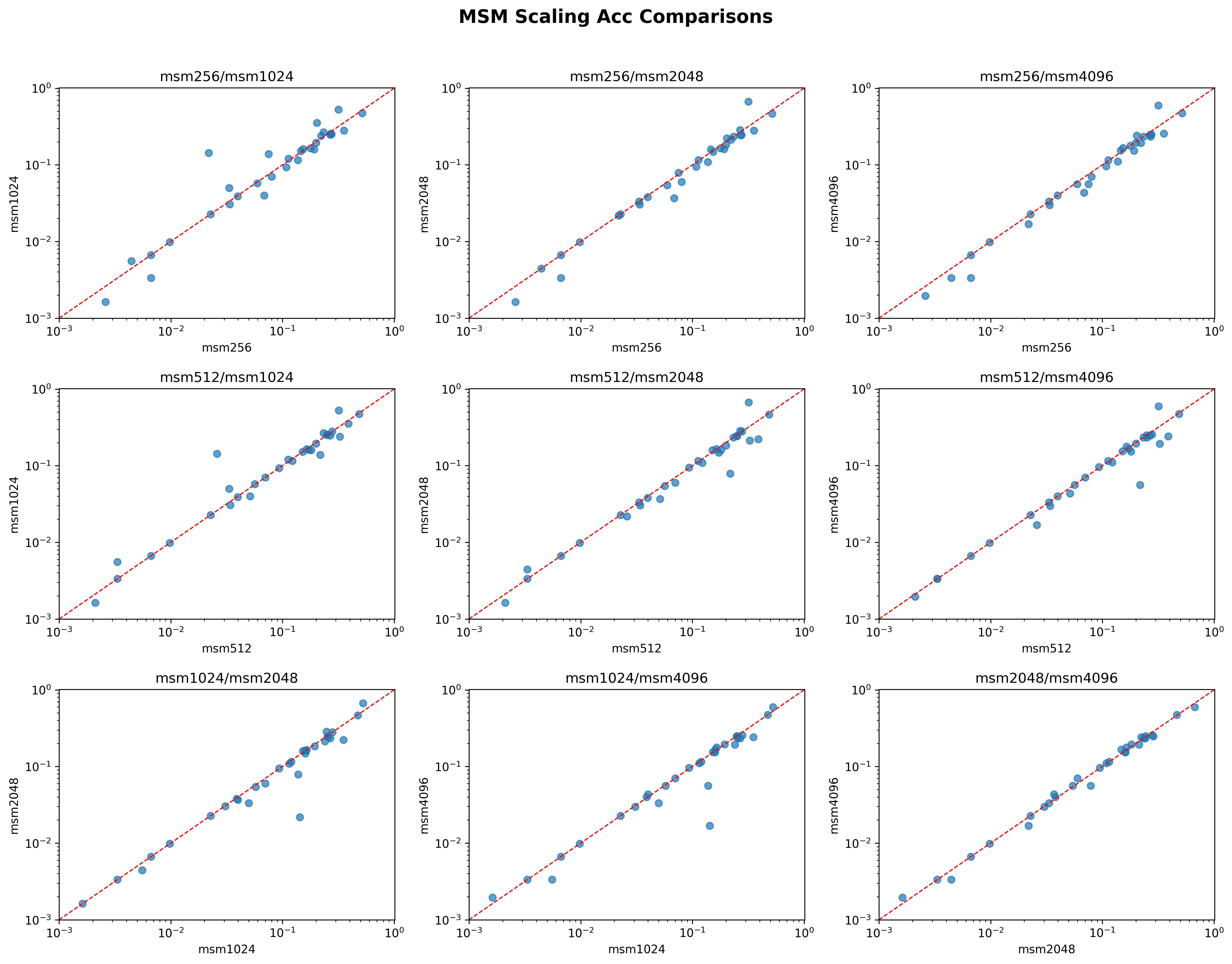}
	
	\caption{MSM Scaling Comparison 5}
	\label{fig:msm scaling distance comparisons 5}
\end{figure}

\subsection{Large Scale MSM Scaling Comparisons}

\begin{figure}[H]
	\centering
	
	\includegraphics[width=1\textwidth]{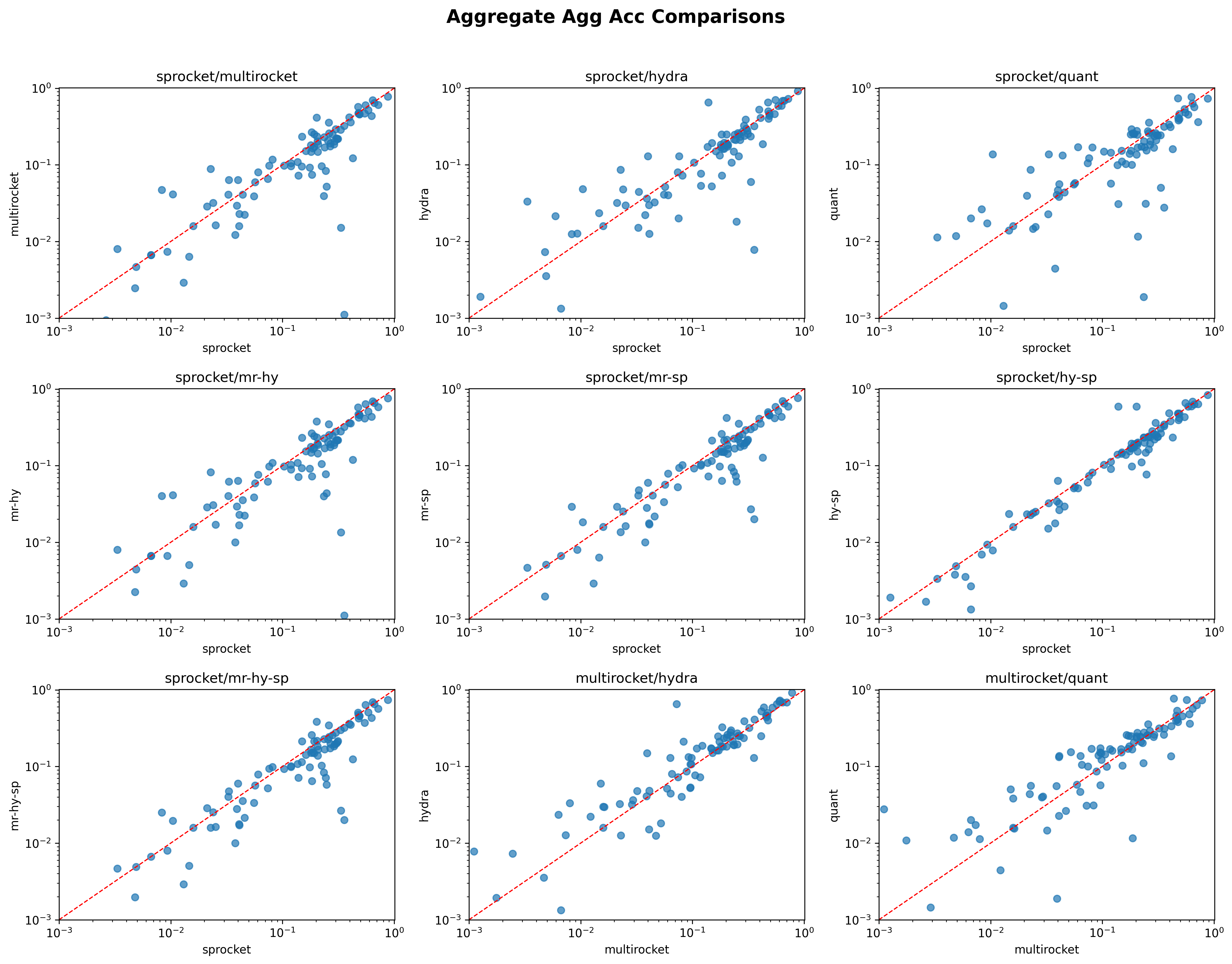}
	
	\caption{MSM Large Scale Comparison 1}
	\label{fig:msm large scal comparison 1}
\end{figure}

\begin{figure}[H]
	\centering
	
	\includegraphics[width=1\textwidth]{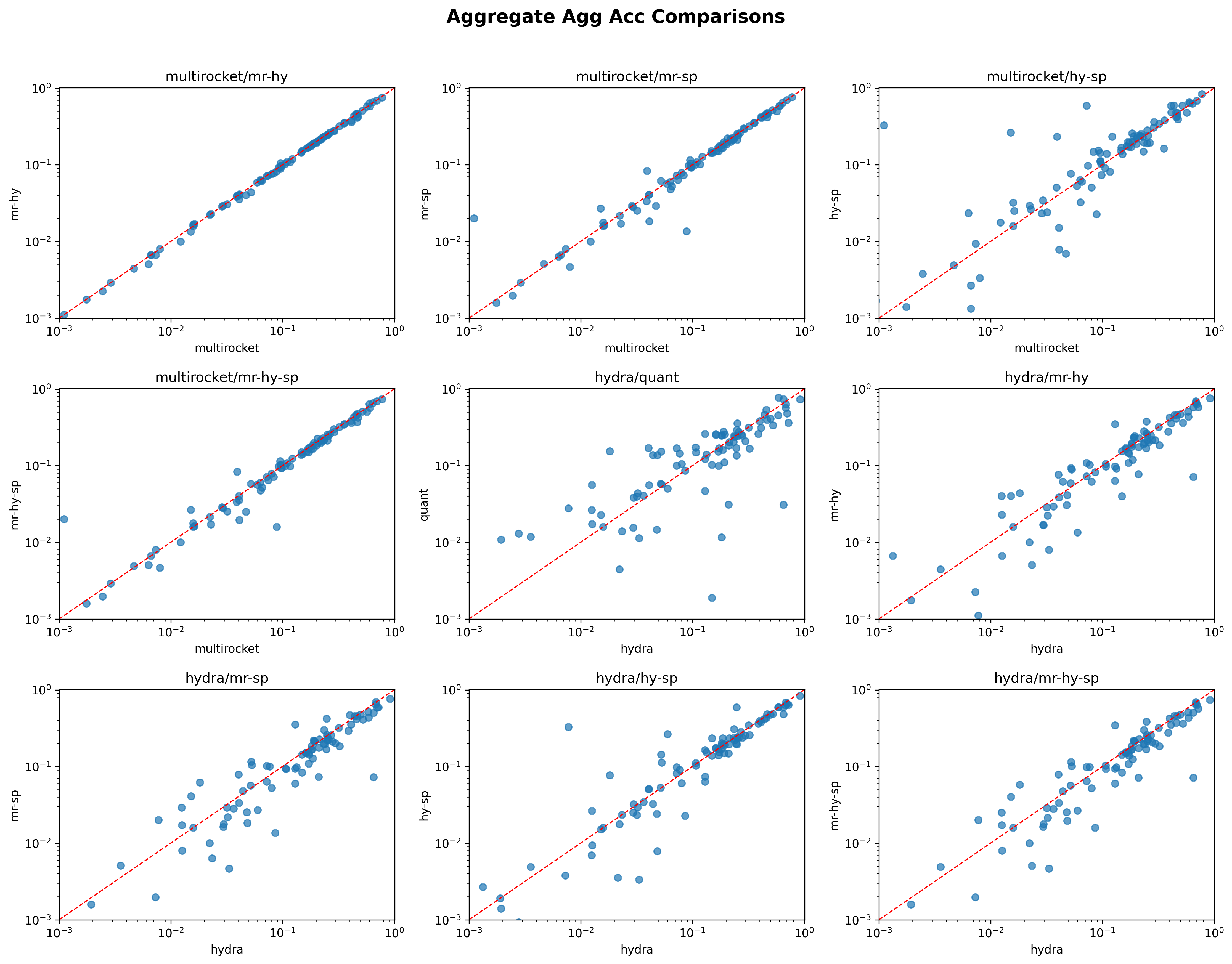}
	
	\caption{MSM Large Scale Comparison 2}
	\label{fig:msm large scal comparison 2}
\end{figure}

\begin{figure}[H]
	\centering
	
	\includegraphics[width=1\textwidth]{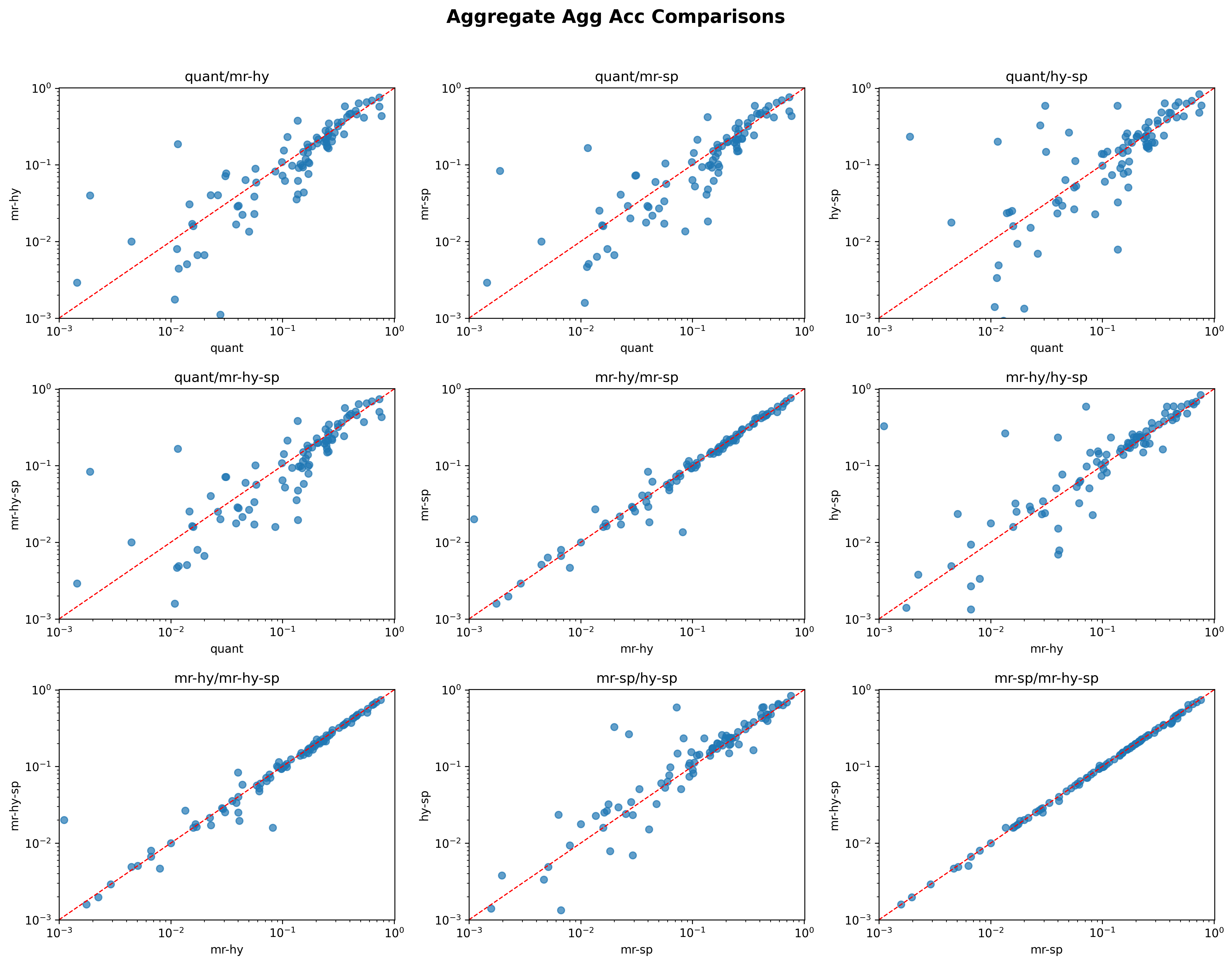}
	
	\caption{MSM Large Scale Comparison 3}
	\label{fig:msm large scal comparison 3}
\end{figure}

\section{Wall Clock Times for Large Scale Tests}

\subsection{MSM Large Scale Testing}
\begin{landscape}
	\begin{center}
		
		\setlength{\LTleft}{0pt}
		\setlength{\LTright}{0pt}
		
		\begin{longtable}{l|cccccccc}
			\caption{Average Wall Clock Time For All Algorithms on All Datasets} \\
			\toprule
			Dataset & sprocket & multirocket & hydra & quant & mr\_hy & mr\_sp & hy\_sp & mr\_hy\_sp \\
			\midrule
			\endfirsthead
			
			\toprule
			Dataset & sprocket & multirocket & hydra & quant & mr\_hy & mr\_sp & hy\_sp & mr\_hy\_sp \\
			\midrule
			\endhead
			
			\midrule
			\multicolumn{9}{r}{Continued on next page} \\
			\endfoot
			
			\bottomrule
			\endlastfoot
			
			DuckDuckGeese & 2663.19 & 6.90 & \textbf{0.26} & 6.85 & 6.85 & 2666.50 & 2660.18 & 2666.75 \\
			PEMS-SF       & 3380.17 & 6.53 & \textbf{1.31} & 17.65 & 7.06 & 3383.70 & 3377.74 & 3383.36 \\
			MindReading & 4263.92 & 7.28 & \textbf{1.78} & 1145.62 & 7.84 & 4267.04 & 4263.71 & 4269.41 \\
			PhonemeSpectra & 1815.93 & 30.70 & \textbf{9.87} & 56.76 & 70.01 & 1858.89 & 1819.89 & 1869.49 \\
			ShapeletSim & 59.93 & 0.74 & 0.55 & \textbf{0.39} & 1.05 & 60.43 & 60.36 & 60.92 \\
			EMOPain & 635.83 & 4.37 & \textbf{3.05} & 14.35 & 5.67 & 637.79 & 636.16 & 638.96 \\
			SmoothSubspace & 0.18 & 0.43 & \textbf{0.17} & 0.23 & 0.47 & 0.52 & 0.32 & 0.62 \\
			MelbournePedestrian & 5.50 & 4.81 & 1.19 & \textbf{0.84} & 4.67 & 7.49 & 4.72 & 7.83 \\
			ItalyPowerDemand & 0.71 & 0.78 & \textbf{0.17} & 0.35 & 0.89 & 1.45 & 0.86 & 1.60 \\
			Chinatown & 0.26 & 0.33 & \textbf{0.14} & 0.21 & 0.40 & 0.53 & 0.37 & 0.66 \\
			JapaneseVowels & 3.99 & 2.18 & 1.61 & \textbf{0.59} & 1.44 & 4.49 & 3.78 & 4.62 \\
			RacketSports & 2.99 & 0.43 & \textbf{0.21} & 0.34 & 0.46 & 1.27 & 1.10 & 1.40 \\
			LSST & 35.56 & 9.91 & \textbf{2.13} & 10.49 & 11.82 & 41.89 & 33.03 & 43.41 \\
			Libras & 1.36 & 1.51 & 0.84 & \textbf{0.39} & 1.54 & 2.43 & 2.00 & 2.13 \\
			FingerMovements & 17.34 & 1.45 & 1.41 & 2.06 & \textbf{1.35} & 15.71 & 14.93 & 15.73 \\
			NATOPS & 9.79 & 1.88 & 1.10 & \textbf{0.89} & 1.73 & 10.38 & 9.70 & 10.22 \\
			SharePriceIncrease & 9.05 & 3.61 & 2.09 & \textbf{1.91} & 3.26 & 10.10 & 8.26 & 10.79 \\
			SyntheticControl & 2.89 & 1.54 & 0.99 & \textbf{0.51} & 2.40 & 3.97 & 2.91 & 3.78 \\
			SonyAIBORobotSurface2 & 2.59 & 0.83 & \textbf{0.29} & 0.46 & 1.04 & 3.36 & 2.86 & 3.65 \\
			ERing & 1.92 & 0.38 & \textbf{0.17} & 0.85 & 0.48 & 2.23 & 2.06 & 2.39 \\
			SonyAIBORobotSurface1 & 1.82 & 0.55 & \textbf{0.23} & 0.43 & 0.70 & 2.30 & 2.04 & 2.53 \\
			PhalangesOutlinesCorrect & 20.97 & 6.57 & \textbf{1.99} & 4.39 & 6.99 & 23.63 & 18.77 & 24.06 \\
			ProximalPhalanxOutlineCorrect & 6.47 & 3.86 & 2.28 & \textbf{1.13} & 2.11 & 6.36 & 5.07 & 8.21 \\
			MiddlePhalanxOutlineCorrect & 6.55 & 1.94 & 1.96 & \textbf{1.07} & 3.13 & 7.27 & 5.87 & 6.97 \\
			DistalPhalanxOutlineCorrect & 6.49 & 3.16 & \textbf{0.48} & 1.01 & 1.57 & 6.27 & 5.87 & 7.25 \\
			ProximalPhalanxTW & 4.33 & 1.94 & 1.19 & \textbf{0.63} & 2.06 & 4.40 & 4.22 & 5.14 \\
			ProximalPhalanxOutlineAgeGroup & 4.43 & 1.95 & 1.91 & \textbf{0.80} & 1.70 & 4.93 & 4.48 & 5.16 \\
			MiddlePhalanxOutlineAgeGroup & 4.06 & 1.46 & 0.70 & \textbf{0.64} & 1.27 & 4.53 & 4.01 & 5.25 \\
			MiddlePhalanxTW & 3.49 & 1.61 & 1.06 & \textbf{0.78} & 2.12 & 4.37 & 4.25 & 4.90 \\
			DistalPhalanxTW & 3.95 & 1.54 & 1.47 & \textbf{0.63} & 2.21 & 4.79 & 3.64 & 4.33 \\
			DistalPhalanxOutlineAgeGroup & 3.98 & 1.34 & 0.93 & \textbf{0.72} & 1.28 & 3.90 & 3.38 & 4.42 \\
			TwoLeadECG & 4.32 & 1.04 & 0.37 & \textbf{0.32} & 1.34 & 5.29 & 4.68 & 5.66 \\
			MoteStrain & 4.90 & 1.11 & 0.40 & \textbf{0.32} & 1.44 & 5.95 & 5.29 & 6.35 \\
			ECG200 & 1.44 & 0.50 & 0.33 & \textbf{0.31} & 0.61 & 1.70 & 1.67 & 1.88 \\
			MedicalImages & 10.59 & 2.95 & 2.20 & \textbf{0.88} & 3.16 & 10.77 & 9.58 & 10.71 \\
			BasicMotions & 3.10 & 0.14 & \textbf{0.13} & 0.31 & 0.21 & 1.51 & 1.54 & 1.64 \\
			TwoPatterns & 68.07 & 7.10 & 3.40 & \textbf{1.89} & 9.29 & 73.67 & 69.42 & 75.47 \\
			CBF & 7.39 & 1.02 & 0.43 & \textbf{0.35} & 1.35 & 8.34 & 7.81 & 8.76 \\
			SwedishLeaf & 15.76 & 2.43 & 1.42 & \textbf{0.97} & 3.25 & 16.35 & 15.91 & 17.09 \\
			BME & 1.49 & 0.28 & 0.43 & \textbf{0.23} & 0.64 & 1.71 & 1.90 & 2.13 \\
			EyesOpenShut & 6.71 & 0.28 & \textbf{0.15} & 0.49 & 0.28 & 6.84 & 6.82 & 6.98 \\
			FacesUCR & 25.37 & 2.97 & 1.99 & \textbf{0.60} & 4.22 & 27.20 & 26.31 & 28.54 \\
			FaceAll & 32.09 & 3.83 & 1.99 & \textbf{1.35} & 5.01 & 34.00 & 32.33 & 34.90 \\
			ECGFiveDays & 7.66 & 0.96 & \textbf{0.76} & 0.99 & 1.63 & 8.54 & 8.42 & 9.31 \\
			ECG5000 & 76.16 & 5.36 & 4.11 & \textbf{1.00} & 8.18 & 81.29 & 78.41 & 83.57 \\
			ArticularyWordRecognition & 52.13 & 2.23 & \textbf{1.84} & 2.50 & 2.51 & 51.60 & 50.90 & 51.77 \\
			PowerCons & 5.40 & 1.25 & 1.05 & \textbf{0.39} & 1.60 & 6.01 & 5.65 & 5.87 \\
			Plane & 2.95 & 0.63 & 0.51 & \textbf{0.33} & 0.69 & 3.41 & 3.25 & 3.50 \\
			GunPointOldVersusYoung & 7.11 & 1.48 & 1.61 & \textbf{0.31} & 1.78 & 7.71 & 7.32 & 7.63 \\
			GunPointMaleVersusFemale & 6.66 & 1.72 & 1.53 & \textbf{0.38} & 2.48 & 7.61 & 7.73 & 8.17 \\
			GunPointAgeSpan & 7.10 & 1.83 & 1.96 & \textbf{0.39} & 2.27 & 7.70 & 7.43 & 7.85 \\
			GunPoint & 2.16 & 0.35 & \textbf{0.17} & 0.30 & 0.42 & 2.43 & 2.31 & 2.59 \\
			UMD & 2.01 & 0.30 & \textbf{0.17} & 0.31 & 0.39 & 2.23 & 2.16 & 2.40 \\
			Wafer & 128.71 & 9.43 & 7.29 & \textbf{1.17} & 17.15 & 140.25 & 131.39 & 144.28 \\
			Handwriting & 24.27 & 1.59 & \textbf{0.70} & 1.04 & 2.03 & 25.60 & 24.84 & 26.20 \\
			ChlorineConcentration & 92.67 & 6.12 & 3.72 & \textbf{1.90} & 8.52 & 95.98 & 93.75 & 98.69 \\
			Adiac & 18.44 & 2.19 & 2.12 & \textbf{1.39} & 2.24 & 18.18 & 17.84 & 18.82 \\
			Epilepsy2 & 218.31 & 24.36 & 8.59 & \textbf{0.77} & 89.77 & 288.70 & 226.87 & 306.67 \\
			Colposcopy & 3.90 & 0.72 & \textbf{0.38} & 0.40 & 0.75 & 4.29 & 4.04 & 4.46 \\
			Fungi & 3.28 & 0.36 & \textbf{0.24} & 0.32 & 0.48 & 3.53 & 3.49 & 3.75 \\
			Epilepsy & 14.63 & 1.70 & 1.18 & \textbf{0.59} & 1.76 & 13.59 & 13.12 & 13.77 \\
			Wine & 2.82 & 0.25 & \textbf{0.16} & 0.33 & 0.29 & 2.97 & 2.95 & 3.12 \\
			Strawberry & 39.42 & 3.05 & 2.64 & \textbf{1.37} & 3.23 & 39.75 & 38.56 & 40.49 \\
			ArrowHead & 5.57 & 0.36 & \textbf{0.26} & 0.30 & 0.49 & 5.81 & 5.81 & 6.06 \\
			ElectricDeviceDetection & 215.68 & 4.91 & 5.44 & \textbf{1.79} & 8.89 & 220.16 & 219.67 & 223.29 \\
			WordSynonyms & 49.14 & 2.42 & 1.81 & \textbf{0.85} & 2.47 & 49.72 & 49.60 & 50.61 \\
			FiftyWords & 47.88 & 2.19 & \textbf{1.59} & 1.82 & 2.58 & 48.67 & 48.63 & 50.48 \\
			Trace & 8.83 & 0.64 & 0.42 & \textbf{0.37} & 0.77 & 9.21 & 9.12 & 9.50 \\
			ToeSegmentation1 & 9.72 & 0.44 & 0.39 & \textbf{0.28} & 0.69 & 10.05 & 10.08 & 10.42 \\
			DodgerLoopWeekend & 5.60 & 0.35 & 0.26 & \textbf{0.26} & 0.46 & 5.81 & 5.83 & 6.06 \\
			DodgerLoopGame & 6.20 & 0.33 & 0.25 & \textbf{0.23} & 0.45 & 6.40 & 6.42 & 6.64 \\
			DodgerLoopDay & 7.85 & 0.35 & \textbf{0.22} & 0.31 & 0.44 & 8.08 & 8.03 & 8.28 \\
			CricketZ & 55.73 & 2.53 & 2.01 & \textbf{1.14} & 2.97 & 56.07 & 55.42 & 56.75 \\
			CricketY & 56.56 & 1.75 & 1.40 & \textbf{1.08} & 3.23 & 57.98 & 57.75 & 59.32 \\
			CricketX & 56.02 & 2.07 & 1.58 & \textbf{1.15} & 2.91 & 57.22 & 56.68 & 57.93 \\
			FreezerRegularTrain & 168.66 & 4.39 & 3.89 & \textbf{0.67} & 8.03 & 172.79 & 172.46 & 176.60 \\
			FreezerSmallTrain & 134.02 & 4.01 & 3.58 & \textbf{0.55} & 7.46 & 137.89 & 137.60 & 141.47 \\
			UWaveGestureLibraryZ & 335.95 & 6.22 & 5.91 & \textbf{2.90} & 12.75 & 342.19 & 341.39 & 347.15 \\
			UWaveGestureLibraryY & 358.37 & 6.13 & 6.21 & \textbf{3.19} & 12.80 & 364.54 & 363.31 & 370.48 \\
			UWaveGestureLibraryX & 361.95 & 6.40 & 5.98 & \textbf{2.97} & 13.94 & 369.29 & 368.64 & 375.56 \\
			UWaveGestureLibrary & 47.53 & 1.47 & 1.30 & \textbf{0.69} & 2.04 & 48.76 & 48.41 & 49.11 \\
			Lightning7 & 9.76 & 0.37 & \textbf{0.23} & 0.35 & 0.46 & 10.00 & 9.95 & 10.22 \\
			ToeSegmentation2 & 9.64 & 0.37 & 0.29 & \textbf{0.29} & 0.52 & 9.87 & 9.91 & 10.16 \\
			DiatomSizeReduction & 18.13 & 0.53 & 0.52 & \textbf{0.32} & 0.91 & 18.52 & 18.63 & 19.04 \\
			FaceFour & 7.52 & 0.31 & \textbf{0.21} & 0.27 & 0.37 & 7.68 & 7.70 & 7.88 \\
			GestureMidAirD3 & 30.59 & 1.61 & 1.79 & \textbf{0.86} & 2.43 & 31.93 & 31.10 & 32.17 \\
			GestureMidAirD2 & 33.72 & 1.84 & 2.08 & \textbf{0.88} & 2.55 & 34.52 & 34.01 & 35.41 \\
			GestureMidAirD1 & 33.03 & 1.31 & 1.31 & \textbf{1.14} & 2.31 & 33.63 & 33.78 & 34.47 \\
			Symbols & 116.93 & 1.52 & 1.76 & \textbf{0.42} & 3.14 & 118.32 & 118.68 & 120.07 \\
			HandMovementDirection & 198.87 & 1.01 & \textbf{0.57} & 1.34 & 1.24 & 199.54 & 199.34 & 200.07 \\
			Heartbeat & 2193.58 & 2.62 & \textbf{2.23} & 4.10 & 3.83 & 2194.98 & 2194.59 & 2195.90 \\
			Yoga & 803.03 & 5.64 & 6.02 & \textbf{1.16} & 10.14 & 807.20 & 807.84 & 812.77 \\
			OSULeaf & 81.53 & 1.63 & 1.96 & \textbf{0.62} & 2.77 & 82.77 & 82.67 & 83.50 \\
			Meat & 21.25 & 0.32 & \textbf{0.23} & 0.32 & 0.40 & 21.44 & 21.45 & 21.66 \\
			Fish & 92.63 & 1.40 & 1.72 & \textbf{0.62} & 2.44 & 93.63 & 93.33 & 94.54 \\
			FordA & 2472.92 & 22.51 & \textbf{13.86} & 16.51 & 40.60 & 2494.93 & 2483.25 & 2508.32 \\
			FordB & 2369.59 & 71.60 & \textbf{17.58} & 24.21 & 97.10 & 2421.89 & 2376.32 & 2436.19 \\
			Ham & 45.46 & 0.78 & 1.01 & \textbf{0.64} & 1.26 & 46.01 & 46.03 & 46.66 \\
			
		\end{longtable}
		
	\end{center}
\end{landscape}

\subsection{Euclidean Large Scale Testing}
\begin{landscape}[H]
	\begin{center}
		
		\setlength{\LTleft}{0pt}
		\setlength{\LTright}{0pt}
		
		\begin{longtable}{l|cccccccc}
			\caption{Average Wall Clock Time For All Algorithms on All Datasets with Euclidean SPROCKET} \\
			\toprule
			Dataset & sprocket & multirocket & hydra & quant & mr\_hy & mr\_sp & hy\_sp & mr\_hy\_sp \\
			\midrule
			\endfirsthead
			
			\toprule
			Dataset & sprocket & multirocket & hydra & quant & mr\_hy & mr\_sp & hy\_sp & mr\_hy\_sp \\
			\midrule
			\endhead
			
			\midrule
			\multicolumn{9}{r}{Continued on next page} \\
			\endfoot
			
			\bottomrule
			\endlastfoot
			
			DuckDuckGeese & 7.29 & 7.78 & \textbf{0.28} & 7.06 & 7.62 & 11.03 & 3.89 & 11.29 \\
			PEMS-SF & 10.63 & 7.74 & \textbf{1.03} & 15.36 & 8.04 & 14.19 & 8.26 & 15.44 \\
			MindReading & 7.74 & 10.00 & \textbf{2.18} & 1085.49 & 10.61 & 13.10 & 7.74 & 15.19 \\
			PhonemeSpectra & 14.08 & 56.10 & \textbf{10.38} & 62.79 & 92.14 & 76.54 & 16.06 & 83.04 \\
			ShapeletSim & \textbf{0.11} & 0.80 & 0.56 & 0.47 & 1.11 & 0.71 & 0.58 & 1.20 \\
			EMOPain & 2.73 & 4.40 & \textbf{2.39} & 13.73 & 5.86 & 4.80 & 3.70 & 6.76 \\
			SmoothSubspace & \textbf{0.10} & 0.46 & 0.18 & 0.23 & 0.45 & 0.41 & 0.24 & 0.51 \\
			MelbournePedestrian & 2.05 & 4.50 & 1.06 & \textbf{0.92} & 5.46 & 5.04 & 1.16 & 6.35 \\
			ItalyPowerDemand & \textbf{0.12} & 0.81 & 0.15 & 0.34 & 0.86 & 0.84 & 0.24 & 0.99 \\
			Chinatown & \textbf{0.08} & 0.31 & 0.13 & 0.21 & 0.38 & 0.35 & 0.18 & 0.45 \\
			JapaneseVowels & 0.79 & 1.48 & 1.22 & 1.11 & 1.77 & 1.97 & \textbf{0.77} & 1.83 \\
			RacketSports & 2.47 & 0.45 & \textbf{0.20} & 0.39 & 0.49 & 0.42 & 0.25 & 0.56 \\
			LSST & 6.25 & 13.58 & \textbf{3.20} & 12.92 & 19.47 & 18.17 & 3.28 & 19.92 \\
			Libras & 0.62 & 1.45 & 1.13 & \textbf{0.44} & 1.58 & 1.52 & 1.13 & 1.32 \\
			FingerMovements & 2.38 & 1.30 & 1.97 & 1.61 & 1.84 & 2.10 & \textbf{0.97} & 1.67 \\
			NATOPS & 1.04 & 1.52 & 1.12 & \textbf{0.84} & 1.56 & 1.65 & 1.22 & 1.70 \\
			SharePriceIncrease & 1.56 & 3.16 & \textbf{1.36} & 1.80 & 4.30 & 3.66 & 2.48 & 4.60 \\
			SyntheticControl & 0.80 & 2.24 & 0.93 & \textbf{0.43} & 2.18 & 1.40 & 0.70 & 1.77 \\
			SonyAIBORobotSurface2 & \textbf{0.15} & 0.89 & 0.33 & 0.48 & 1.12 & 0.95 & 0.45 & 1.26 \\
			ERing & \textbf{0.12} & 0.35 & 0.20 & 0.53 & 0.44 & 0.36 & 0.28 & 0.56 \\
			SonyAIBORobotSurface1 & \textbf{0.10} & 0.58 & 0.24 & 0.27 & 0.74 & 0.61 & 0.32 & 0.85 \\
			PhalangesOutlinesCorrect & 3.21 & 7.06 & \textbf{2.35} & 3.75 & 8.81 & 10.77 & 4.96 & 9.37 \\
			ProximalPhalanxOutlineCorrect & 1.55 & 3.02 & 2.56 & \textbf{1.12} & 2.79 & 2.14 & 1.64 & 2.84 \\
			MiddlePhalanxOutlineCorrect & 2.36 & 2.99 & 1.22 & 1.24 & 1.90 & 2.30 & \textbf{1.01} & 3.39 \\
			DistalPhalanxOutlineCorrect & 1.64 & 2.77 & 1.21 & 1.07 & 3.02 & 2.02 & \textbf{0.67} & 2.57 \\
			ProximalPhalanxTW & 1.12 & 1.65 & 1.60 & \textbf{0.58} & 2.61 & 2.05 & 0.63 & 1.74 \\
			ProximalPhalanxOutlineAgeGroup & 1.10 & 2.05 & 1.45 & \textbf{0.70} & 2.10 & 2.17 & 0.70 & 1.51 \\
			MiddlePhalanxOutlineAgeGroup & 1.56 & 2.20 & 0.87 & \textbf{0.68} & 1.15 & 2.44 & 0.88 & 2.13 \\
			MiddlePhalanxTW & 1.07 & 2.72 & 1.14 & \textbf{0.76} & 1.59 & 1.67 & 0.82 & 1.91 \\
			DistalPhalanxTW & 1.32 & 2.35 & 0.95 & 0.64 & 1.76 & 1.04 & \textbf{0.48} & 1.32 \\
			DistalPhalanxOutlineAgeGroup & 0.99 & 1.30 & 1.13 & \textbf{0.62} & 1.66 & 1.53 & 0.73 & 1.45 \\
			TwoLeadECG & \textbf{0.15} & 1.10 & 0.39 & 0.32 & 1.41 & 1.18 & 0.52 & 1.55 \\
			MoteStrain & \textbf{0.14} & 1.16 & 0.42 & 0.37 & 1.50 & 1.24 & 0.53 & 1.64 \\
			ECG200 & \textbf{0.21} & 0.51 & 0.33 & 0.36 & 0.53 & 0.49 & 0.40 & 0.63 \\
			MedicalImages & 1.37 & 2.96 & 1.43 & \textbf{0.92} & 2.01 & 1.89 & 1.81 & 2.72 \\
			BasicMotions & 1.87 & 0.17 & 0.14 & 0.35 & 0.22 & \textbf{0.12} & 0.15 & 0.26 \\
			TwoPatterns & \textbf{1.64} & 8.72 & 2.66 & 2.13 & 15.27 & 13.58 & 3.32 & 15.71 \\
			CBF & \textbf{0.11} & 1.09 & 0.44 & 0.32 & 1.43 & 1.10 & 0.54 & 1.53 \\
			SwedishLeaf & 1.37 & 2.85 & 2.03 & \textbf{0.97} & 3.35 & 2.49 & 1.01 & 2.24 \\
			BME & \textbf{0.07} & 0.31 & 0.18 & 0.28 & 0.39 & 0.29 & 0.22 & 0.46 \\
			EyesOpenShut & \textbf{0.09} & 0.30 & 0.15 & 0.51 & 0.29 & 0.23 & 0.20 & 0.37 \\
			FacesUCR & 1.01 & 2.98 & 2.42 & \textbf{0.54} & 4.52 & 3.44 & 2.57 & 4.54 \\
			FaceAll & \textbf{1.93} & 3.57 & 2.12 & 2.11 & 5.53 & 4.49 & 3.15 & 5.91 \\
			ECGFiveDays & \textbf{0.14} & 0.91 & 0.52 & 0.31 & 1.35 & 0.96 & 0.64 & 1.49 \\
			ECG5000 & 1.68 & 7.12 & 3.89 & \textbf{1.07} & 12.85 & 10.46 & 3.61 & 12.74 \\
			ArticularyWordRecognition & 2.41 & 1.66 & 1.54 & 2.92 & 1.98 & 1.76 & \textbf{1.37} & 2.32 \\
			PowerCons & \textbf{0.32} & 1.34 & 1.29 & 0.37 & 1.94 & 1.23 & 0.93 & 1.55 \\
			Plane & \textbf{0.25} & 0.59 & 0.41 & 0.31 & 0.65 & 0.57 & 0.46 & 0.78 \\
			GunPointOldVersusYoung & 0.87 & 1.55 & 1.74 & \textbf{0.29} & 2.23 & 2.07 & 1.63 & 1.78 \\
			GunPointMaleVersusFemale & 0.91 & 1.90 & 1.64 & \textbf{0.38} & 2.10 & 1.62 & 1.46 & 2.05 \\
			GunPointAgeSpan & 1.05 & 2.14 & 1.69 & \textbf{0.41} & 2.47 & 1.96 & 1.38 & 2.09 \\
			GunPoint & \textbf{0.11} & 0.33 & 0.18 & 0.30 & 0.39 & 0.35 & 0.25 & 0.51 \\
			UMD & \textbf{0.07} & 0.32 & 0.18 & 0.30 & 0.38 & 0.29 & 0.22 & 0.45 \\
			Wafer & 3.01 & 15.77 & 4.03 & \textbf{1.21} & 22.83 & 18.19 & 5.18 & 23.60 \\
			Handwriting & \textbf{0.48} & 1.73 & 0.80 & 1.24 & 2.19 & 1.91 & 1.09 & 2.55 \\
			ChlorineConcentration & \textbf{1.80} & 6.42 & 3.12 & 2.18 & 11.03 & 8.43 & 3.76 & 11.07 \\
			Adiac & \textbf{1.47} & 2.31 & 1.69 & 1.54 & 2.97 & 2.62 & 1.91 & 2.57 \\
			Epilepsy2 & 1.71 & 50.86 & 21.61 & \textbf{0.96} & 174.55 & 131.00 & 23.19 & 172.58 \\
			Colposcopy & \textbf{0.20} & 0.94 & 0.42 & 0.59 & 0.95 & 0.76 & 0.45 & 1.02 \\
			Fungi & \textbf{0.05} & 0.40 & 0.25 & 0.38 & 0.52 & 0.33 & 0.27 & 0.58 \\
			Epilepsy & 3.48 & 1.69 & 1.01 & \textbf{0.62} & 1.77 & 1.62 & 1.12 & 1.93 \\
			Wine & \textbf{0.06} & 0.30 & 0.16 & 0.35 & 0.33 & 0.23 & 0.19 & 0.40 \\
			Strawberry & 1.71 & 2.84 & 3.37 & \textbf{1.30} & 3.14 & 2.95 & 2.53 & 3.70 \\
			ArrowHead & \textbf{0.09} & 0.42 & 0.26 & 0.28 & 0.54 & 0.38 & 0.33 & 0.64 \\
			ElectricDeviceDetection & 1.55 & 6.05 & 4.41 & \textbf{1.09} & 11.67 & 8.14 & 5.67 & 12.04 \\
			WordSynonyms & 1.00 & 2.73 & 1.69 & \textbf{0.85} & 2.68 & 2.07 & 2.02 & 2.56 \\
			FiftyWords & 1.73 & 2.47 & 1.83 & \textbf{1.64} & 2.60 & 2.09 & 1.73 & 3.23 \\
			Trace & \textbf{0.23} & 0.72 & 0.52 & 0.38 & 0.84 & 0.62 & 0.61 & 0.96 \\
			ToeSegmentation1 & \textbf{0.12} & 0.51 & 0.38 & 0.39 & 0.79 & 0.53 & 0.47 & 0.91 \\
			DodgerLoopWeekend & \textbf{0.09} & 0.39 & 0.26 & 0.31 & 0.49 & 0.31 & 0.30 & 0.55 \\
			DodgerLoopGame & \textbf{0.08} & 0.39 & 0.27 & 0.32 & 0.50 & 0.30 & 0.31 & 0.56 \\
			DodgerLoopDay & \textbf{0.09} & 0.41 & 0.23 & 0.37 & 0.48 & 0.33 & 0.27 & 0.56 \\
			CricketZ & 2.14 & 2.80 & 2.11 & \textbf{1.25} & 2.82 & 2.65 & 2.08 & 3.94 \\
			CricketY & \textbf{0.88} & 1.90 & 1.64 & 1.23 & 3.04 & 1.82 & 1.56 & 3.16 \\
			CricketX & 1.49 & 2.65 & 1.62 & \textbf{1.17} & 3.23 & 2.53 & 2.04 & 2.94 \\
			FreezerRegularTrain & 0.84 & 4.84 & 3.86 & \textbf{0.74} & 9.01 & 5.84 & 4.66 & 9.82 \\
			FreezerSmallTrain & 0.71 & 4.81 & 3.67 & \textbf{0.64} & 9.61 & 6.12 & 4.37 & 10.03 \\
			UWaveGestureLibraryZ & 3.38 & 9.10 & 6.09 & \textbf{2.93} & 18.16 & 13.66 & 8.46 & 19.47 \\
			UWaveGestureLibraryY & \textbf{2.38} & 9.74 & 6.68 & 3.07 & 19.15 & 14.08 & 11.86 & 19.48 \\
			UWaveGestureLibraryX & \textbf{2.20} & 8.71 & 7.03 & 2.54 & 16.52 & 12.64 & 7.35 & 18.76 \\
			UWaveGestureLibrary & 0.70 & 1.57 & 1.16 & \textbf{0.61} & 2.03 & 1.53 & 1.33 & 1.92 \\
			Lightning7 & \textbf{0.07} & 0.36 & 0.23 & 0.35 & 0.47 & 0.33 & 0.27 & 0.55 \\
			ToeSegmentation2 & \textbf{0.08} & 0.37 & 0.29 & 0.33 & 0.52 & 0.31 & 0.34 & 0.60 \\
			DiatomSizeReduction & \textbf{0.12} & 0.52 & 0.51 & 0.27 & 0.88 & 0.49 & 0.61 & 0.99 \\
			FaceFour & \textbf{0.07} & 0.31 & 0.21 & 0.28 & 0.38 & 0.23 & 0.25 & 0.44 \\
			GestureMidAirD3 & \textbf{0.51} & 1.67 & 1.53 & 0.87 & 1.89 & 1.26 & 1.62 & 2.31 \\
			GestureMidAirD2 & \textbf{1.01} & 1.82 & 1.86 & 1.61 & 2.40 & 1.59 & 1.46 & 2.41 \\
			GestureMidAirD1 & \textbf{0.95} & 1.77 & 1.77 & 1.77 & 2.40 & 1.75 & 1.54 & 2.07 \\
			Symbols & \textbf{0.35} & 1.65 & 1.77 & 0.47 & 3.26 & 1.85 & 2.11 & 3.61 \\
			HandMovementDirection & \textbf{0.49} & 1.10 & 0.62 & 1.71 & 1.31 & 1.18 & 0.97 & 1.71 \\
			Heartbeat & 2.40 & 3.13 & \textbf{2.12} & 4.45 & 3.27 & 3.39 & 2.72 & 4.53 \\
			Yoga & 2.79 & 5.69 & 6.92 & \textbf{1.35} & 13.27 & 7.69 & 7.56 & 13.35 \\
			OSULeaf & 0.97 & 1.86 & 1.85 & \textbf{0.69} & 2.89 & 2.18 & 1.51 & 2.46 \\
			Meat & \textbf{0.07} & 0.35 & 0.23 & 0.33 & 0.41 & 0.27 & 0.27 & 0.49 \\
			Fish & 0.74 & 1.61 & 1.95 & \textbf{0.63} & 2.78 & 1.75 & 1.56 & 2.19 \\
			FordA & \textbf{12.01} & 35.30 & 16.22 & 18.09 & 62.56 & 49.73 & 18.56 & 72.50 \\
			FordB & \textbf{12.82} & 43.55 & 22.24 & 26.76 & 94.80 & 109.70 & 23.77 & 107.19 \\
			Ham & \textbf{0.29} & 1.02 & 1.04 & 0.58 & 1.66 & 1.08 & 1.00 & 1.57 \\

		\end{longtable}
		
	\end{center}
\end{landscape}

\end{appendices}

\end{document}